\documentclass[10pt]{amsart} 

\usepackage{amsmath}
\usepackage{amssymb}

\usepackage{graphicx}
\usepackage{xcolor}
\usepackage{subcaption}
\usepackage{multirow}
\usepackage{enumitem}
\usepackage{mathrsfs}
\usepackage[rightcaption]{sidecap}
\usepackage{rotating}
\usepackage{tikz}

\newtheorem{theorem}{Theorem}[section]

\theoremstyle{definition}
\newtheorem{example}[theorem]{Example}
\theoremstyle{remark}
\newtheorem{remark}[theorem]{Remark}

\newcommand{\R}{\mathbb R}
\newcommand{\dx}{\mathrm d}

\newcommand{\JS}{\mathrm{JS}}

\newcommand{\Z}{\mathrm Z}

\newcommand{\SNN}{\mathrm{SNN}}
\newcommand{\LN}{\mathrm{LN}}
\newcommand{\GELU}{\mathrm{GELU}}
\newcommand{\erf}{\mathrm{erf}}
\newcommand{\e}{\mathrm e}
\newcommand{\bfu}{\mathbf u}
\newcommand{\bff}{\mathbf f}
\newcommand{\bfg}{\mathbf g}
\newcommand{\cfl}{\mathrm{CFL}}
\newcommand{\weno}{\mathcal{WENO}}
\newcommand{\N}{\mathcal N}
\newcommand{\vecf}{F}
\newcommand{\vecone}{\mathbf 1}

\title[A third-order WENO scheme with shallow neural network]{A third-order finite difference weighted essentially non-oscillatory scheme with shallow neural network}
\author{Kwanghyuk Park}
\address{Graduate School of Artificial Intelligence $\&$ POSTECH MINDS (Mathematical Institute for Data Science), Pohang University of Science and Technology, Pohang 37673, Korea}
\email{pkh0219@postech.ac.kr}

\author{Xinjuan Chen}
\address{Department of Mathematics, College of Science, Jimei University, Xiamen, Fujian 361021, China}
\email{chenxinjuan@jmu.edu.cn}

\author{Dongjin Lee}
\address{Graduate School of Artificial Intelligence $\&$ POSTECH MINDS (Mathematical Institute for Data Science), Pohang University of Science and Technology, Pohang 37673, Korea}
\email{dongjinlee@postech.ac.kr}

\author{Jiaxi Gu}
\address{Department of Mathematics $\&$ POSTECH MINDS (Mathematical Institute for Data Science), Pohang University of Science and Technology, Pohang 37673, Korea}
\email{jiaxigu@postech.ac.kr}

\author{Jae-Hun Jung}
\address{Department of Mathematics $\&$ POSTECH MINDS (Mathematical Institute for Data Science), Pohang University of Science and Technology, Pohang 37673, Korea}
\email{jung153@postech.ac.kr}

\makeatletter
\@namedef{subjclassname@2020}{\textup{}2020 Mathematics Subject Classification}
\makeatother

\begin{document}

\maketitle

\begin{abstract}
In this paper, we introduce the finite difference weighted essentially non-oscillatory (WENO) scheme based on the neural network for hyperbolic conservation laws. 
We employ the supervised learning and design two loss functions, one with the mean squared error and the other with the mean squared logarithmic error, where the WENO3-JS weights are computed as the labels.
Each loss function consists of two components where the first component compares the difference between the weights from the neural network and WENO3-JS weights, while the second component matches the output weights of the neural network and the linear weights. 
The former of the loss function enforces the neural network to follow the WENO properties, implying that there is no need for the post-processing layer.
Additionally the latter leads to better performance around discontinuities.
As a neural network structure, we choose the shallow neural network (SNN) for computational efficiency with the Delta layer consisting of the normalized undivided differences.
These constructed WENO3-SNN schemes shows the outperformed results in one-dimensional examples and improved behavior in two-dimensional examples, compared with the simulations from WENO3-JS and WENO3-Z. 
\end{abstract}

\section{Introduction} \label{sec:intro}
Computational fluid dynamics (CFD) has emerged as a crucial and challenging field, drawing significant interest for both theoretical and practical applications.
The complex flow structures, which includes but are not limited to shock, contact discontinuity, rarefaction wave and vortices, pose significant challenges for the numerical simulation.
Hyperbolic conservation laws, which are frequently considered in CFD, particularly demand numerical schemes to handle the complex flow structure while maintaining high-order accuracy for smooth regions. 
This dual requirement has led to the development of various numerical methods, with the essentially non-oscillatory (ENO) scheme and weighted essentially non-oscillatory (WENO) scheme among the most efficient methods.

The third-order finite volume WENO scheme was first introduced by Liu et al. \cite{Liu}, in order to increase the order of accuracy in smooth regions under the premise of retaining ENO properties for discontinuities. 
Jiang and Shu \cite{Jiang} subsequently constructed a general framework of the smoothness indicators and proposed the fifth-order finite difference WENO (WENO5-JS) scheme, which has since been widely adopted for problems containing both shocks and complicated smooth solution structures.
However, the WENO5-JS scheme is of dissipation around discontinuities, which may affect the accuracy of numerical simulations, particularly in capturing fine-scale flow structures.
Using a different approach, Borges et al. \cite{Borges} introduced the global smoothness indicator and developed Z-type nonlinear weights $\omega^\Z_k$, leading to the WENO5-Z scheme, which decreases the numerical dissipation near discontinuities and maintains the ENO behavior.
Those fifth-order WENO schemes could be easily translated to third-order WENO schemes \cite{Shu,Don}, namely WENO3-JS and WENO3-Z.

Recently machine learning has seen a growing presence within the CFD field.
One promising area is the integration of machine learning technique with the WENO schemes, where the specific model is trained offline to offer potential improvements in both accuracy and efficiency.
In \cite{Sun,Wen,Xue}, the discontinuity detector based on the neural network was applied to the hybrid WENO scheme.
Kossaczk\'{a} et al. \cite{Kossaczka} incorporated a small convolutional neural network to adjust smoothness indicators in the fifth-order WENO scheme.
The use of machine learning to directly calculate the nonlinear weights of WENO schemes is gaining popularity.
Wang et al. \cite{Wang} designed a reinforcement learning policy network to optimize the nonlinear weights of the numerical fluxes corresponding to the substencils.
In \cite{Bezgin21}, Bezgin et al. developed a convolutional neural network to output the WENO weights and the dispersion coefficient at the cell face of the finite volume scheme for the linear diffusive-dispersive regularizations of the scalar cubic conservation law.
Bezgin et al. \cite{Bezgin22} introduced the Delta layer to the multilayer perceptrons, which predicted the nonlinear weights for the third-order WENO scheme within the finite volume framework.
However, during testing, the resulting weights would pass through an ENO layer in order to retain the ENO property.

Adopting the notion of the Delta layer, we propose the artificial neural network, which mimics the WENO weighting function, to give the nonlinear weights for the third-order finite difference WENO schemes. 
We choose the shallow neural network (SNN) without the ENO layer in \cite{Bezgin22} for computational efficiency and reduced dissipation.
The training process is composed of two stages and the supervised learning is employed.
In the first stage, the neural network is trained to generate the linear weights for the smooth stencils where we use the dataset, which consists of a variety of smooth functions.
In the second stage, with the dataset from the piecewise smooth function and the WENO3-JS nonlinear weights $\omega^\JS_k$ as labels, our goal is to optimize the neural network for simulations with less dissipation around discontinuities.
Two kinds of loss functions are specified, where one is based on mean squared error and the other is the mean squared log error. 
Each loss function consists of two terms, where one term calculates the difference between the output weights and the labels for consistency, which drives the neural network to yield the nonlinear weights close to $\omega^\JS_k$ near discontinuities, whereas the other term compares the ratio of the output weights, forcing the neural network to return to the linear weights in smooth regions.
By carefully tuning the hyperparameters, the proposed WENO3-SNN schemes preserves the ENO behavior at discontinuities while achieving high-order accuracy in smooth regions.
Since the neural network learns the classical WENO3-JS weights in a direct way, we do not add any post-processing layer, such as ENO layer in \cite{Bezgin22}, in order to retain the ENO property, thus simplifying the implementation.
Additionally, the resulting numerical solution from the neural network is less dissipative by using the second term in each loss function.
Both one- and two-dimensional numerical examples are provided to illustrate the performance of the proposed WENO3-SNN schemes over WENO3-JS and WENO3-Z. 

The remainder of the paper is organized as follows. 
Section \ref{sec:weno} briefly reviews the classical third-order finite difference WENO schemes, including WENO3-JS and WENO3-Z. 
In Section \ref{sec:snn}, we introduce the architecture of the WENO3-SNNs and describe the training process in details. 
Section \ref{sec:nr} presents the order of accuracy as well as numerical performance to several one-dimensional and two-dimensional problems, which validates the new proposed schemes.
Finally, we draw a conclusion and outline the further developments in Section \ref{sec:conclusion}.

\section{WENO scheme for one-dimensional scalar equation} \label{sec:weno}
Consider the one-dimensional scalar hyperbolic equation,
\begin{equation} \label{eq:1d_scalar_hyperbolic}
 \frac{\partial u}{\partial t} + \frac{\partial f(u)}{\partial x} = 0.
\end{equation} 
We discretize the spatial domain $[a, b]$ into the uniform grid with $N$ points, 
$$ x_i = a + \frac{\Delta x}{2} + i \Delta x, \quad i = 0, \cdots, N-1, $$
where $\Delta x = (b-a)/N$ is the grid size. 
The grid point $x_i$ is also the cell center for the $i$th cell $I_i = [x_{i-1/2}, x_{i+1/2}]$ with the cell boundaries $x_{i\pm1/2} = x_i \pm \Delta x/2$. 
Fixing the time $t$ and applying the method of lines to \eqref{eq:1d_scalar_hyperbolic} gives
\begin{equation} \label{eq:1D_hyperbolic_discrete}
 \frac{du(x_i, t)}{dt} = - \left. \frac{\partial f \left( u(x,t) \right)}{\partial x} \right |_{x=x_i}.
\end{equation} 
Define the flux function $h(x)$ implicitly by
$$
   f \left( u(x) \right) = \frac{1}{\Delta x} \int^{x+\Delta x/2}_{x-\Delta x/2} h(\xi) \dx \xi, 
$$
where the time variable $t$ is dropped as it is fixed.
Differentiating both sides with respect to $x$ and evaluating at $x=x_i$, we obtain
\begin{equation} \label{eq:partial_f}
 \left. \frac{\partial f}{\partial x} \right |_{x=x_i} = \frac{h_{i+1/2} - h_{i-1/2}}{\Delta x},
\end{equation}
with $h_{i \pm 1/2} = h(x_{i \pm 1/2})$.
Combining \eqref{eq:1D_hyperbolic_discrete} and \eqref{eq:partial_f} shows that
$$
 \frac{du(x_i, t)}{dt} = - \frac{h_{i+1/2} - h_{i-1/2}}{\Delta x}.
$$
Our goal is now to reconstruct the fluxes $h_{i \pm 1/2}$ at the cell boundaries such that the numerical approximation achieves high order accuracy in smooth regions and precludes spurious oscillations around discontinuities.

\begin{figure}[htbp]
\centering
\includegraphics[width=0.7\textwidth]{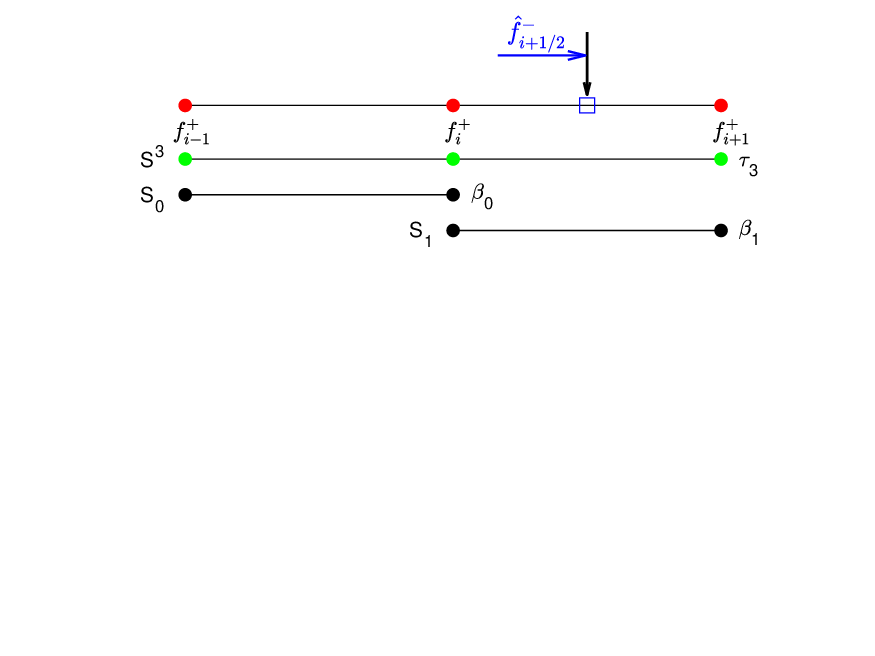}
\vspace{-4cm}
\caption{The construction of the numerical flux $\hat{f}^-_{i+1/2}$ depends on the stencil $\left( f^+_{i-1}, f^+_i, f^+_{i+1} \right)$.}
\label{fig:stencil_l}
\end{figure}

\begin{figure}[htbp]
\centering
\includegraphics[width=0.7\textwidth]{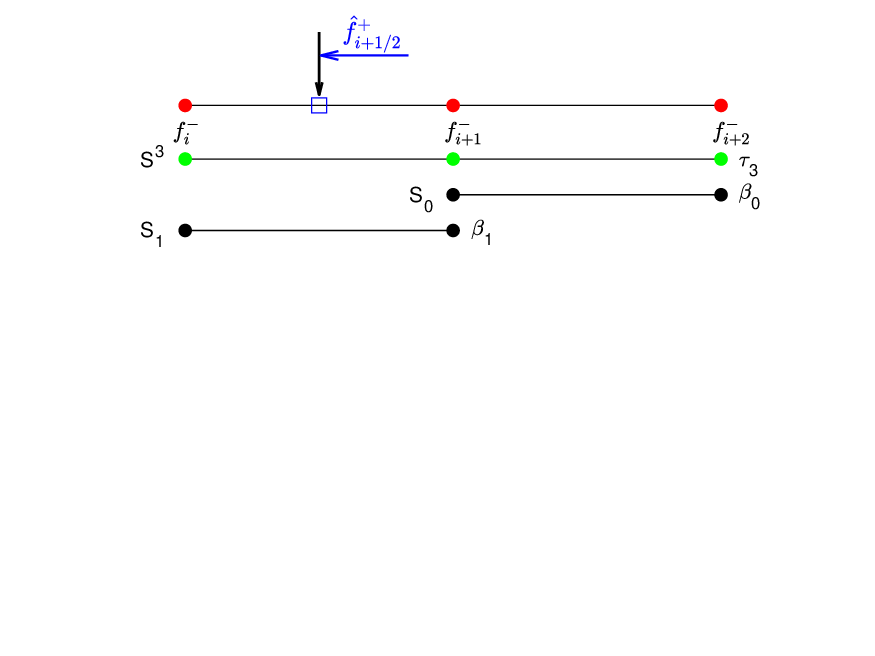}
\vspace{-4cm}
\caption{The construction of the numerical flux $\hat{f}^+_{i+1/2}$ depends on the stencil $\left( f^-_{i+2}, f^-_{i+1}, f^-_i \right)$.}
\label{fig:stencil_r}
\end{figure}

To approximate the flux $h_{i+1/2}$, we first make use of the flux splitting, for example, the Lax-Friedrichs splitting:
$$ f^{\pm}(u) = \frac{1}{2} \left( f(u) \pm \alpha u \right), $$
with $\alpha = \max_u |f'(u)|$ over the relevant range of $u$.
The numerical flux from the left-hand side $\hat{f}^-_{i+1/2}$, in Fig. \ref{fig:stencil_l}, takes the form
\begin{equation} \label{eq:numerical_flux_approximation_minus}
 \hat{f}^-_{i+1/2} = \omega_0 \hat{f}^{0+}_{i+1/2} + \omega_1 \hat{f}^{1+}_{i+1/2},
\end{equation}
where
$$ \hat{f}^{0+}_{i+1/2} = -\frac{1}{2} f^+_{i-1} + \frac{3}{2} f^+_i, \quad \hat{f}^{1+}_{i+1/2} = \frac{1}{2} f^+_i + \frac{1}{2} f^+_{i+1}, $$
with $f^+_j = f^+(u_j)$.
The key to the success of WENO is the choice of the nonlinear weights $\omega_k$, which requires
\begin{equation} \label{eq:weights}
 \omega_k \geqslant 0, \quad \omega_0 + \omega_1 = 1,
\end{equation}
for stability and consistency.
In \cite{Shu}, the smoothness indicator $\beta_k$ of the substencil $S_k$ is defined as
$$
 \beta_0 = \left( f^+_{i-1} - f^+_i \right)^2, \quad \beta_1 = \left( f^+_i - f^+_{i+1} \right)^2.
$$
The nonlinear weights $\omega^{\JS}_k$ in \eqref{eq:numerical_flux_approximation_minus} are given by
$$
 \omega^{\JS}_k = \frac{\alpha_k}{\alpha_0 + \alpha_1}, \quad \alpha_k = \frac{d_k}{(\beta_k + \varepsilon)^2}, \quad k=0,1,
$$
where $d_0 = \frac{1}{3}, \, d_1 = \frac{2}{3}$ are linear weights and $\varepsilon$ is a small constant to avoid the denominator being zero with $\varepsilon=10^{-6}$ in \cite{Shu}.
With a different approach, Don and Borges \cite{Don} introduced the global smoothness indicator $\tau_3$ as the absolute difference between $\beta_0$ and $\beta_1$,
$$ \tau_3 = \left| \beta_0 - \beta_1 \right|. $$
The Z-type nonlinear weights $\omega^\Z_k$ are defined as
$$
 \omega^\Z_k = \frac{\alpha_k}{\alpha_0 + \alpha_1}, \quad \alpha_k = d_k \left( 1 + \frac{\tau_3}{\beta_k + \varepsilon} \right), \quad k=0,1,
$$
with $\varepsilon = 10^{-40}$ in \cite{Borges}.
Similarly, the numerical flux from the right-hand side $\hat{f}^+_{i+1/2}$, in Fig. \ref{fig:stencil_r}, is of the form
\begin{gather*}
 \hat{f}^+_{i+1/2} = \omega_0 \hat{f}^{0-}_{i+1/2} + \omega_1 \hat{f}^{1-}_{i+1/2}, \\
 \hat{f}^{0-}_{i+1/2} = -\frac{1}{2} f^-_{i+2} + \frac{3}{2} f^-_{i+1}, \quad \hat{f}^{1-}_{i+1/2} = \frac{1}{2} f^-_{i+1} + \frac{1}{2} f^-_i.
\end{gather*}
The nonlinear weights $\omega_k$ can be obtained by the above WENO weighting stategies.
The numerical flux $\hat{f}_{i+1/2}$, which is the sum of $\hat{f}^-_{i+1/2}$ and $\hat{f}^+_{i+1/2}$, is the approximation of $h_{i+1/2}$.

\begin{remark} \label{rmk:weighting_function}
We can view each WENO weighting strategy as a function $\weno^\N:\R^3 \to [0, \, 1]^2$ defined by $\weno^\N (\vecf)= \omega$, with $\N \in \{ \JS, \Z \}, \, \vecf = (f_0, f_1, f_2)$ and $\omega = \left( \omega_0, \omega_1 \right)$.
Then neither of the WENO weighting functions is scale-invariant, that is, $\weno^\N (\vecf) \ne \weno^\N (\lambda \vecf)$ for some real number $\lambda$.
To see this, consider the three-point stencils $\vecf_1=(10^{-3}, 10^{-3}, 0)$ and $\vecf_2=(10^{-20}, \, 10^{-20}, \, 0)$ for the respective WENO3-JS and WENO3-Z weighting functions, and the scaling $\lambda = \sqrt{3}$.
The calculations
\begin{align*}
 \weno^\JS (\vecf_1) &= \left( \frac{2}{3}, \frac{1}{3} \right), &  \weno^\JS (\lambda \vecf_1) &= \left( \frac{8}{9},\frac{1}{9} \right), \\
 \weno^\Z  (\vecf_2) &= \left( \frac{2}{5},\frac{3}{5} \right),  &  \weno^\Z  (\lambda \vecf_2) &= \left( \frac{8}{15},\frac{7}{15} \right),
\end{align*}
verify that the WENO weighting functions are not scale-invariant.
However, both of the WENO weighting functions are translation-invariant, i.e., for all $\delta \in \R$, 
$$ \weno(\vecf + \delta \cdot \vecone) = \weno(\vecf), $$
with $\vecone = (1,\,1,\,1)$.
This is because the translation of the stencil does not change the values of the smoothness indicators $\beta_k$, and hence has no effect on the nonlinear weights.
Therefore, the WENO3-JS and WENO3-Z weighting functions are not scale-invariant but translation-invariant.
\end{remark}

\section{Shallow neural network for WENO weighting function} \label{sec:snn}
From Section \ref{sec:weno}, we see that the WENO weighting function maps the three-point stencil $\vecf$ to the nonlinear weights $\omega$.
In this section, we introduce the data-driven WENO weighting function based on the shallow neural network (SNN) with the input $\vecf$ and the output $\omega$, which mimics the WENO3-JS weighting function.

\subsection{SNN Architecture}
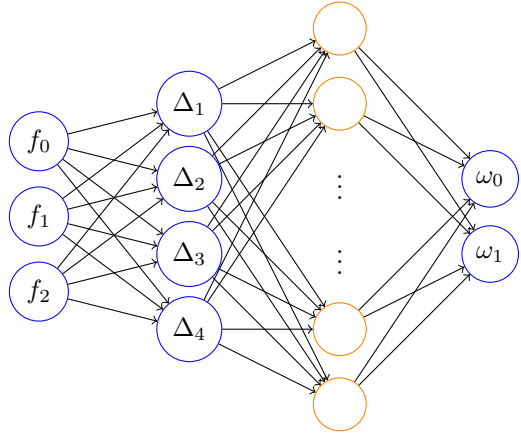
\begin{figure}[htbp]
\centering
\begin{tikzpicture}
\node[circle, minimum width=15pt, minimum height=15pt, draw=blue] (11) at(0,1)  {$f_0$};
\node[circle, minimum width=15pt, minimum height=15pt, draw=blue] (12) at(0,0)  {$f_1$};
\node[circle, minimum width=15pt, minimum height=15pt, draw=blue] (13) at(0,-1) {$f_2$};
\node[circle, minimum width=15pt, minimum height=15pt, draw=blue] (21) at(2,1.5)  {$\Delta_1$};
\node[circle, minimum width=15pt, minimum height=15pt, draw=blue] (22) at(2,0.5)  {$\Delta_2$};
\node[circle, minimum width=15pt, minimum height=15pt, draw=blue] (23) at(2,-0.5)  {$\Delta_3$};
\node[circle, minimum width=15pt, minimum height=15pt, draw=blue] (24) at(2,-1.5) {$\Delta_4$};
\node[circle, minimum width=15pt, minimum height=20pt, draw=orange] (31) at(4,2.5)  {};
\node[circle, minimum width=15pt, minimum height=20pt, draw=orange] (32) at(4,1.5)  {};
\node (33) at(4,0.5)  {$\vdots$};
\node (34) at(4,-0.5) {$\vdots$};
\node[circle, minimum width=15pt, minimum height=20pt, draw=orange] (35) at(4,-1.5) {};
\node[circle, minimum width=15pt, minimum height=20pt, draw=orange] (36) at(4,-2.5) {};
\node[circle, minimum width=15pt, minimum height=15pt, draw=blue] (41) at(6,0.5)  {$\omega_0$};
\node[circle, minimum width=15pt, minimum height=15pt, draw=blue] (42) at(6,-0.5)  {$\omega_1$};
\draw[->] (11)--(21);
\draw[->] (11)--(22);
\draw[->] (11)--(23);
\draw[->] (11)--(24);
\draw[->] (12)--(21);
\draw[->] (12)--(22);
\draw[->] (12)--(23);
\draw[->] (12)--(24);
\draw[->] (13)--(21);
\draw[->] (13)--(22);
\draw[->] (13)--(23);
\draw[->] (13)--(24);
\draw[->] (21)--(31);
\draw[->] (21)--(32);
\draw[->] (21)--(35);
\draw[->] (21)--(36);
\draw[->] (22)--(31);
\draw[->] (22)--(32);
\draw[->] (22)--(35);
\draw[->] (22)--(36);
\draw[->] (23)--(31);
\draw[->] (23)--(32);
\draw[->] (23)--(35);
\draw[->] (23)--(36);
\draw[->] (24)--(31);
\draw[->] (24)--(32);
\draw[->] (24)--(35);
\draw[->] (24)--(36);
\draw[->] (31)--(41);
\draw[->] (32)--(41);
\draw[->] (35)--(41);
\draw[->] (36)--(41);
\draw[->] (31)--(42);
\draw[->] (32)--(42);
\draw[->] (35)--(42);
\draw[->] (36)--(42);
\end{tikzpicture}
\caption{Schematic of the SNN architecture.}
\label{fig:snn}
\end{figure}

Our SNN architecture is made up of the input layer, a pre-processing layer, one hidden layer and the output layer, as shown in Fig. \ref{fig:snn}.
In the input layer, the neural network receives the three-point stencil $\vecf=(f_0, f_1, f_2)$ that is either $\left( f^+_{i-1}, f^+_i, f^+_{i+1} \right)$ or $\left( f^-_{i+2}, f^-_{i+1}, f^-_i \right)$.
Then the stencil $\vecf$ passes through the pre-processing layer called the Delta layer in \cite{Bezgin22}, where four features $\Delta_j, \, j=1,2,3,4$, are defined as 
\begin{equation} \label{eq:delta_layer}
\begin{aligned}
 & \tilde{\Delta}_1 = \left| f_0-f_1 \right|, \, 
   \tilde{\Delta}_2 = \left| f_1-f_2 \right|, \,
   \tilde{\Delta}_3 = \left| f_0-f_2 \right|, \, 
   \tilde{\Delta}_4 = \left| f_0-2f_1+f_2 \right|, \\
 & \Delta_j = \tilde{\Delta}_j / \max \left( \tilde{\Delta}_1, \tilde{\Delta}_2, \varepsilon \right), \, \varepsilon = 10^{-12}.
\end{aligned}     
\end{equation}
The Delta layer transforms the input data to the normalized undivided differences. 
Those features are used to measure the smoothness of the stencil as the ratio of each substencil within the stencil is a significant factor in determining the nonlinear weights \cite{Gu}. 
In addition, those refined features exhibit translation-invariance as the WENO weighting functions in the previous section, but scale-invariance only in the case that 
$$ 
   \max \left( \tilde{\Delta}_1, \tilde{\Delta}_2, \varepsilon \right) = \max \left( \tilde{\Delta}_1, \tilde{\Delta}_2 \right), \quad 
   \max \left( \lambda\tilde{\Delta}_1,\lambda\tilde{\Delta}_1,\varepsilon\right)=\max \left( \lambda\tilde{\Delta}_1, \lambda\tilde{\Delta}_2 \right).
$$
It is also observed empirically that this Delta layer in our neural network plays an important role in inhibiting the oscillations around discontinuities.
After the Delta layer, the features go to the hidden layer that enables the neural network to learn the relationship between the three-point stencil and the nonlinear weights.
The hidden layer is composed of $16$ neurons and the activation function, which is the Gaussian Error Linear Unit (GELU) defined by
$$ \GELU (x) = \frac{x}{2} \left( 1+\erf \left( x/ \sqrt{2} \right) \right), $$
with $\erf$ the error function. 
The GELU activation function with its smooth, bounded and stationary properties, has been successfully applied to several state-of-the-art neural network models, such as BERT \cite{Devlin}, ViT \cite{Dosovitskiy}, and GPT-3 \cite{Brown}, demonstrating its versatility and effectiveness.
The output layer produces two nonlinear weights $\omega_0$ and $\omega_1$, which satisfy the conditions \eqref{eq:weights} with the use of the softmax function.
Therefore, for the input three-point stencil $\vecf$, the output nonlinear weights $\omega$ in our neural network is calculated as follows:
\begin{equation} \label{eq:snn_weighting}
 \omega = \sigma_1 \left( W^1 \left( \sigma_0 \left( W^0 \Delta (\vecf) + b^0 \right) \right) + b^1 \right),
\end{equation}
where $\Delta$ represents the computations \eqref{eq:delta_layer} in the Delta layer, $W^0$ and $W^1$ are weight matrices of respective sizes $2\times16$ and $16\times4$, $b^0$ and $b^1$ are bias vectors of respective sizes $16\times1$ and $2\times1$, and $\sigma_0$ is the activation function GELU and $\sigma_1$ is the softmax function. 
This is referred to as the SNN based WENO weighting function $\weno^{\SNN}$ and the resulting WENO scheme is WENO3-SNN.
Note that all the elements in the weight matrices $W^0$ and $W^1$, as well as the bias vectors $b^0$ and $b^1$, are parameters of the neural network. 

\begin{remark} 
Similar to the WENO3-JS and WENO3-Z weighting functions, the WENO3-SNN weighting function is not scale-invariant but translation-invariant.
Take $\vecf_2=(10^{-20},10^{-20},0)$ again in Remark \ref{rmk:weighting_function}, and $\lambda=2$.
Then 
$$ \max \left( \tilde{\Delta}_1, \tilde{\Delta}_2, \varepsilon \right) = \max(0,10^{-8},\varepsilon)= \varepsilon, $$ 
where the constant $\varepsilon$ dominates.
As mentioned earlier, $\Delta(\vecf_2)$ and $\Delta(\lambda \vecf_2)$ would not be the same,
$$
   \Delta(\vecf_2) = (0, 10^{-8}, 10^{-8}, 10^{-8}), \quad \Delta(\lambda \vecf_2) = (0, 2\cdot10^{-8}, 2\cdot10^{-8}, 2\cdot10^{-8}),
$$
and hence $\weno^{\SNN} (\vecf_2) \ne \weno^{\SNN} (\lambda \vecf_2)$ by \eqref{eq:snn_weighting}.
This means that the scale-invariance does not hold for the WENO3-SNN weighting function.
To illustrate the translation-invariance, we return to the computations \eqref{eq:delta_layer} in the Delta layer.
It is easy to see that there is no difference between the undivided differences with translation and without translation.
Thus $\tilde{\Delta}_j, \, j=1,2,3,4$, give exactly the same values with/without the translation, implying that the output weights are identical and the WENO3-SNN weighting function is of invariance for every translation.
\end{remark}

\begin{remark}
In \cite{Bezgin22}, the trained neural network is difficult to output essentially zero weights, which might lead to spurious oscillations, negative density or negative pressure for test examples including very strong shocks, e.g., the blastwaves interaction problem.
So a post-processing layer, called ENO layer, is added after the output layer during the numerical experimentation in order to maintain the ENO property.
The ENO layer, which is essentially a sharp cutoff function, forces one nonlinear weight to zero if it is less than the threshold. 
In our SNN architecture, the ENO layer is not utilized during testing, as it is possible for the neural network \eqref{eq:snn_weighting} to output essentially zero weights shown in Table \ref{tab:weights_comparison}, and the ENO reconstruction causes more numerical dissipation near discontinuities.
Without the ENO layer in our WENO3-SNN scheme, we do not observe the significant spurious oscillations, nor the negative density/pressure for the implemented numerical experiments in Section \ref{sec:nr}.
\end{remark}

\subsection{Training}
The training is proceeded in two stages. 
First we initialize the neural network such that its output is close to the linear weights $d_k$ for the smooth stencils.
Then we employ the supervised learning with the dataset from the piecewise smooth function and the WENO3-JS nonlinear weights $\omega^\JS_k$ as labels. 
We want to optimize the the neural network that would output the linear weights in smooth regions and improve the approximate solution with less dissipation around discontinuities. 

\subsubsection{Initialization}
In the first stage, the goal is to train the parameters of the neural network, from which the weights are close to the linear weights $d_k$ for the smooth stencils.
The dataset consisting of the three-point stencils, is generated from the smooth functions that include constant functions, polynomials, trigonometric function and exponential functions.
We use the mean squared log error to define the loss function $\mathcal{L}^0$ for the initialization,
\begin{equation} \label{eq:loss_initialization}
 \mathcal{L}^0 = \frac{1}{N} \sum_{l=1}^{N} \left[ \log \left( 2 \omega^{\SNN,l}_0 \right) - \log \left( \omega^{\SNN,l}_1 \right) \right]^2,  
\end{equation}
with $N$ the number of stencils.
If $\mathcal{L}^0=0$, we have $2 \omega^{\SNN,l}_0 = \omega^{\SNN,l}_1$ for all $l$.
Combining this with the softmax function gives $\omega^{\SNN,l}_k = d_k, \, k=0,1$.
Thus the neural network is able to yield the linear weights $d_k$ under the perfect condition $\mathcal{L}^0=0$.
To train the neural network, we choose the Adam optimizer with the learning rate $10^{-3}$ and the weight decay $0.01$.
Table \ref{tab:initialization} shows the output weights from the trained neural network for the smooth stencils in the initialization stage. 
Though there is some slight deviation from the linear weights $d_k$, we obtain a good approximation of the linear weights from the neural network for the initialization.

\begin{table}[htbp]
\renewcommand{\arraystretch}{1.1}
\scriptsize
\centering
\caption{SNN weights in the initialization stage}      
\begin{tabular}{cc} 
\hline  
Stencil & $\left( \omega^\SNN_0, \, \omega^\SNN_1 \right)$ \\ 
\hline 
(1.2602, 1.5574, 1.9648) & (0.3337, 0.6663) \\
(1.0453e-4, 3.8753e-6, -3.8753e-6) & (0.3337, 0.6663) \\ 
(0, 0.2139, 0.1931) & (0.3337, 0.6663) \\
(0.0611, -0.0304, 0.0335) & (0.3337, 0.6663) \\
(-1.0329e-4, -8.1116e-5, -1.2136e-4) & (0.3337, 0.6663) \\
(-5.2602e-4, -8.3374e-3, 3.6296e-3) & (0.3337, 0.6663) \\
\hline
\end{tabular}
\label{tab:initialization}
\end{table}

\subsubsection{Training dataset}
In the second stage, the supervised learning is employed.
The training dataset is from the one-dimensional linear advection equation,
\begin{equation} \label{eq:advection_1d}
 u_t + u_x = 0, \quad -1 \leqslant x \leqslant 1,
\end{equation}
with the piecewise initial condition
\begin{equation} \label{eq:train_ic}
\begin{aligned}
 & u(x,0) = \left\{ 
   			 \begin{array}{ll} 
   			  \frac{1}{6} \left[ G(x, \beta, z-\delta) + 4 G(x, \beta, z) + G(x, \beta, z+\delta)\right], & -0.8 \leqslant x \leqslant -0.6, \\
			      1, & -0.4 \leqslant x \leqslant -0.2, \\
			      1 - \left| 10(x-0.1) \right|, & 0 \leqslant x \leqslant 0.2, \\
			      \frac{1}{6} \left[ F(x, \alpha, y-\delta) + 4 F(x, \alpha, y) + F(x, \alpha, y+\delta)\right], & 0.4 \leqslant x \leqslant 0.6, \\
			      0, & \text{otherwise},
			     \end{array} 
           \right. \\
 & G(x, \beta, z) = \e^{-\beta(x-z)^2}, \, F(x, \alpha, y) = \sqrt{\max \left( 1-\alpha^2 (x-y)^2, 0 \right)}, \\
 & \delta = 0.005, \, \beta = \frac{\ln 2}{36 \delta^2}, \, z = -0.7, \, \alpha = 10, \, y = 0.5.
\end{aligned}    
\end{equation}
The input data, which consists of the three-point stencils, is obtained by applying a uniform grid with $\Delta x = 0.01$ to the spatial domain $[-1,1]$ and then using the Lax-Friedrichs splitting.
The corresponding labels are the nonlinear weights $\omega^\JS = (\omega^\JS_0, \omega^\JS_1)$ computed by the WENO3-JS weighting function.

\subsubsection{Loss functions}
When designing the loss functions, we expect to include the WENO properties:
\begin{enumerate}[label=\arabic*., font=\itshape]
\item In smooth regions, the nonlinear weights $\omega_k$ follow the linear weights $d_k$ in order to achieve the spatial third-order accuracy.
\item When there exists a discontinuity, the nonlinear weight $\omega_k$ corresponding to the discontinuous substencil is close to $0$, which guarantees the ENO performance.
\end{enumerate}
Note that the first property is partially done in the initialization stage.
We wish to build the neural network that yields the linear weights over the smooth stencils and preserves the ENO performance around discontinuities with less dissipation.
To manifest those WENO properties, we consider two loss functions, of which each involves two parts.

\begin{figure}[htbp]
\centering
\includegraphics[height=0.3\textwidth]{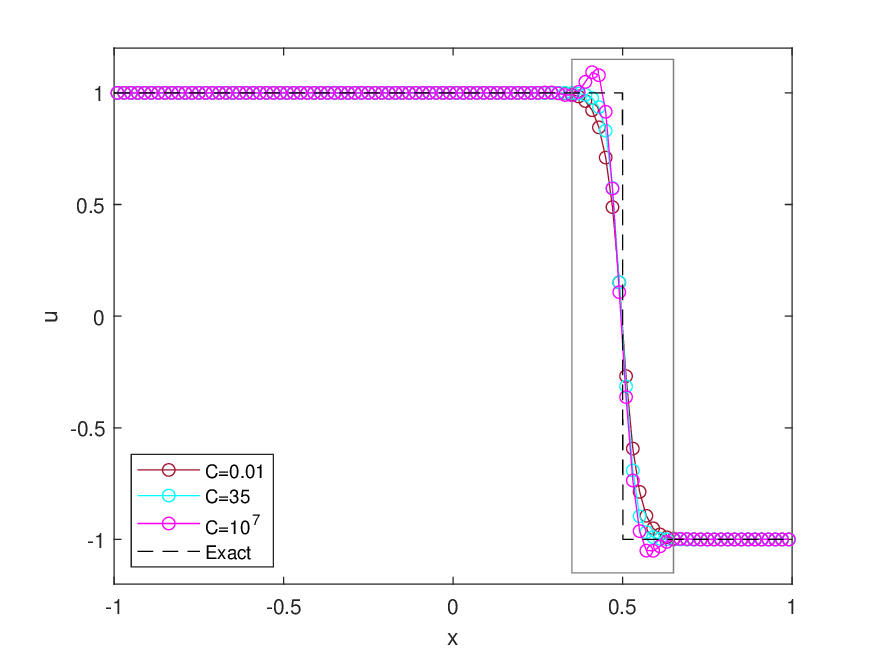}
\includegraphics[height=0.3\textwidth]{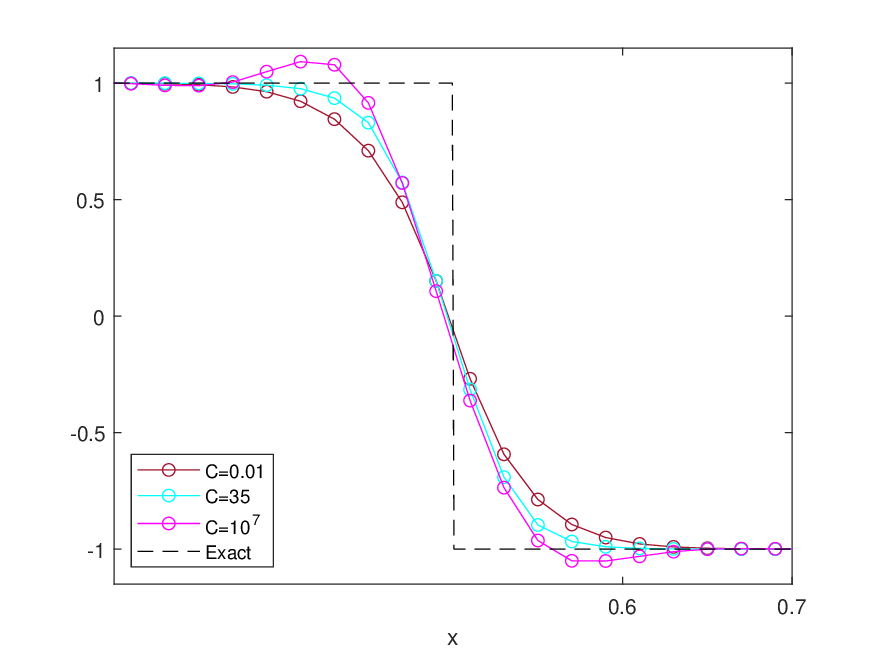}
\caption{Solution profiles with $C=0.01, \, 35, \, 10^7$ for the Riemann problem of \eqref{eq:advection_1d} at $T=0.5$ approximated by WENO3-SNN1. 
The dashed black line is the exact solution.}
\label{fig:MSE_C}
\end{figure}

Before defining the first loss function, we introduce the smoothness indicator $\lambda_l$ for the $l$th stencil by
\begin{equation} \label{eq:SI}
 \lambda_l = \e^{-(r_l-1)/C}, \quad 
 r_l = \max \left\{ \frac{2 \omega^{\JS,l}_0}{\omega^{\JS,l}_1}, \frac{\omega^{\JS,l}_1}{2 \omega^{\JS,l}_0} \right\}.
\end{equation}
The parameter $r_l$, derived from WENO3-JS nonlinear weights $\omega^\JS_k$, estimates the extent to which the stencil contains the discontinuity.
If the stencil is smooth, the nonlinear weights $\omega^\JS_k$ are close to $d_k$, and as a result $r_l$ approaches to one.
With a discontinuity in the stencil, $r_l$ is different from $1$ by many orders of magnitude.
It follows that $\lambda_l$ is close to one for the smooth stencil, whereas $\lambda_l$ goes to zero when there is a discontinuity.
Based on the smoothness indicator $\lambda_l$, the loss function $\mathcal{L}^1$, with the mean squared error (MSE), is defined by
\begin{equation} \label{eq:loss_local}
\begin{aligned}
 \mathcal{L}^1 &= \mathcal{L}^1_\JS + \mathcal{L}^1_\LN, \\
 \mathcal{L}^1_{\JS} &= \sum_{l=1}^N \left( 1-\lambda_l \right) \sum_{k=0}^1 \left( \omega^{\SNN,l}_k-\omega^{\JS,l}_k \right)^2, \\
 \mathcal{L}^1_{\LN} &= \sum_{l=1}^N \lambda_l \left( 2\omega^{\SNN,l}_0-\omega^{\SNN,l}_1 \right)^2,
\end{aligned}
\end{equation}
where $N$ is the number of stencils, and $\omega^{\SNN,l}_k, \, k=0,1$ are the nonlinear weights of the $l$th stencil from the neural network.
The first part $\mathcal{L}^1_{\JS}$ is essential for the neural network to learn the WENO3-JS weighting function.
We notice that the loss function with only the $\mathcal{L}_{\JS}^1$ part ($\lambda_l=0$) maintains the ENO behavior.
However, the approximation from the WENO3-JS scheme is more dissipative than WENO3-Z near the discontinuities. 
In order to decrease the dissipation around the discontinuities and limit the nonlinear weights $\omega^\SNN_k$ to the linear weights in smooth regions, we add the linear part $\mathcal{L}_{\LN}^1$ to the loss function.
The smoothness indicator $\lambda_l$ is expected to have an effect on the nonlinear weights $\omega^{\SNN,l}_k$ for the $l$th stencil.
For the smooth stencil, $\lambda_l \approx 1$ and the linear part dominates, which returns the nonlinear weights to $d_k$. 
But for the stencil containing a discontinuity, $\lambda_l \approx 0$ and the linear part is negligible, so $1-\lambda_l$ drives the neural network to learn the WENO3-JS weighting function.
The hyperparameter $C$ in \eqref{eq:SI} quantifies how much the output weights from the neural network match the linear weights $d_k$.
For $C$ sufficiently small, $\lambda_l$ is always close to $0$, where the $\mathcal{L}^1_{\JS}$ part controls the loss function.
Then the nonlinear weights $\omega^\SNN_k$ resemble $\omega^\JS_k$, which causes more dissipation around discontinuities.
However, as $C$ increases, the dissipation around discontinuities will be reduced, but some oscillations may occur. 
Fig. \ref{fig:MSE_C} shows the performance of different values of $C$ to the simulations of the Riemann problem for one-dimensional linear advection equation \eqref{eq:advection_1d} at the final time $T=0.5$, which agrees with our expectations above.
According to the numerical results in Fig. \ref{fig:MSE_C}, we choose $C=35$ for the smoothness indicator $\lambda_l$.

\begin{figure}[htbp]
\centering
\includegraphics[height=0.3\textwidth]{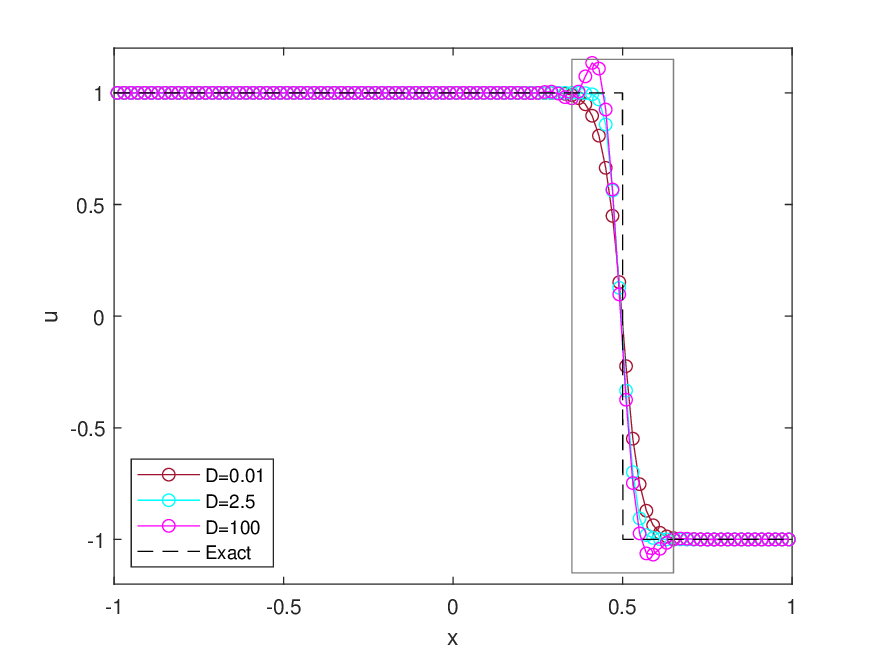}
\includegraphics[height=0.3\textwidth]{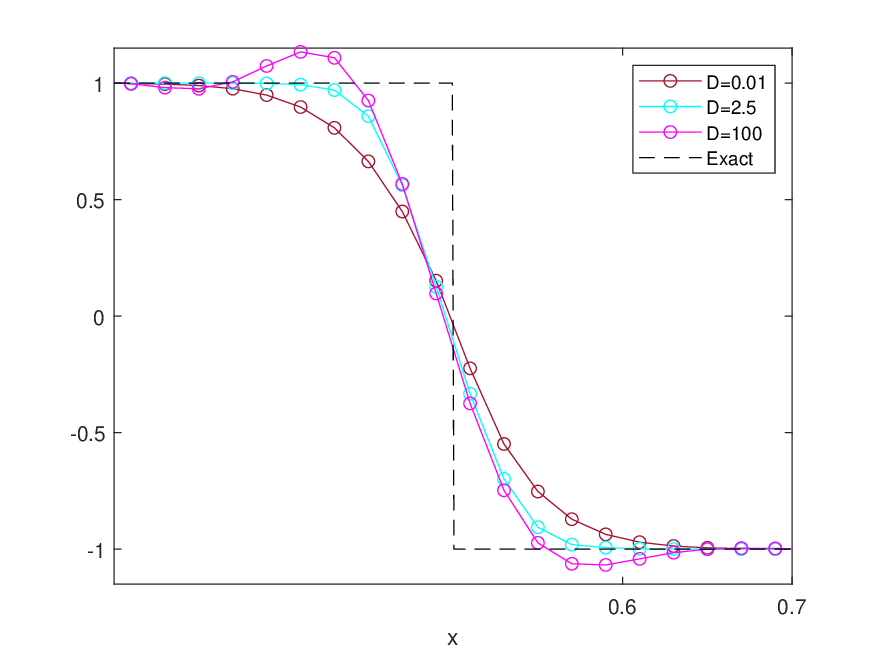}
\caption{Solution profiles with $D=0.01, \, 2.5, \, 100$ for the Riemann problem of \eqref{eq:advection_1d} at $T=0.5$ approximated by WENO3-SNN2.
The dashed black line is the exact solution.}
\label{fig:MSLE_D}
\end{figure}

We define the other loss function $\mathcal{L}^2$, with the mean squared log error (MSLE), as
\begin{equation} \label{eq:loss_global}
\begin{aligned}
 \mathcal{L}^2 &= \mathcal{L}^2_\JS + D \, \mathcal{L}^2_\LN, \\
 \mathcal{L}^2_\JS &= \sum_{l=1}^N \sum_{k=0}^1 \left[ \log\left(\omega^{\SNN,l}_k\right)-\log\left(\omega^{\JS,l}_k\right) \right]^2, \\
 \mathcal{L}^2_\LN &= \sum_{l=1}^N \left[ \log \left( 2 \omega^{\SNN,l}_0 \right) - \log \left( \omega^{\SNN,l}_1 \right) \right]^2.
\end{aligned}
\end{equation}
Similar to the loss function $\mathcal{L}^1$, this loss function is composed of two parts: the $\mathcal{L}^2_\JS$ part guides the neural network in learning the WENO3-JS weighting function, whereas the linear part $\mathcal{L}^2_\LN$, which takes the same form as in \eqref{eq:loss_initialization}, reduces the numerical dissipation near discontinuities and pushes the output weights $\omega^\SNN_k$ towards the linear weights $d_k$.
To achieve the same effect as the hyperparameter $C$ in \eqref{eq:SI}, we scale $\mathcal{L}^2_\LN$ by the hyperparameter $D$.
Unlike $\mathcal{L}^1$, this hyperparameter $D$ acts on every component in the linear part $\mathcal{L}^2_{LN}$ equally regardless of the smoothness of the $l$th stencil.
Fig. \ref{fig:MSLE_D} shows that how this neural network performs with different values of $D$ for the Riemann problem of the one-dimensional linear advection equation \eqref{eq:advection_1d} at the final time $T=0.5$.
We can see that the numerical dissipation diminishes with the growth of $D$, but the spurious oscillations around the discontinuity are noticeable if $D$ is large to some degree.  
Thus we set $D = 2.5$ in \eqref{eq:loss_global}, based on the numerical results in Fig. \ref{fig:MSLE_D}.

We apply the Adam optimizer with the same setting (learning rate $10^{-3}$ and weight decay $0.01$) as in the initialization stage, for the two neural networks, which are referred as WENO3-SNN1 and WENO3-SNN2 for the loss functions $\mathcal{L}^1$ \eqref{eq:loss_local} and $\mathcal{L}^2$ \eqref{eq:loss_global}, respectively.
Table \ref{tab:weights_comparison} compares the nonlinear weights from different WENO weighting functions.
We see that each WENO3-SNN weighting function could produce the essentially zero weight.
\begin{table}[htbp]
\renewcommand{\arraystretch}{1.1}
\scriptsize
\centering
\caption{Comparison of nonlinear weights from different WENO weighting functions}      
\begin{tabular}{ccccc}
\hline 
Stencil & \multicolumn{4}{c}{$\left( \omega_0, \, \omega_1 \right)$} \\
          \cline{2-5}
        & WENO3-JS & WENO3-Z & WENO3-SNN1 & WENO3-SNN2 \\
\hline         
(1, 1, 0)                & (1-2.0000e-12, 2.0000e-12) & (1, 4.0000e-40)     & (0.9972, 2.7788e-3) & (0.9984, 1.5913e-3) \\
(0, 1, 1)                & (5.0000e-13, 1-5.0000e-13) & (1.0000e-40, 1)     & (9.5522e-3, 0.9905) & (3.8937e-4, 0.9996) \\
(1, 0.95, 0)             & (9.9998e-1, 1.5359e-5)     & (0.9891, 0.0109)    & (0.9924, 7.6075e-3) & (0.9890, 0.0110) \\
(0.0628, 0.0314, 0.9997) & (1-2.2161e-6, 2.2161e-6)   & (0.9958, 4.1865e-3) & (1, 5.4471e-32)     & (1, 2.4276e-37) \\
(0.0286, 0.9999, 0.9686) & (5.4028e-7, 1-5.4028e-7)   & (1.0368e-3, 0.9990) & (0.0136, 0.9864)    & (1.0662e-3, 0.9989) \\
(0.0157, 0.9843, 0.9529) & (5.5334e-7, 1-5.5334e-7)   & (1.0493e-3, 0.9990) & (0.0133, 0.9867)    & (9.9179e-4, 0.9990) \\
\hline
\end{tabular}
\label{tab:weights_comparison}
\end{table}

\section{Numerical examples} \label{sec:nr}
In this section, we present some numerical experiments to compare the proposed WENO scheme, WENO3-SNN1 and WENO3-SNN2, with the WENO3-JS and WENO3-Z schemes.
We use the one-dimensional linear advection equation and Euler equations to verify the order of accuracy of the WENO schemes in terms of $L_1$ and $L_{\infty}$ error norms.
The rest of examples show the numerical results from WENO3-SNNs, in comparison with WENO3-JS and WENO3-Z.
We choose $\epsilon = 10^{-6}$ for the WENO3-JS scheme whereas $\epsilon = 10^{-40}$ for WENO3-Z.
For one- and two-dimensional scalar problems, we use the Lax-Friedrich flux splitting. 
For one-dimensional system problems, we apply the characteristic-wise Lax-Friedrich flux splitting. 
For two-dimensional system problems, we implement the schemes with the characteristic-wise Lax-Friedrich flux splitting in a dimension-by-dimension fashion. 
As for the time integration, we employ the explicit third-order total variation diminishing Runge-Kutta method \cite{ShuOsherI} 
\begin{align*}
 u^{(1)} &= u^n + \Delta t L(u^n), \\
 u^{(2)} &= \frac{3}{4} u^n + \frac{1}{4} u^{(1)} + \frac{1}{4} \Delta t L \left( u^{(1)} \right), \\
 u^{n+1} &= \frac{1}{3} u^n + \frac{2}{3} u^{(2)} + \frac{2}{3} \Delta t L \left( u^{(2)} \right),
\end{align*}
where $L$ is the spatial operator.
We set $\cfl = 0.4$. 

\subsection{One-dimensional scalar problems} 

\begin{example} \label{ex:advection_1d_order}
We first examine the order of accuracy for the one-dimensional linear advection equation \eqref{eq:advection_1d} with the initial condition $u(x,0) = \sin(\pi x)$ and the periodic boundary condition. 
The exact solution is given by $u(x,t)= \sin \left( \pi (x-t) \right)$.
The numerical solution is computed up to the final time $T=2$ with the time step $\Delta t = \cfl \cdot \Delta x$. 

The $L_1$ and $L_{\infty}$ errors versus $N$, as well as the order of accuracy, for the WENO3-JS, WENO3-Z, WENO3-SNN1, and WENO3-SNN2 schemes are displayed in Tables \ref{tab:advection_1d_order_L1} and \ref{tab:advection_1d_order_Linf}, respectively.
Although none of the WENO schemes achieves the expected order of accuracy, both WENO3-SNN schemes perform better than WENO3-JS and WENO3-S in terms of accuracy and convergence. 
\end{example}

\begin{table}[htbp]
\renewcommand{\arraystretch}{1.1}
\scriptsize
\centering
\caption{$L_1$ error and numerical order for Example \ref{ex:advection_1d_order}.}      
\begin{tabular}{clcrlcrlcrlc} 
\hline  
N & \multicolumn{2}{l}{WENO3-JS} & & \multicolumn{2}{l}{WENO3-Z} & & \multicolumn{2}{l}{WENO3-SNN1} & & \multicolumn{2}{l}{WENO3-SNN2} \\ 
    \cline{2-3}                      \cline{5-6}                     \cline{8-9}                        \cline{11-12}                   
  & Error & Order                & & Error & Order               & & Error & Order                  & & Error & Order \\
\hline 
10  & 2.99e-1 & --     & & 2.22e-1 & --     & & 2.10e-1 & --     & & 1.75e-1 & --     \\  
20  & 9.05e-2 & 1.7226 & & 7.25e-2 & 1.6136 & & 6.89e-2 & 1.6061 & & 5.29e-2 & 1.7232 \\  
40  & 3.82e-2 & 1.2437 & & 2.04e-2 & 1.8277 & & 1.71e-2 & 2.0135 & & 1.31e-2 & 2.0117 \\
80  & 9.58e-3 & 1.9955 & & 4.81e-3 & 2.0850 & & 3.85e-3 & 2.1471 & & 2.92e-3 & 2.1667 \\ 
160 & 2.33e-3 & 2.0414 & & 1.06e-3 & 2.1898 & & 7.88e-4 & 2.2904 & & 6.36e-4 & 2.2004 \\
\hline
\end{tabular}
\label{tab:advection_1d_order_L1}
\end{table}

\begin{table}[htbp]
\renewcommand{\arraystretch}{1.1}
\scriptsize
\centering
\caption{$L_{\infty}$ error and numerical order for Example \ref{ex:advection_1d_order}.}      
\begin{tabular}{clcrlcrlcrlc} 
\hline  
N & \multicolumn{2}{l}{WENO3-JS} & & \multicolumn{2}{l}{WENO3-Z} & & \multicolumn{2}{l}{WENO3-SNN1} & & \multicolumn{2}{l}{WENO3-SNN2} \\ 
    \cline{2-3}                      \cline{5-6}                     \cline{8-9}                        \cline{11-12}                   
  & Error & Order                & & Error & Order               & & Error & Order                  & & Error & Order \\
\hline 
10  & 5.30e-1 & --     & & 4.31e-1 & --     & & 4.19e-1 & --     & & 3.62e-1 & --     \\  
20  & 2.09e-1 & 1.3433 & & 1.51e-1 & 1.5135 & & 1.42e-1 & 1.5603 & & 1.22e-1 & 1.5711 \\  
40  & 8.74e-2 & 1.2573 & & 5.91e-2 & 1.3526 & & 5.35e-2 & 1.4078 & & 4.60e-2 & 1.4041 \\
80  & 3.50e-2 & 1.3180 & & 2.22e-2 & 1.4135 & & 1.94e-2 & 1.4669 & & 1.69e-2 & 1.4470 \\ 
160 & 1.36e-2 & 1.3644 & & 8.14e-3 & 1.4474 & & 6.84e-3 & 1.5021 & & 6.08e-3 & 1.4727 \\
\hline
\end{tabular}
\label{tab:advection_1d_order_Linf}
\end{table}

\begin{figure}[htbp]
\centering
\includegraphics[height=0.33\textwidth]{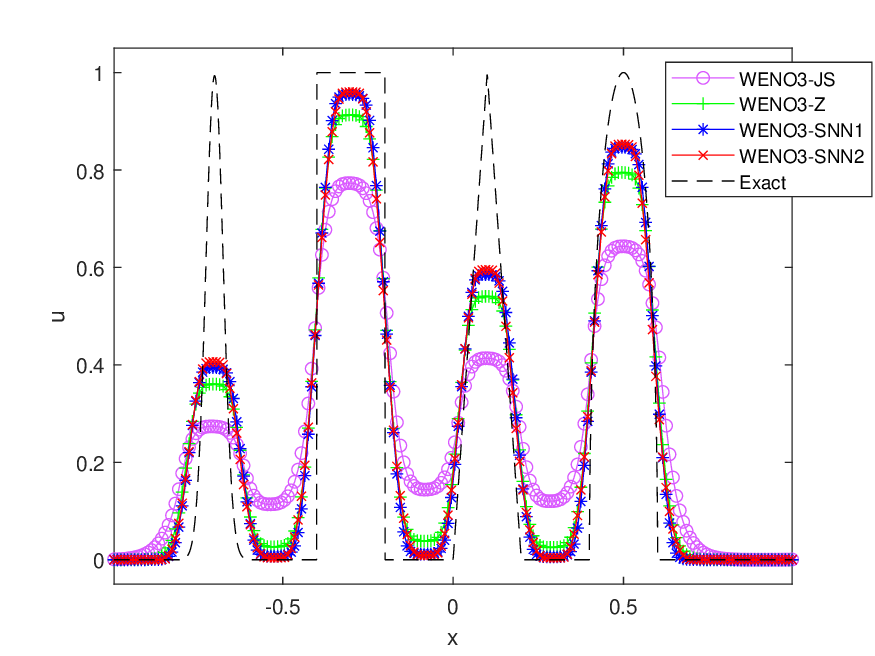}
\includegraphics[height=0.33\textwidth]{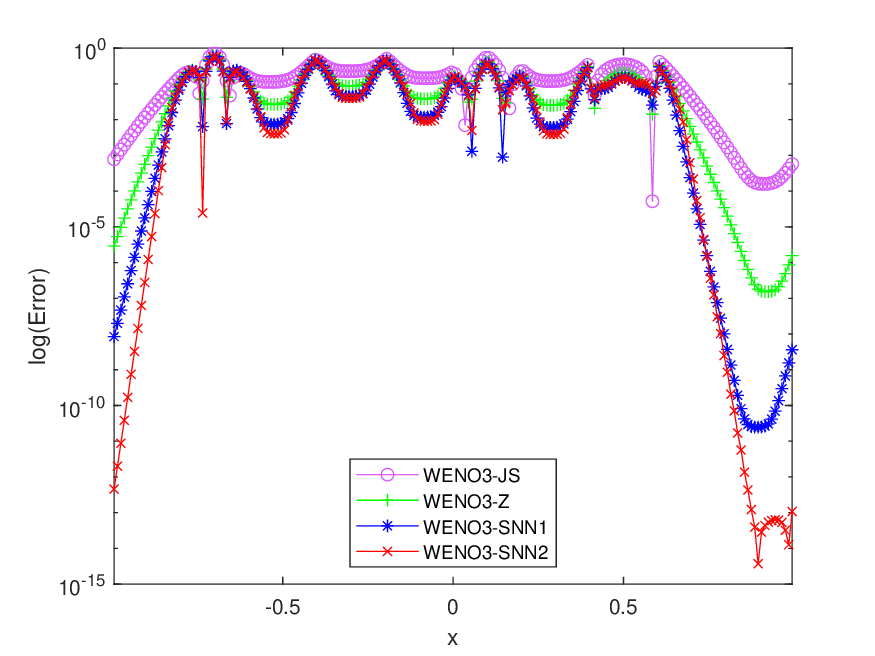}
\caption{Solution profiles (left) and log-scale pointwise errors (right) for Example \ref{ex:advection_1d} at $T=8$ approximated by WENO3-JS (purple), WENO3-Z (green), WENO3-SNN1 (blue) and WENO3-SNN2 (red) with $N = 200$. 
The dashed black line is the exact solution.}
\label{fig:advection_1d}
\end{figure}

\begin{example} \label{ex:advection_1d}
We continue to solve the above advection equation \eqref{eq:advection_1d} with the initial condition given by \eqref{eq:train_ic} and the periodic boundary condition.
The computational domain $[-1,1]$ is divided into $N=200$ uniform cells.
The final time is $T=8$ and the time step is $\Delta t = \cfl \cdot \Delta x$.
Fig. \ref{fig:advection_1d} displays the numerical solutions and the log-scale pointwise errors at the grid points, showing an overall improvement of WENO3-SNNs over the other two WENO schemes.
\end{example}

\begin{figure}[htbp]
\centering
\includegraphics[width=0.325\textwidth]{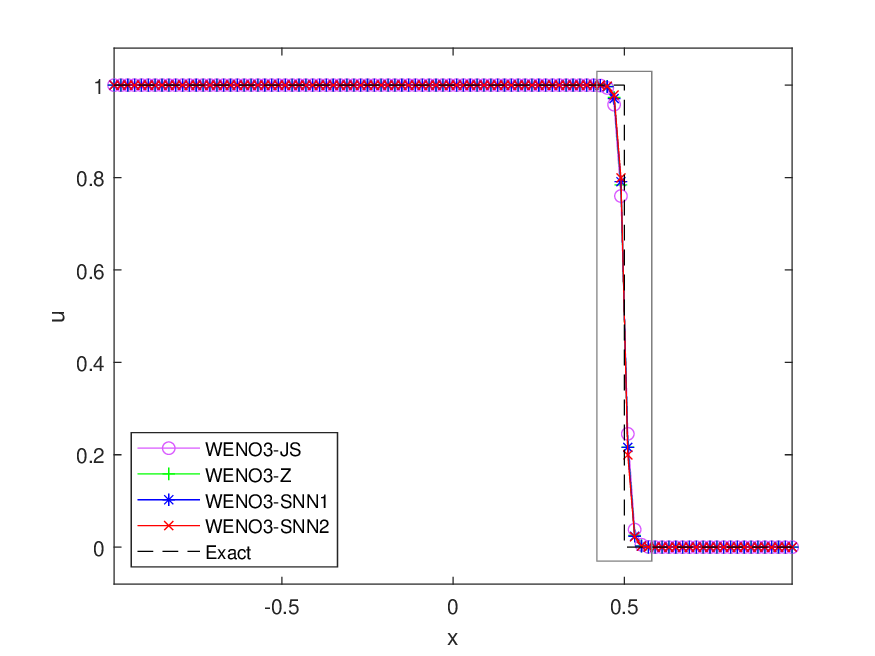}
\includegraphics[width=0.325\textwidth]{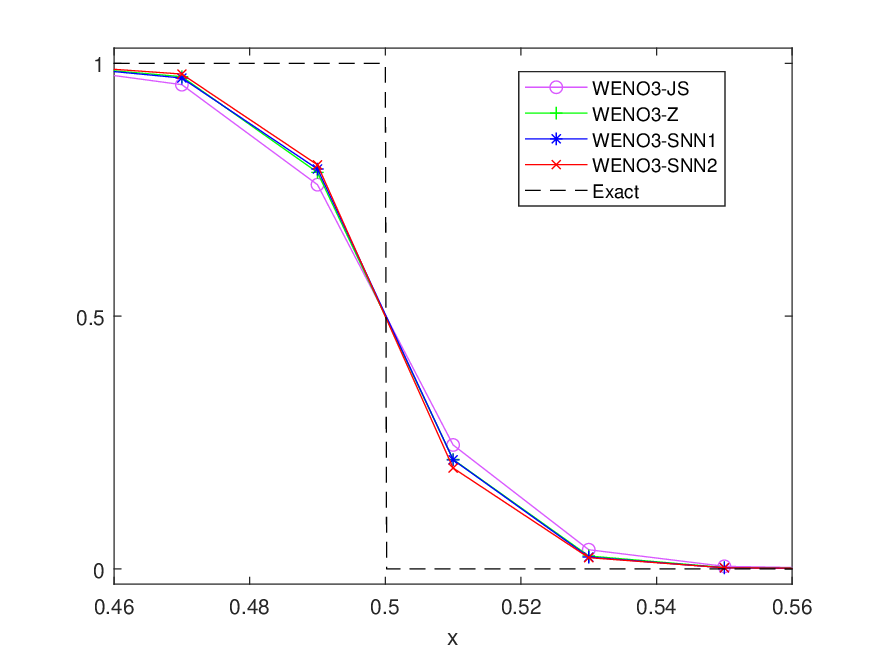}
\includegraphics[width=0.325\textwidth]{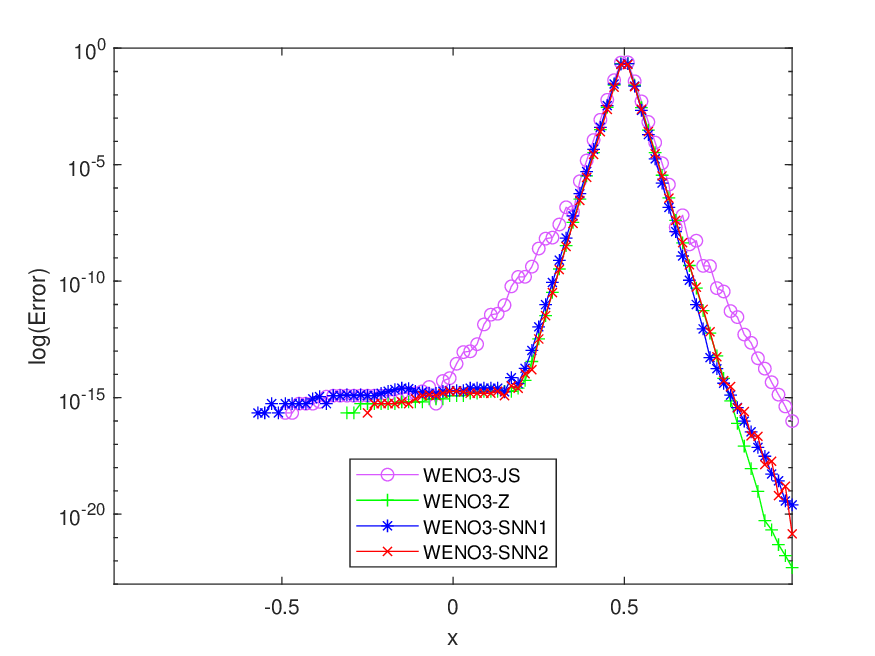}
\caption{Solution profiles for Example \ref{ex:burgers_1d} at $T=1$ (left), close-up view of the solutions in the box (middle) and log-scale pointwise error (right) approximated by WENO3-JS (purple), WENO3-Z (green), WENO3-SNN1 (blue) and WENO3-SNN2 (red) with $N = 100$. 
The dashed black line is the exact solution.}
\label{fig:burgers}
\end{figure}

\begin{example} \label{ex:burgers_1d}
Consider the Riemann problem for the Burgers' equation 
\begin{align*}
 u_t + \left(\frac{1}{2} u^2 \right)_x &= 0, \\
                                u(x,0) &= \left\{ 
                                           \begin{array}{ll} 
                                            1, & x \leqslant 0, \\ 
                                            0, & x > 0,
                                           \end{array} 
                                          \right.
\end{align*}
to which the exact solution is also a shock wave
$$
   u(x,t) = \left\{ 
             \begin{array}{ll} 
              1, & x - \frac{1}{2} t \leqslant 0, \\ 
              0, & x - \frac{1}{2} t > 0.
             \end{array} 
            \right.
$$
The shock moves to the right at the position $x = \frac{1}{2} t$ for $t>0$.
We divide the computational domain $[-1, 1]$ into $N = 100$ uniform cells.
Fig. \ref{fig:burgers} shows the approximate solutions by WENO schemes with the exact solution at the final time $T = 1$, and the pointwise errors on a logarithmic scale.
We can see that WENO3-JS, WENO3-Z, WENO3-SNN1 and WENO3-SNN2 with less dissipation yield increasingly sharper approximations around the shock while keeping the smooth regions without noticeable oscillations.
Among all, WENO3-SNN2 exhibits the most accurate solution profile around the shock.
\end{example}

\begin{figure}[htbp]
\centering
\includegraphics[width=0.325\textwidth]{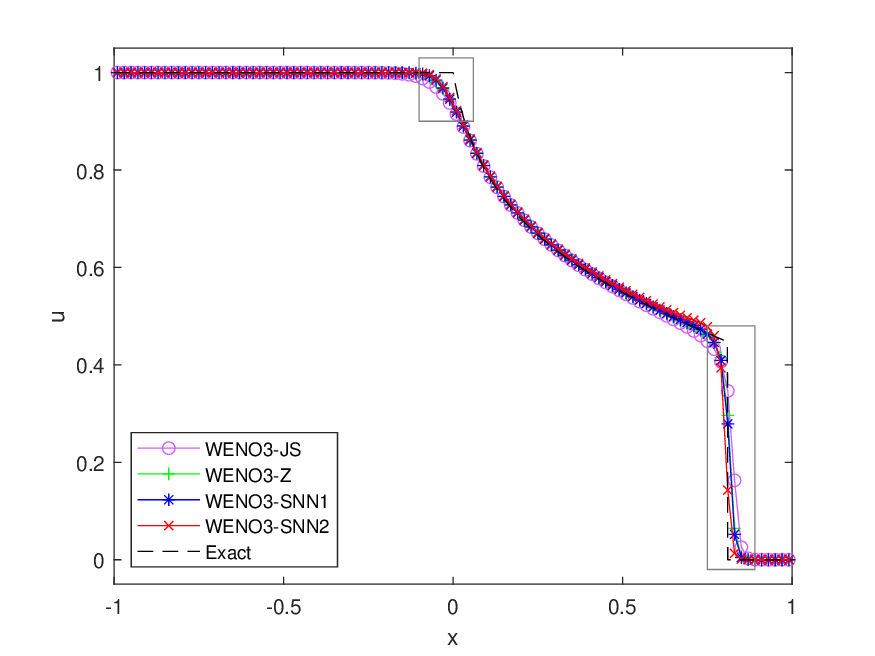}
\includegraphics[width=0.325\textwidth]{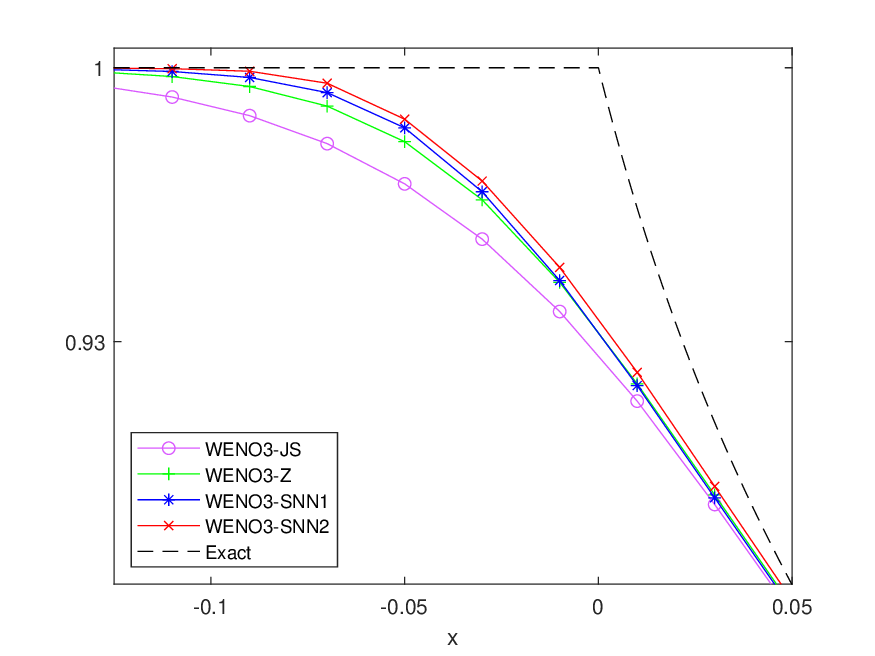}
\includegraphics[width=0.325\textwidth]{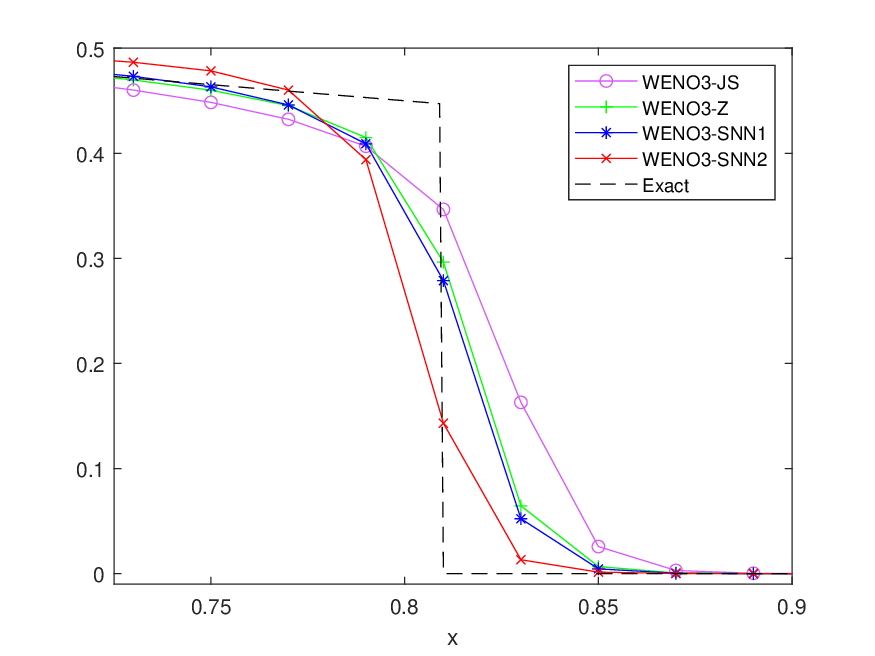}
\caption{Solution profiles for Example \ref{ex:buckley} at $T=0.5$ (left), close-up view of the solutions in the boxes from left to right (middle, right) approximated by WENO3-JS (purple), WENO3-Z (green), WENO3-SNN1 (blue) and WENO3-SNN2 (red) with $N = 80$. 
The dashed black line is the exact solution.}
\label{fig:buckley}
\end{figure}

\begin{example} \label{ex:buckley}
The Buckley-Leverett equation is of the form \eqref{eq:1d_scalar_hyperbolic} with the nonconvex flux
$$ f(u)= \frac{4u^2}{4u^2+(1-u)^2}. $$
We test the Riemann problem with the initial condition set as 
$$
   u(x,0) = \left\{ 
             \begin{array}{ll} 
              1, & x \leqslant 0, \\ 
              0, & x > 0.
             \end{array} 
            \right.
$$
The computational domain $[-1, \, 1]$ is discretized with $N = 80$ grid points and the final time is $T=0.5$ in the simulation.
Numerical results are given in Fig. \ref{fig:buckley}.
It shows that WENO3-SNN2 yields sharper solution than WENO3-JS, WENO3-Z and WENO3-SNN1 near the shock front.
\end{example}

\begin{figure}[htbp]
\centering
\includegraphics[height=0.3\textwidth]{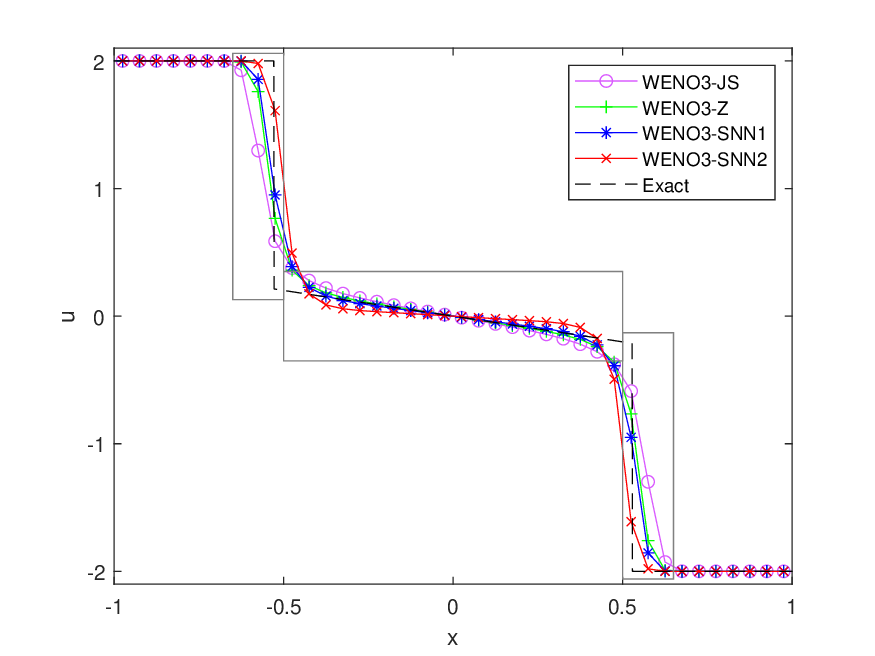}
\includegraphics[height=0.3\textwidth]{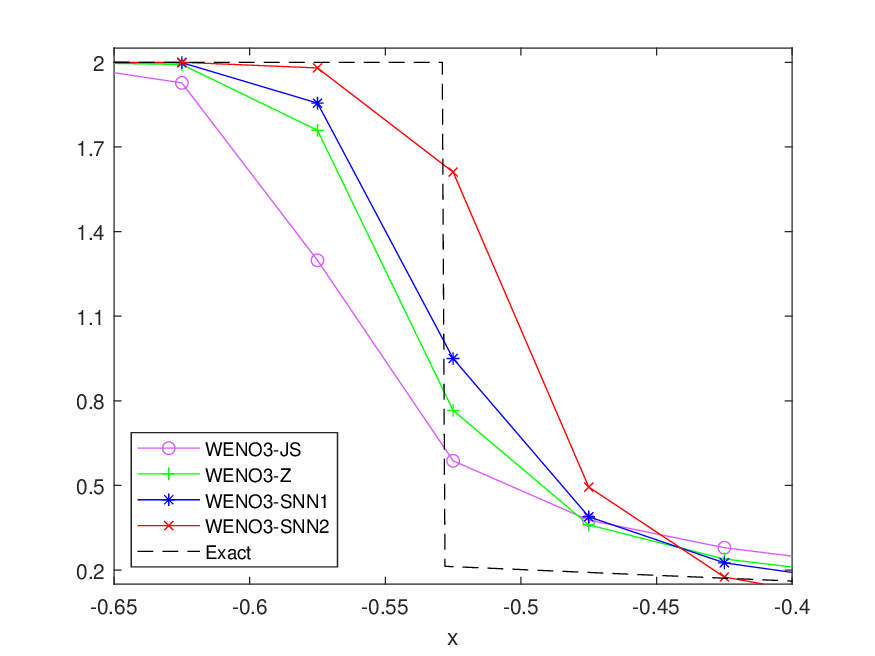}
\includegraphics[height=0.3\textwidth]{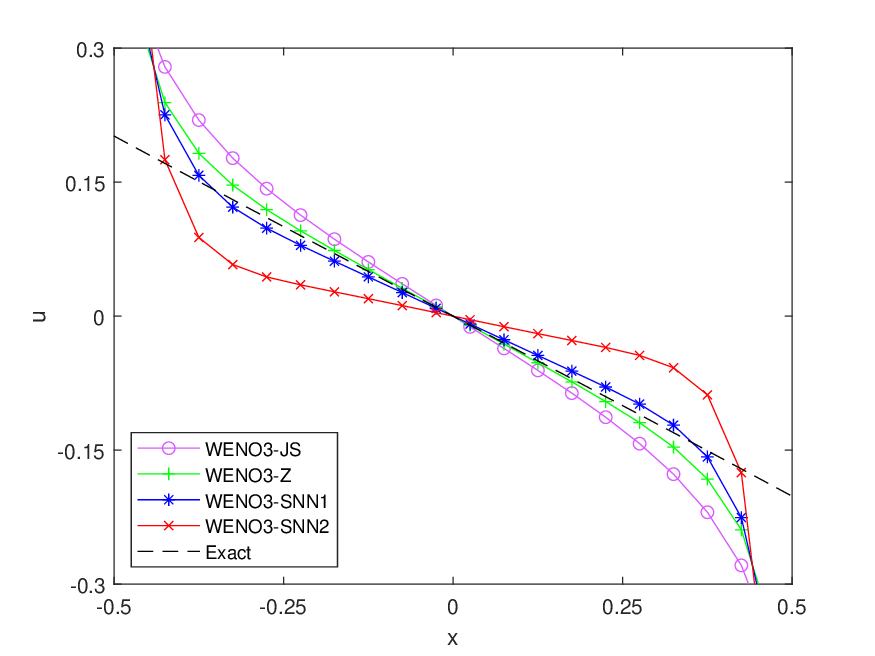}
\includegraphics[height=0.3\textwidth]{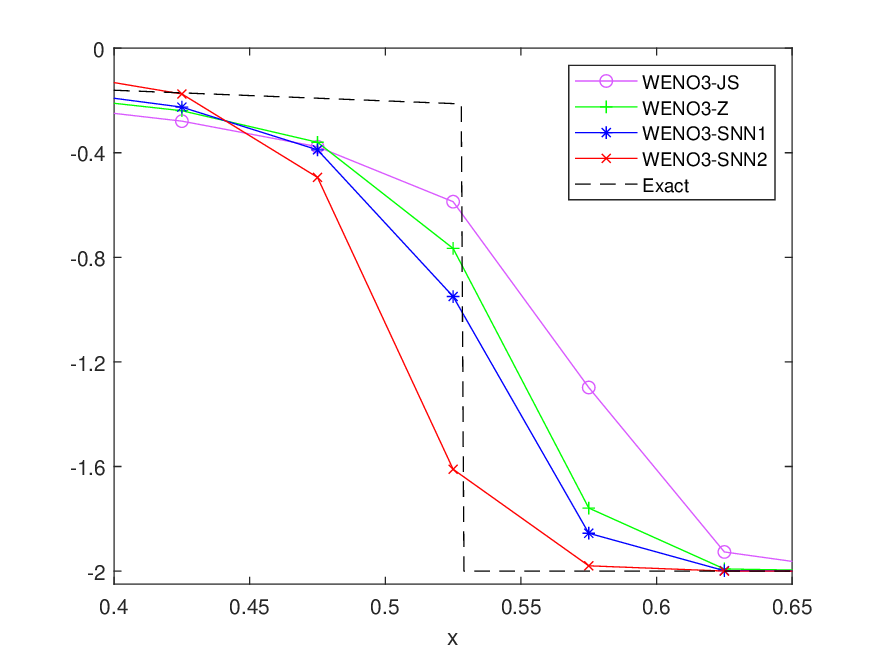}
\caption{Solution profiles for Example \ref{ex:nonconvex} with $u_L = 2$ and $u_R = -2$ at $T = 1$ (left), close-up view of the solutions in the boxes (top right, bottom left, bottom right) approximated by WENO3-JS (purple), WENO3-Z (green), WENO3-SNN1 (blue) and WENO3-SNN2 (red) with $N = 40$.
The dashed black line is the exact solution.}
\label{fig:nonconvex_IC0}
\end{figure}

\begin{figure}[htbp]
\centering
\includegraphics[width=0.495\textwidth]{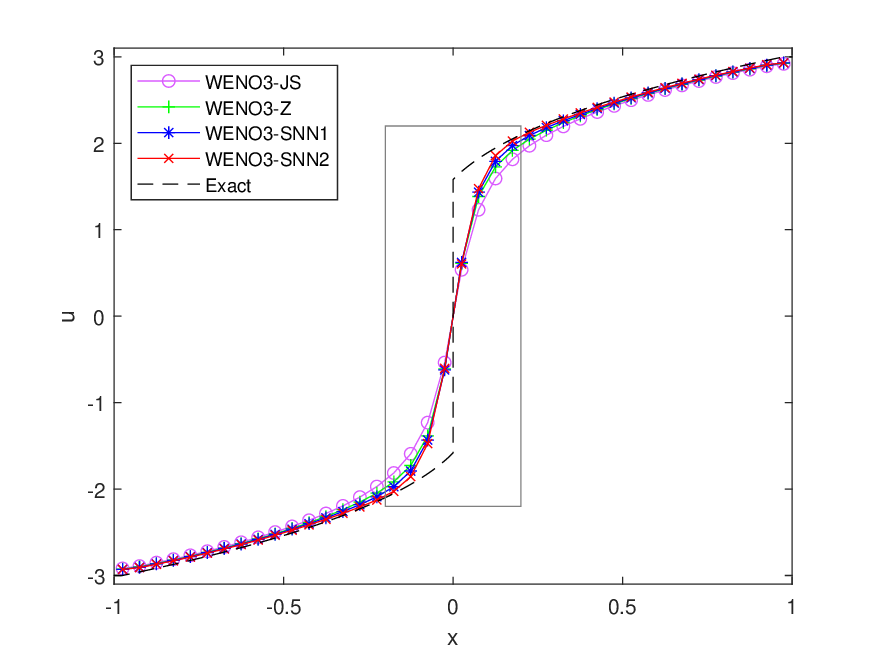}
\includegraphics[width=0.495\textwidth]{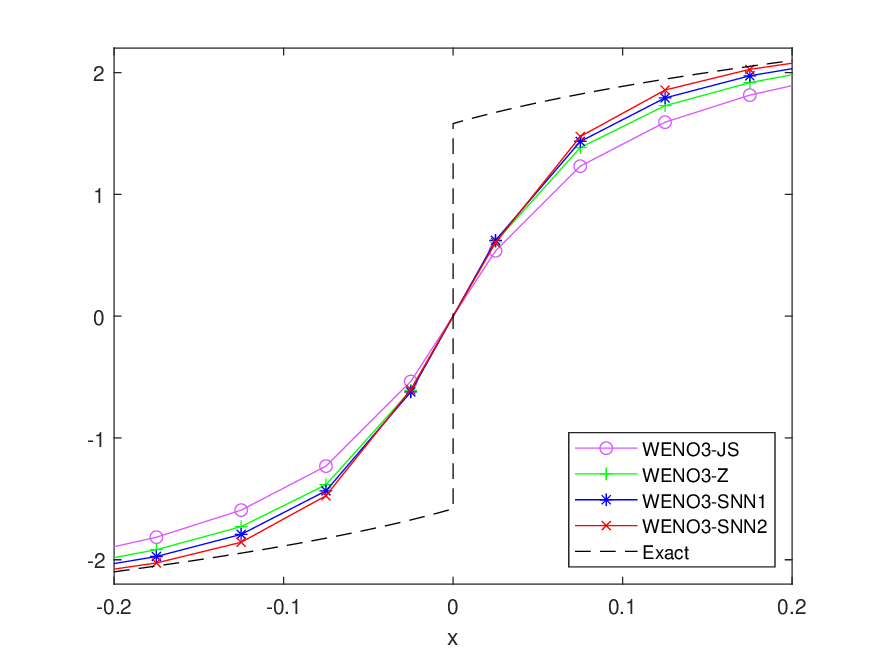}
\caption{Solution profiles for Example \ref{ex:nonconvex} with $u_L=-3$ and $u_R=3$ at $T=0.05$ (left),  close-up view of the solutions in the box (right) approximated by WENO3-JS (purple), WENO3-Z (green), WENO3-SNN1 (blue) and WENO3-SNN2 (red) with $N = 40$. 
The dashed black line is the exact solution.}
\label{fig:nonconvex_IC1}
\end{figure}

\begin{example} \label{ex:nonconvex}
In this experiment, we solve the Riemann problem of \eqref{eq:1d_scalar_hyperbolic} with another nonconvex flux of the form,
$$ f(u) = \frac{1}{4} (u^2-1)(u^2-4). $$
The initial conditions is
$$
   u(x,0) = \left\{ 
             \begin{array}{ll} 
              u_L, & x \leqslant 0, \\ 
              u_R, & x > 0.
             \end{array} 
            \right.
$$
We set $N=40$ grid points for the computational domain $[-1, \, 1]$.

If $u_L = 2$ and $u_R = -2$, then the exact solution consists of two shocks and one rarefaction wave in between. 
We run the simulation up to the final time $T=1$.
Fig. \ref{fig:nonconvex_IC0} shows the numerical results by four WENO schemes at the final time.
The solution computed by WENO3-SNN1 performs the best in the area of the rarefaction wave, while the solution by WENO3-SNN2 shows sharper approximations around the regions of two shocks than the other three WENO schemes.

If $u_L = -3$ and $u_R = 3$, then the exact solution is a stationary shock at $x=0$.
We present the numerical solutions at the final time $T=0.05$, as shown in Fig. \ref{fig:nonconvex_IC1}, where the numerical solution by WENO3-SNN2 are closer to the exact solution than WENO3-JS, WENO3-Z and WENO3-SNN1 around the stationary shock due to its low dissipation.
\end{example}

\subsection{One-dimensional system problems}
For one-dimensional system problem, we consider the Euler equations of gas dynamics
\begin{equation} \label{eq:euler_1d}
 \bfu_t + \bff(\bfu)_x = 0, 
\end{equation}
where the column vector $\bfu$ of the conserved variables and the flux vector $\bff$ in the $x$ direction are
$$
   \bfu = \left[ \rho, \, \rho u, \, E \right]^T, \quad \bff(\bfu) = \left[ \rho u, \, \rho u^2 + P, \, u(E+P) \right]^T,
$$
with $\rho, u$ and $P$ the primitive variables representing density, velocity and pressure, respectively.
The specific kinetic energy $E$ is
$$
   E = \frac{P}{\gamma - 1} + \frac{1}{2} \rho u^2 
$$
with $\gamma = 1.4$ for the ideal gas.

\begin{example} \label{ex:euler_1d_order}
We check the order of accuracy for the one-dimensional Euler equations \eqref{eq:euler_1d} with the initial condition
$$
   (\rho, u, P)(x,0) = \left( 1 + 0.5 \sin(\pi x), \, 1, \, 1 \right),
$$
and the periodic boundary condition.
The exact solution is given by
$$
   (\rho, u, P)(x,t) = \left( 1 + 0.5 \sin \left( \pi (x-t) \right), \, 1, \, 1 \right).
$$
The numerical solution is computed up to the final time $T=2$ with the time step $\Delta t = \cfl \cdot \Delta x$. 

The $L_1$ and $L_{\infty}$ errors versus $N$ with the order of accuracy for the WENO schemes, are displayed in Tables \ref{tab:euler_1d_order_L1} and \ref{tab:euler_1d_order_Linf}, respectively.
Similar to Example \ref{ex:advection_1d_order}, both WENO3-SNNs perform better than WENO3-JS and WENO3-S in terms of accuracy and convergence, though none of the WENO schemes achieves the third-order accuracy. 
\end{example}
\begin{table}[htbp]
\renewcommand{\arraystretch}{1.1}
\scriptsize
\centering
\caption{$L_1$ error and numerical order for Example \ref{ex:euler_1d_order}.}      
\begin{tabular}{clcrlcrlcrlc} 
\hline  
N & \multicolumn{2}{l}{WENO3-JS} & & \multicolumn{2}{l}{WENO3-Z} & & \multicolumn{2}{l}{WENO3-SNN1} & & \multicolumn{2}{l}{WENO3-SNN2} \\ 
    \cline{2-3}                      \cline{5-6}                     \cline{8-9}                        \cline{11-12}                   
  & Error & Order                & & Error & Order               & & Error & Order                  & & Error & Order \\
\hline  
10  & 1.50e-1 & --     & & 1.10e-1 & --     & & 1.05e-1 & --     & & 8.72e-2 & --     \\  
20  & 4.55e-2 & 1.7179 & & 3.67e-2 & 1.5885 & & 3.50e-2 & 1.5805 & & 2.70e-2 & 1.6899 \\  
40  & 1.92e-2 & 1.2447 & & 1.03e-2 & 1.8295 & & 8.67e-3 & 2.0148 & & 6.62e-3 & 2.0290 \\
80  & 4.82e-3 & 1.9929 & & 2.43e-3 & 2.0872 & & 1.96e-3 & 2.1484 & & 1.47e-3 & 2.1766 \\ 
160 & 1.17e-3 & 2.0427 & & 5.33e-4 & 2.1894 & & 3.98e-4 & 2.2956 & & 3.18e-4 & 2.2032 \\
\hline
\end{tabular}
\label{tab:euler_1d_order_L1}
\end{table}

\begin{table}[htbp]
\renewcommand{\arraystretch}{1.1}
\scriptsize
\centering
\caption{$L_{\infty}$ error and numerical order for Example \ref{ex:euler_1d_order}.}      
\begin{tabular}{clcrlcrlcrlc}
\hline 
N & \multicolumn{2}{l}{WENO3-JS} & & \multicolumn{2}{l}{WENO3-Z} & & \multicolumn{2}{l}{WENO3-SNN1} & & \multicolumn{2}{l}{WENO3-SNN2} \\
    \cline{2-3}                      \cline{5-6}                     \cline{8-9}                        \cline{11-12}                   
  & Error & Order                & & Error & Order               & & Error & Order                  & & Error & Order \\
\hline  
10  & 2.65e-1 & --     & & 2.16e-1 & --     & & 2.10e-1 & --     & & 1.83e-1 & --     \\  
20  & 1.05e-1 & 1.3372 & & 7.59e-2 & 1.5054 & & 7.15e-2 & 1.5563 & & 6.12e-2 & 1.5787 \\  
40  & 4.39e-2 & 1.2587 & & 2.97e-2 & 1.3516 & & 2.70e-2 & 1.4039 & & 2.31e-2 & 1.4056 \\
80  & 1.76e-2 & 1.3186 & & 1.12e-2 & 1.4124 & & 9.77e-3 & 1.4662 & & 8.46e-2 & 1.4486 \\ 
160 & 6.83e-3 & 1.3653 & & 4.10e-3 & 1.4470 & & 3.45e-3 & 1.5035 & & 3.05e-3 & 1.4740 \\
\hline
\end{tabular}
\label{tab:euler_1d_order_Linf}
\end{table}

\begin{figure}[htbp]
\centering
\includegraphics[height=0.35\textwidth]{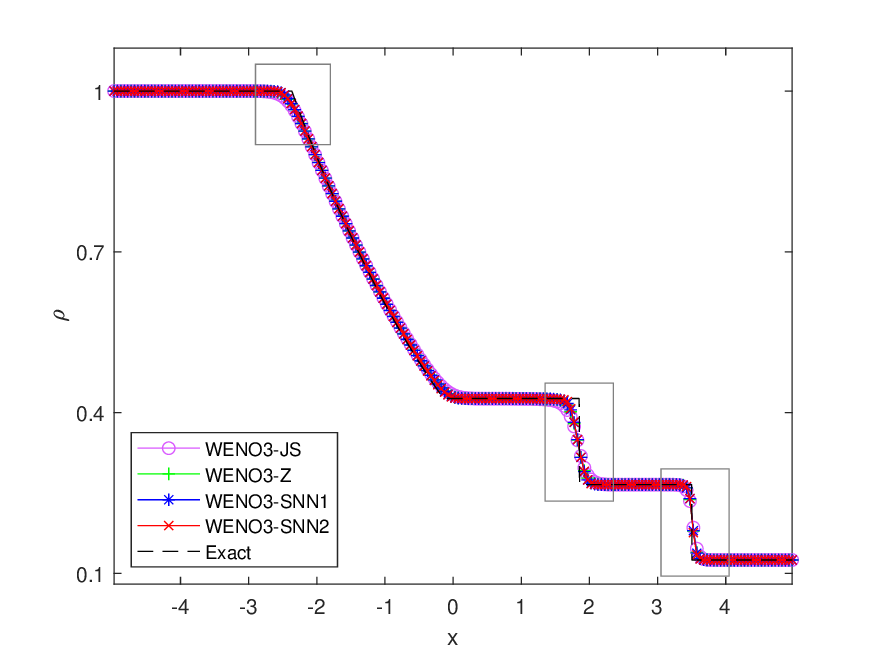} 
\includegraphics[height=0.35\textwidth]{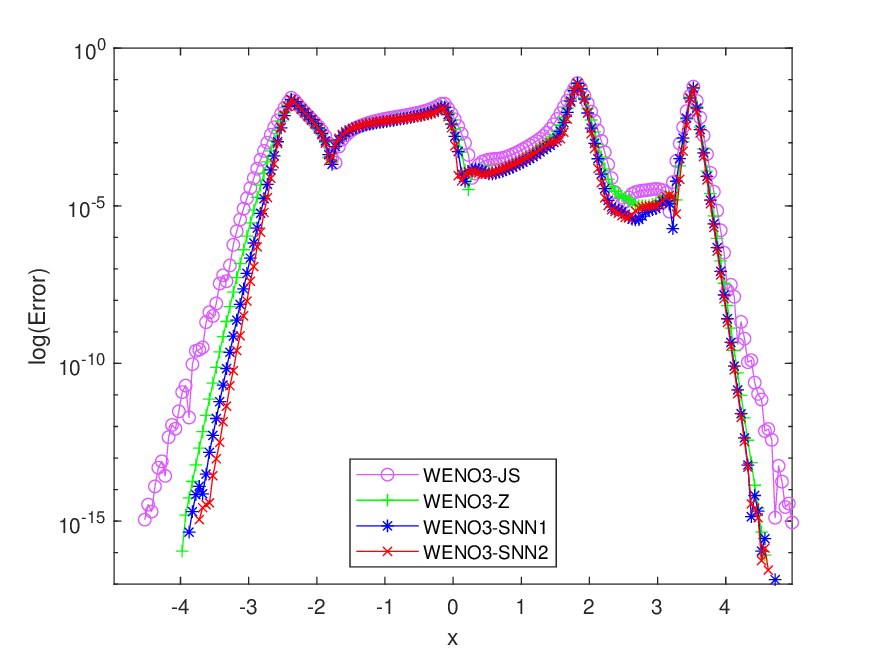} \\
\includegraphics[height=0.23\textwidth]{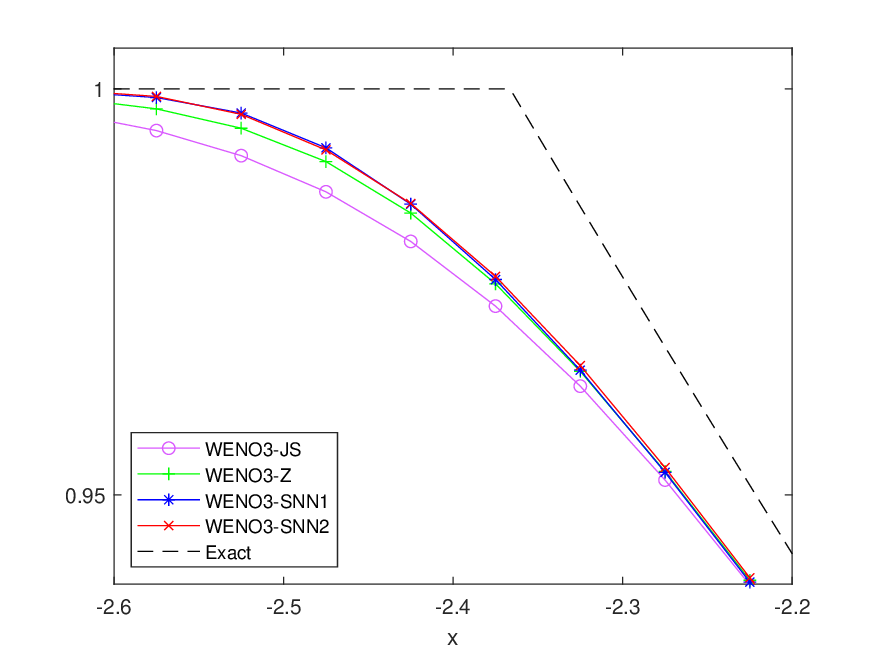}
\includegraphics[height=0.23\textwidth]{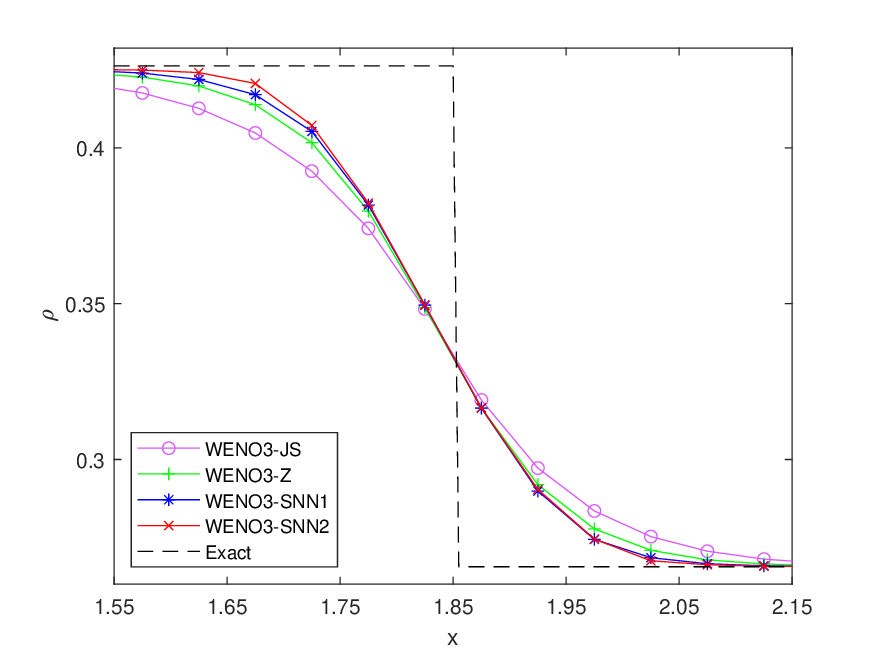}
\includegraphics[height=0.23\textwidth]{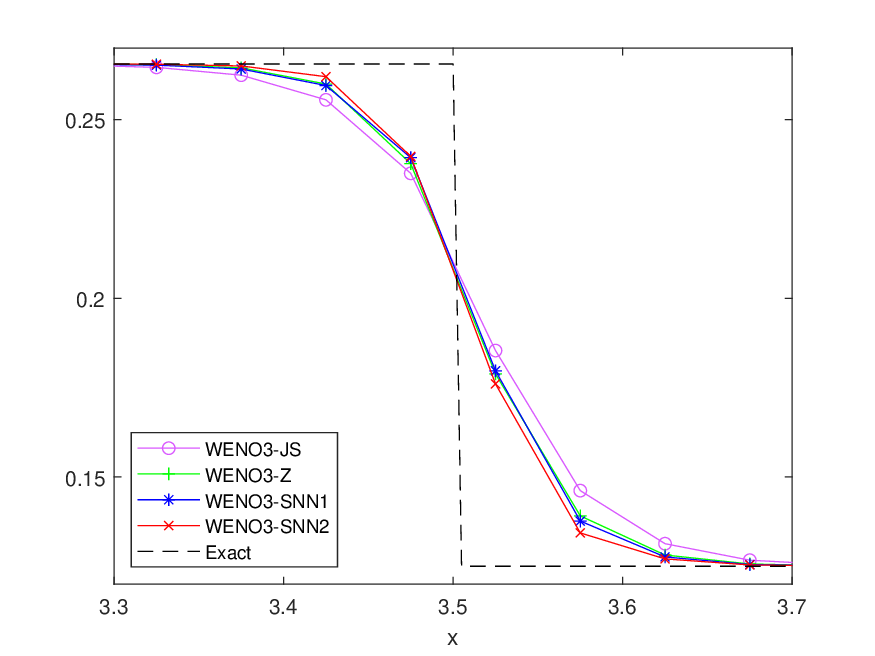}
\caption{Density profiles for the Sod problem \eqref{eq:euler_1d} and \eqref{eq:sod} at $T=2$ (top left), log-scale pointwise error (top right) and close-up view of the solutions in the boxes from left to right (bottom left, bottom middle, bottom right) approximated by WENO3-JS (purple), WENO3-Z (green), WENO3-SNN1 (blue) and WENO3-SNN2 (red) with $N = 200$. 
The dashed black line is the exact solution.}
\label{fig:sod}
\end{figure}

\begin{figure}[htbp]
\centering
\includegraphics[height=0.35\textwidth]{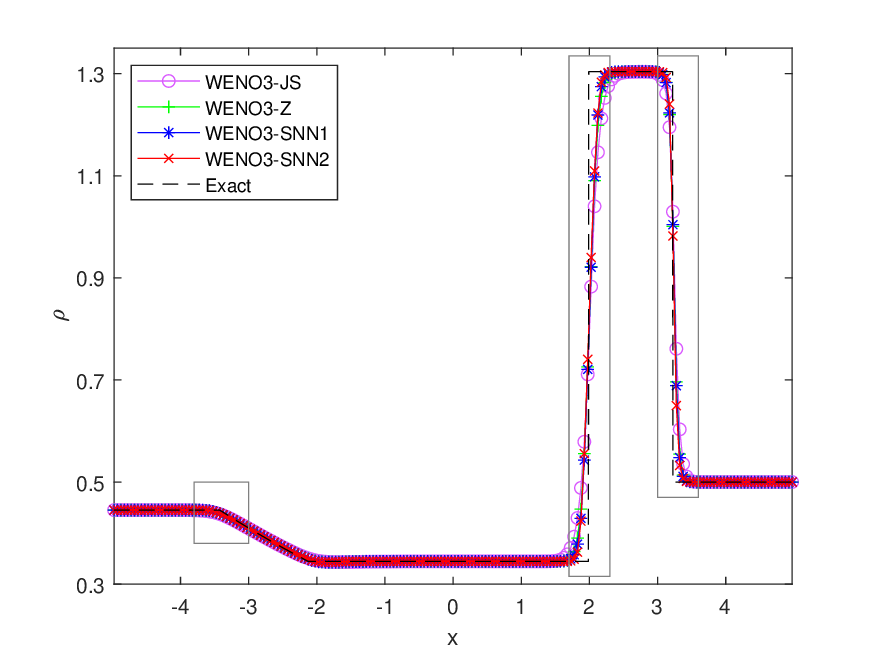}
\includegraphics[height=0.35\textwidth]{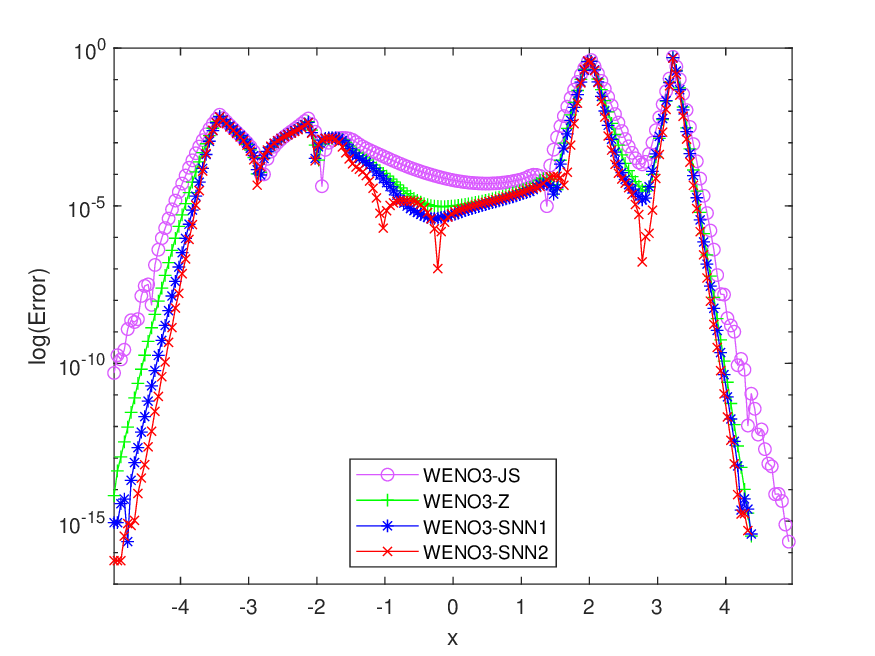}
\includegraphics[height=0.23\textwidth]{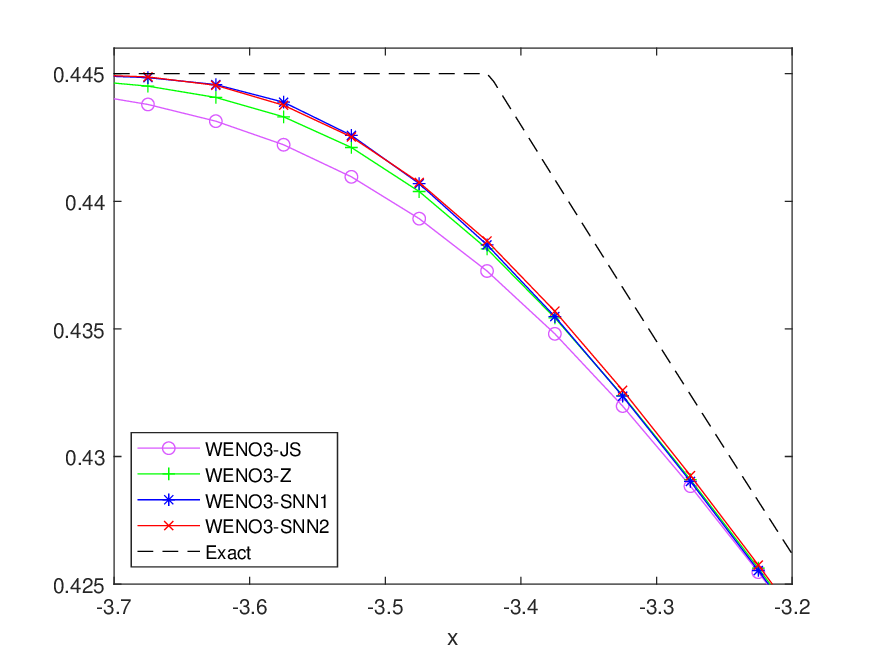}
\includegraphics[height=0.23\textwidth]{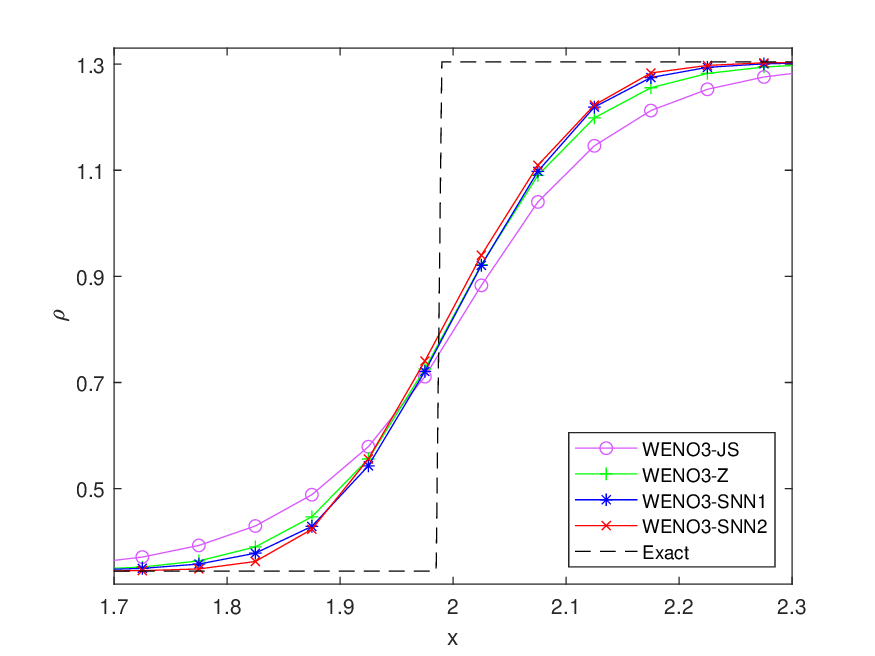}
\includegraphics[height=0.23\textwidth]{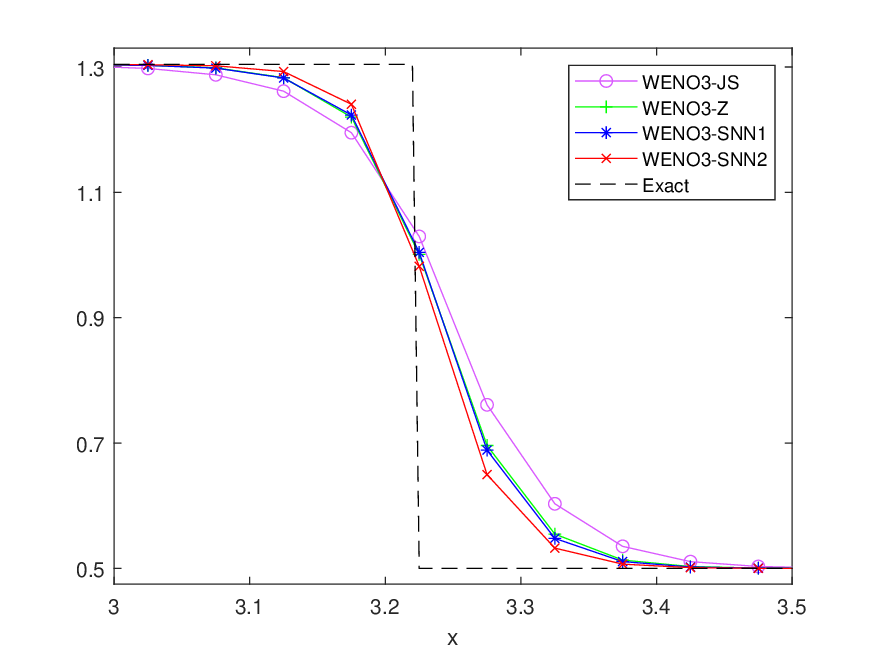}
\caption{Density profiles for the Lax problem \eqref{eq:euler_1d} and \eqref{eq:lax} at $T=1.3$ (top left), log-scale pointwise error (top right) and close-up view of the solutions in the boxes from left to right (bottom left, bottom middle, bottom right) approximated by WENO3-JS (purple), WENO3-Z (green), WENO3-SNN1 (blue) and WENO3-SNN2 (red) with $N = 200$.
The dashed black line is the exact solution.}
\label{fig:lax}
\end{figure}

\begin{figure}[htbp]
\centering
\includegraphics[height=0.35\textwidth]{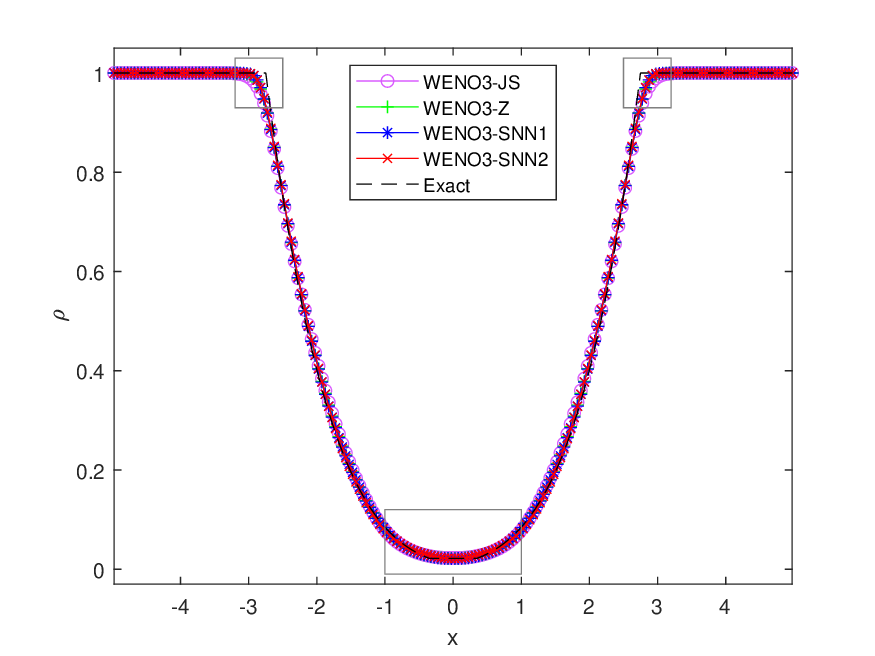}
\includegraphics[height=0.35\textwidth]{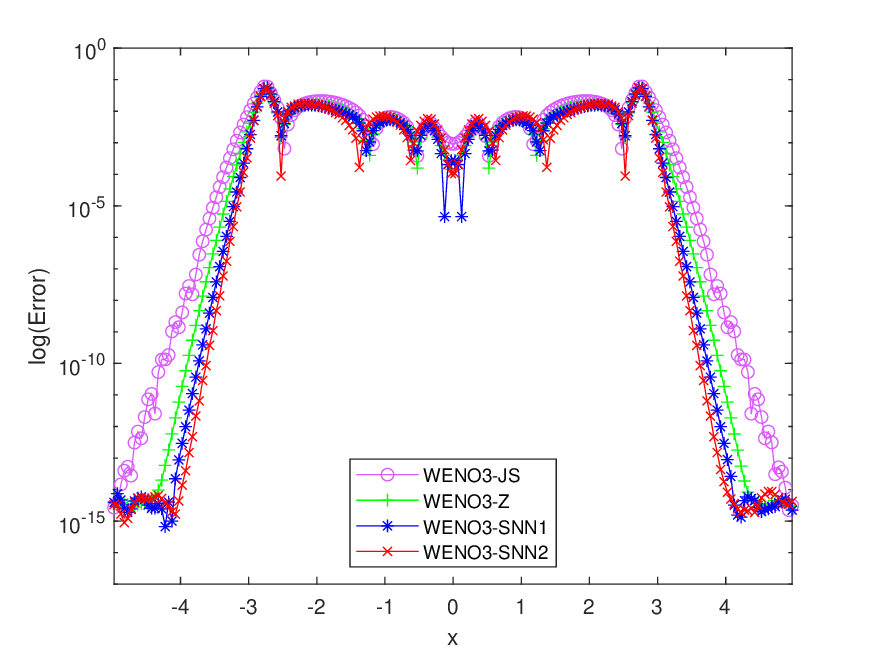}
\includegraphics[height=0.23\textwidth]{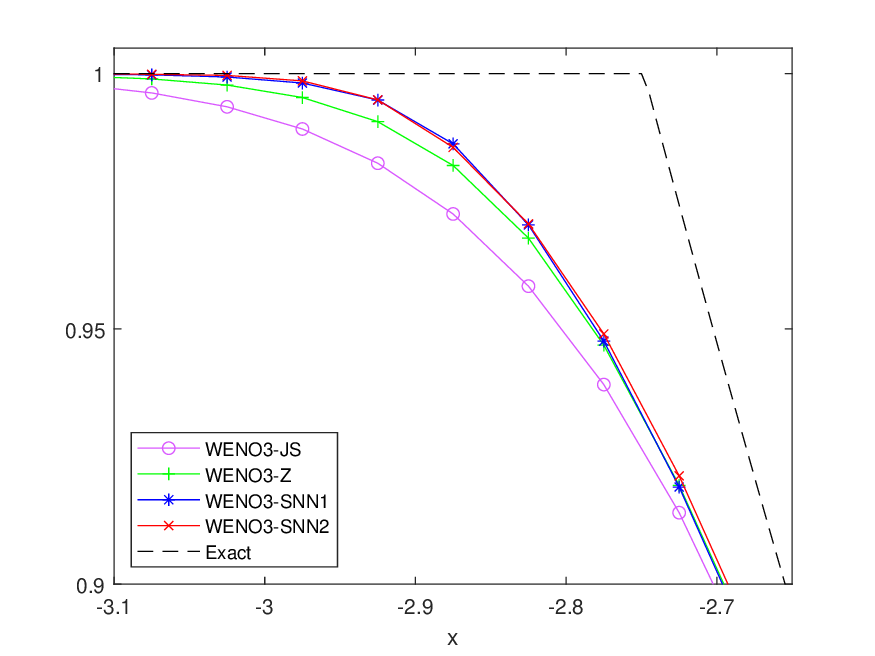}
\includegraphics[height=0.23\textwidth]{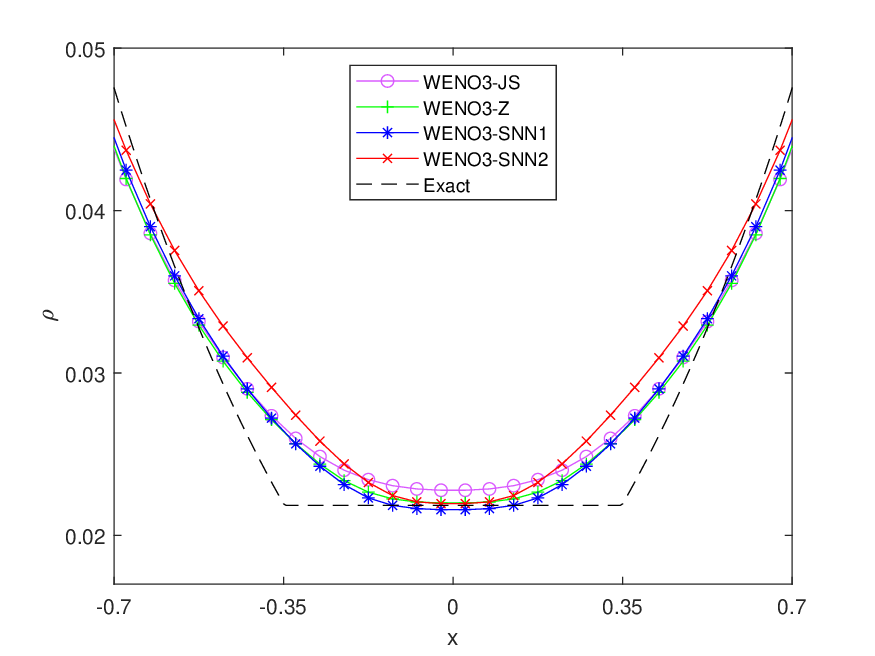}
\includegraphics[height=0.23\textwidth]{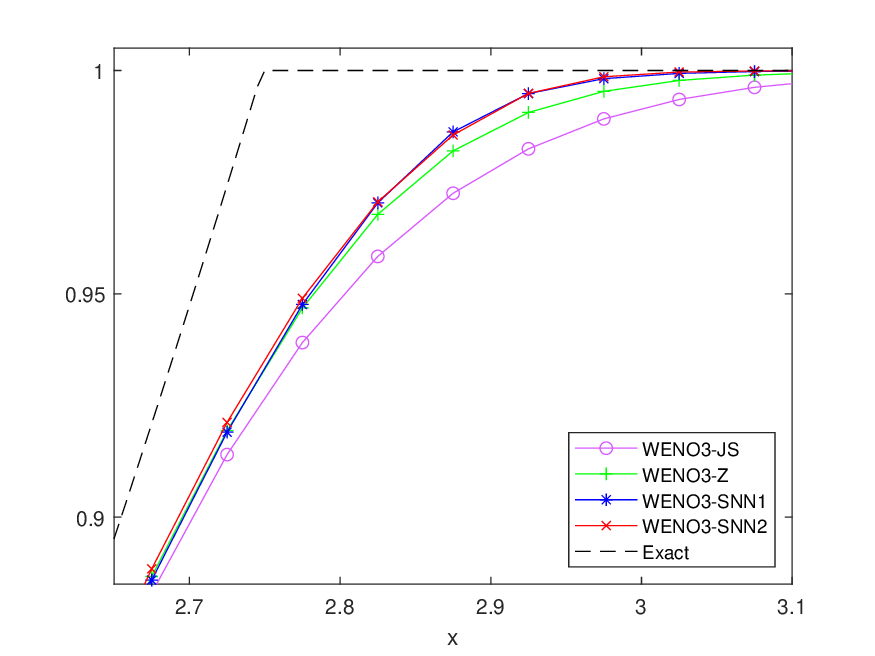}
\caption{Density profiles for the 123 problem \eqref{eq:euler_1d} and \eqref{eq:one23} at $T=1.3$ (top left), log-scale pointwise error (top right) and close-up view of the solutions in the boxes from left to right (bottom left, bottom middle, bottom right) approximated by WENO3-JS (purple), WENO3-Z (green), WENO3-SNN1 (blue) and WENO3-SNN2 (red) with $N = 200$. 
The dashed black line is the exact solution.}
\label{fig:one23}
\end{figure}

\begin{figure}[htbp]
\centering
\includegraphics[height=0.32\textwidth]{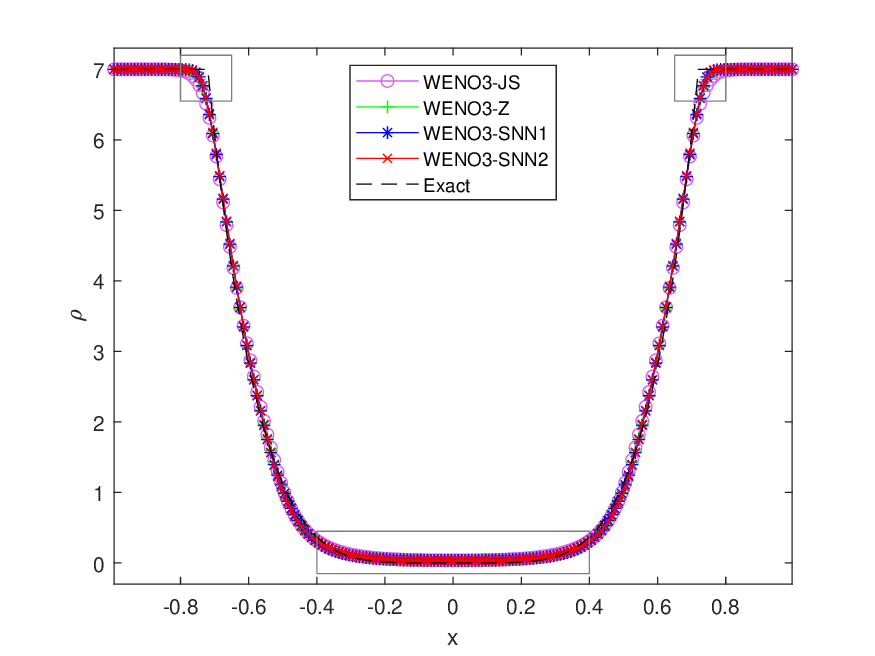}
\includegraphics[height=0.32\textwidth]{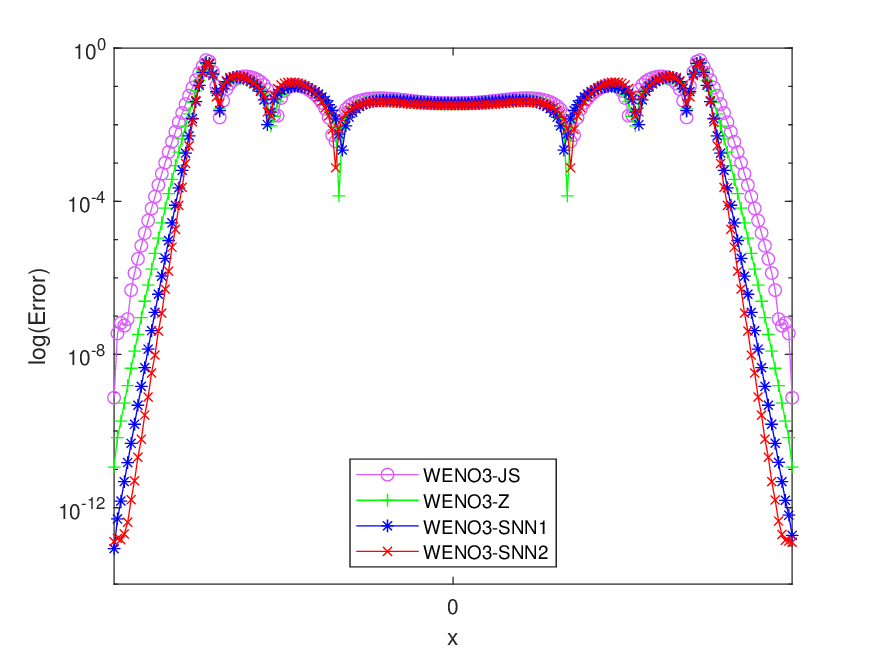}
\includegraphics[height=0.22\textwidth]{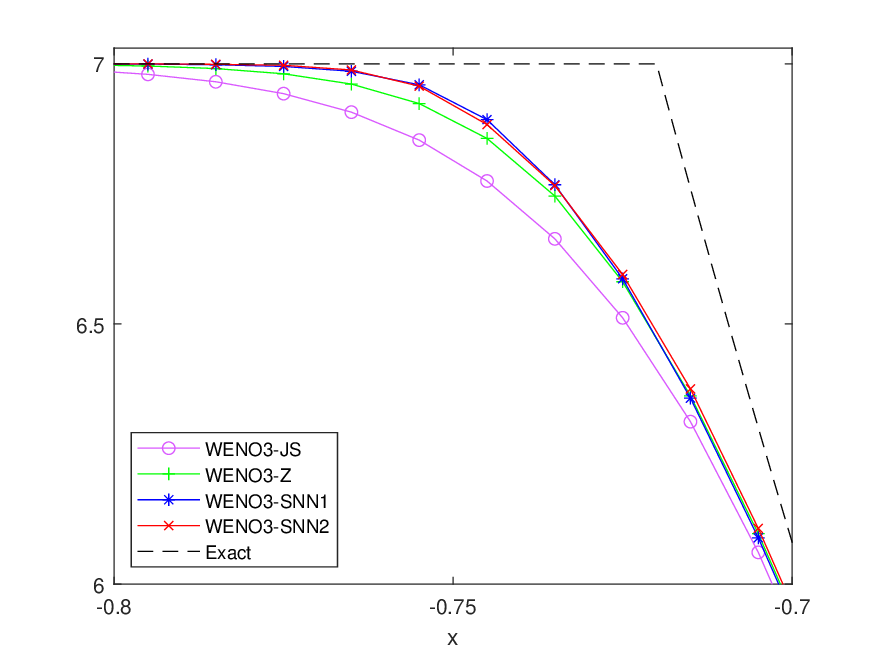}
\includegraphics[height=0.22\textwidth]{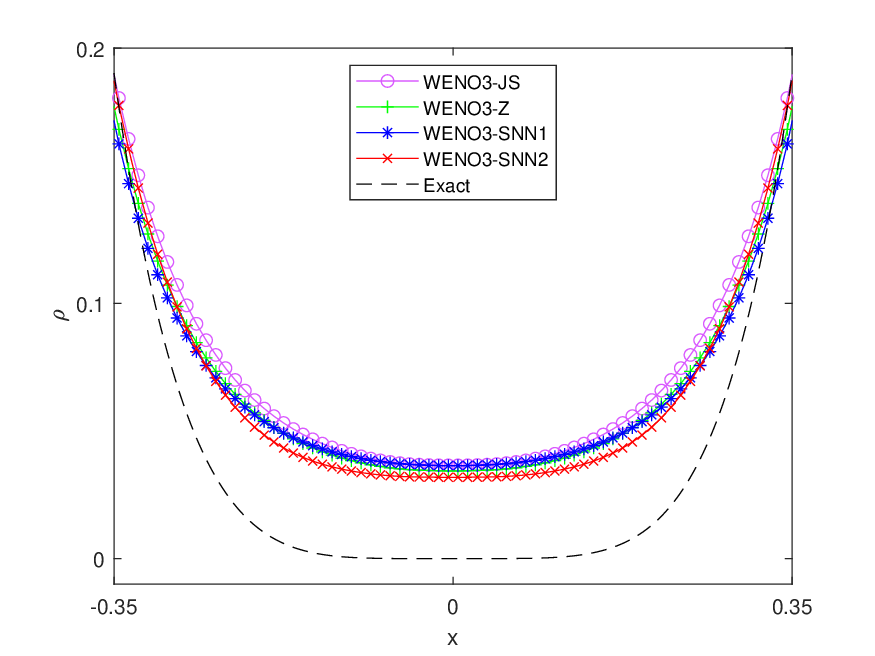}
\includegraphics[height=0.22\textwidth]{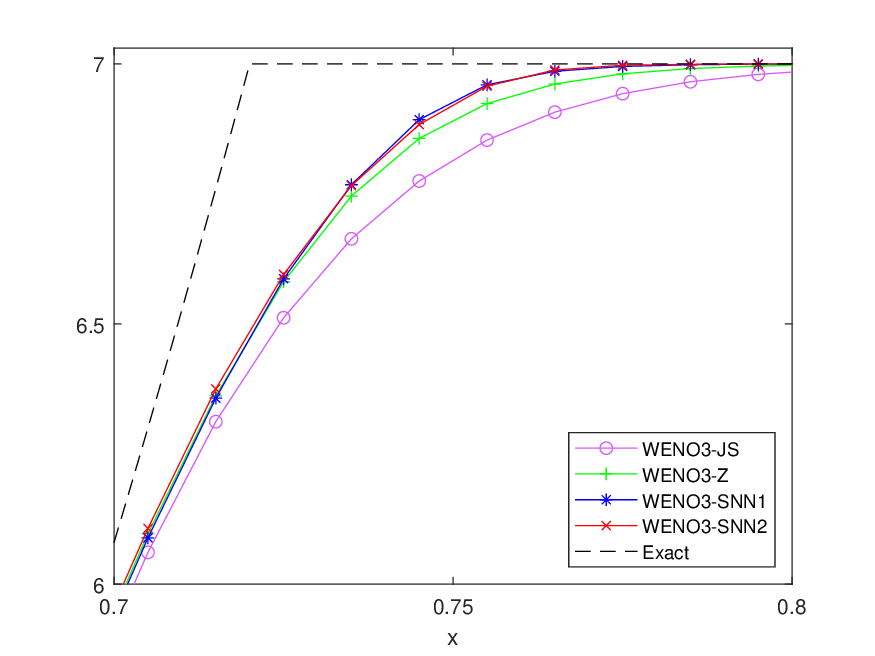}
\caption{Density profiles for the double rarefaction problem \eqref{eq:euler_1d} and \eqref{eq:double_rarefaction} at $T=0.6$ (top left), log-scale pointwise error (top right) and close-up view of the solutions in the boxes from left to right (bottom left, bottom middle, bottom right) approximated by WENO3-JS (purple), WENO3-Z (green), WENO3-SNN1 (blue) and WENO3-SNN2 (red) with $N = 200$. 
The dashed black line is the exact solution.}
\label{fig:double_rarefaction}
\end{figure}

\begin{example} \label{ex:euler_1d}
We start with the Riemann problems for the one-dimensional Euler equations, where the exact solution can be obtained to compare the performance of different WENO schemes.  

The Sod problem is a classic shock tube problem, where the initial condition of the primitive variables
\begin{equation} \label{eq:sod}
 (\rho, u, P ) = \left\{ 
                  \begin{array}{ll} 
                   (1, \, 0, \, 1),       & x \leqslant 0, \\ 
                   (0.125, \, 0, \, 0.1), & x > 0.
                  \end{array} 
                 \right.
\end{equation}
The computational domain is $[-5, \, 5]$ with $N=200$ uniform cells.
Fig. \ref{fig:sod} presents the numerical density by the WENO schemes up to the final time $T=2$ and the pointwise errors on a logarithmic scale.
The density profile approximated by WENO3-SNNs shows the less dissipation around the regions of rarefaction, contact discontinuity and shock wave than WENO3-JS and WENO3-Z.

The Lax problem is alsoa shock tube problem with the initial condition
\begin{equation} \label{eq:lax}
 (\rho, u, P ) = \left\{ 
                  \begin{array}{ll} 
                   (0.445, \, 0.698, \, 3.528), & x \leqslant 0, \\ 
                   (0.5, \, 0, \, 0.571),       & x > 0. 
                  \end{array} 
                 \right. 
\end{equation}
We apply $N=200$ grid points to the computational domain $[-5, \, 5]$.
The final time is $T=1.3$.
We plot the numerical solutions of the density $\rho$ at the final time, along with the log-scale pointwise errors at the grid points, in Fig. \ref{fig:lax}.
The density profile with WENO3-SNN2 is the most accurate around the rarefaction, contact discontinuity and shock regions.

The 123 problem consists of the the Euler equations \eqref{eq:euler_1d} and the initial condition
\begin{equation} \label{eq:one23}
 (\rho, u, P ) = \left\{ 
                  \begin{array}{ll} 
                   (1, \, -2, \, 0.4), & x \leqslant 0, \\ 
                   (1, \, 2, \, 0.4),  & x > 0. 
                  \end{array} 
                 \right. 
\end{equation}
We divide the computational domain $[-5, \, 5]$ into $N=200$ uniform cells.
The simulations of the density $\rho$ at the final time $T=1$ and the pointwise errors on a logarithmic scale are plotted in Fig. \ref{fig:one23}.
The results with WENO3-SNNs are more accurate than those with WENO3-JS and WENO3-Z near the two strong rarefactions. 
In the region of the trivial stationary contact discontinuity, WENO3-SNN1 gives the most accurate density profile.

The double rarefaction problem is the Euler equations \eqref{eq:euler_1d} with the initial condition
\begin{equation} \label{eq:double_rarefaction}
 (\rho, u, P ) = \left\{
                  \begin{array}{ll} 
                   (7, \, -1, \, 0.2), & x \leqslant 0, \\ 
                   (7, \, 1, \, 0.2),  & x > 0.
                  \end{array} 
                 \right.
\end{equation}
The computational domain $[-1, \, 1]$ is discretized with $N=200$ grid points.
Fig. \ref{fig:double_rarefaction} shows the approximations of the density $\rho$ at the final time $T=0.6$ and the log-scale pointwise errors.
We can see that in the regions of the two rarefactions, the numerical solutions of density simulated by WENO3-SNNs are more accurate than those by WENO3-JS and WENO3-Z.
The exact solution contains vacuum shown in the bottom middle figure of Fig. \ref{fig:double_rarefaction}. 
Despite none of the WENO schemes capturing the vacuum exactly, both WENO3-SNNs approximate the low density better than WENO3-JS and WENO3-Z in terms of accuracy.
\end{example}

\begin{figure}[htbp]
\centering
\includegraphics[height=0.32\textwidth]{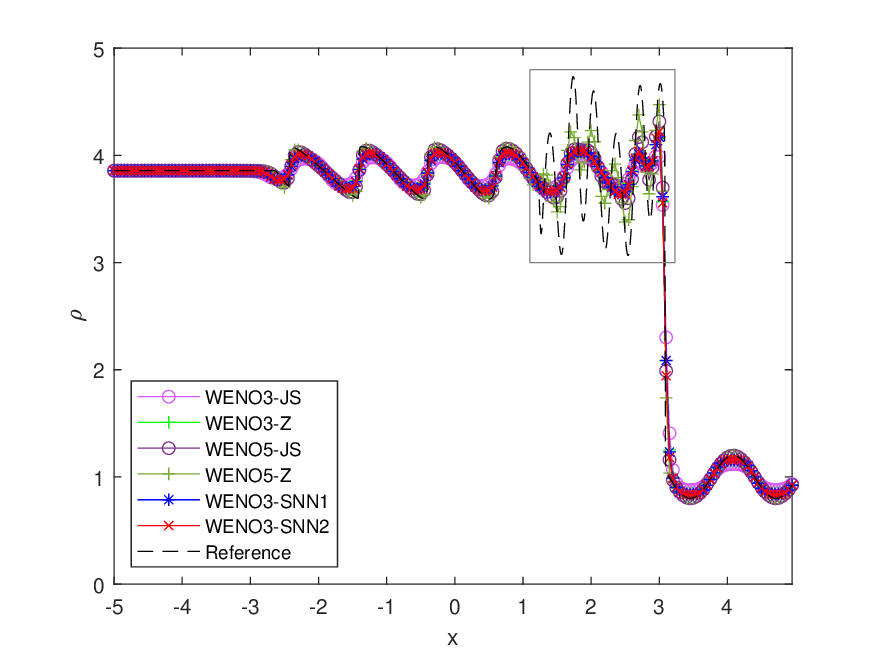}
\includegraphics[height=0.32\textwidth]{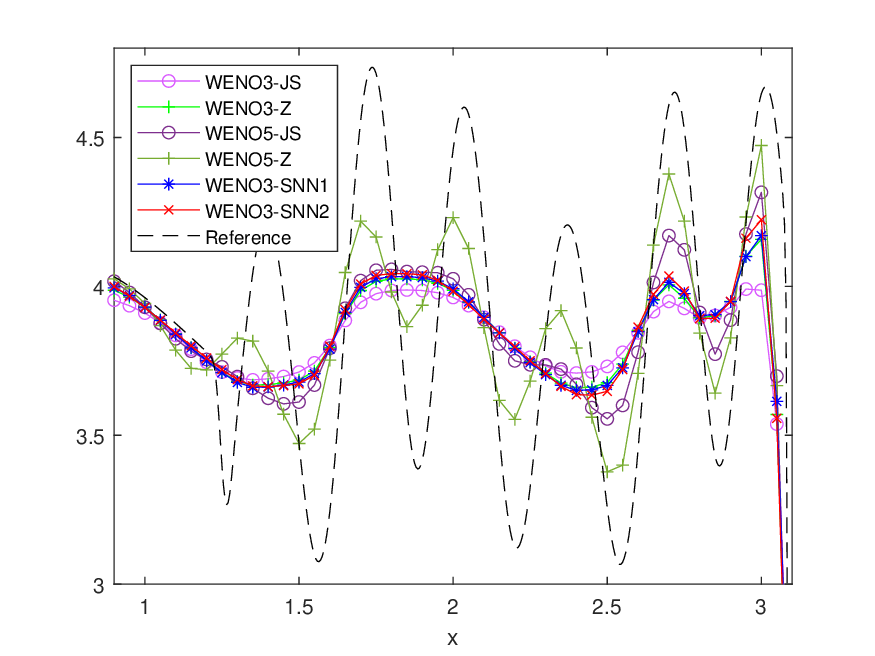}
\caption{Density profiles for Example \ref{ex:shock_entropy_wave} with $k=5$ at $T=2$ (left) and close-up view of the solutions in the box (right) approximated by WENO3-JS (purple), WENO3-Z (green), WENO5-JS (dark purple), WENO5-Z (dark green), WENO3-SNN1 (blue) and WENO3-SNN2 (red) with $N = 200$. 
The dashed black line is generated by fifth-order WENO5-M with $N = 2000$.}
\label{fig:shock_entropy_wave_k5}
\end{figure}

\begin{figure}[htbp]
\centering
\includegraphics[height=0.32\textwidth]{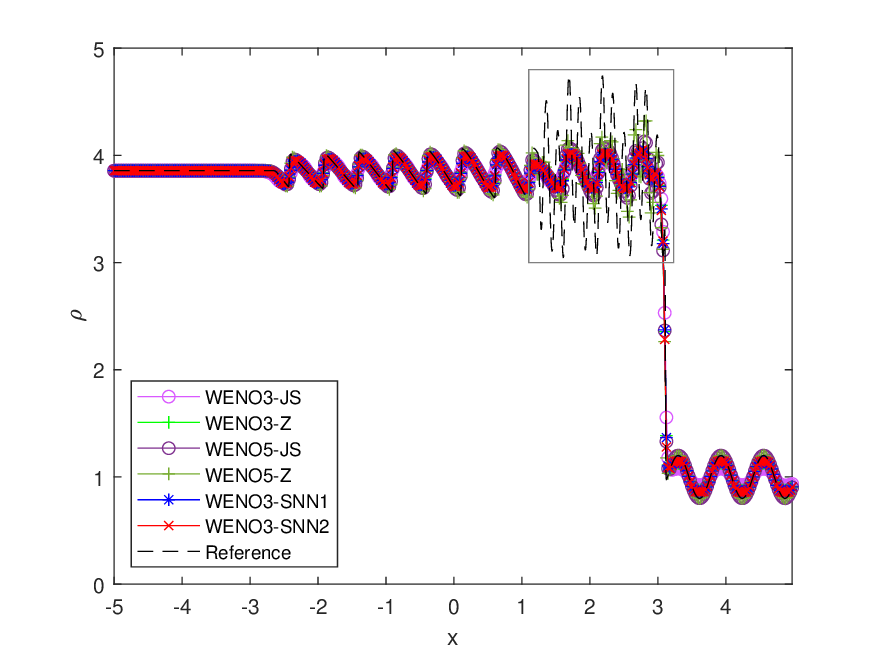}
\includegraphics[height=0.32\textwidth]{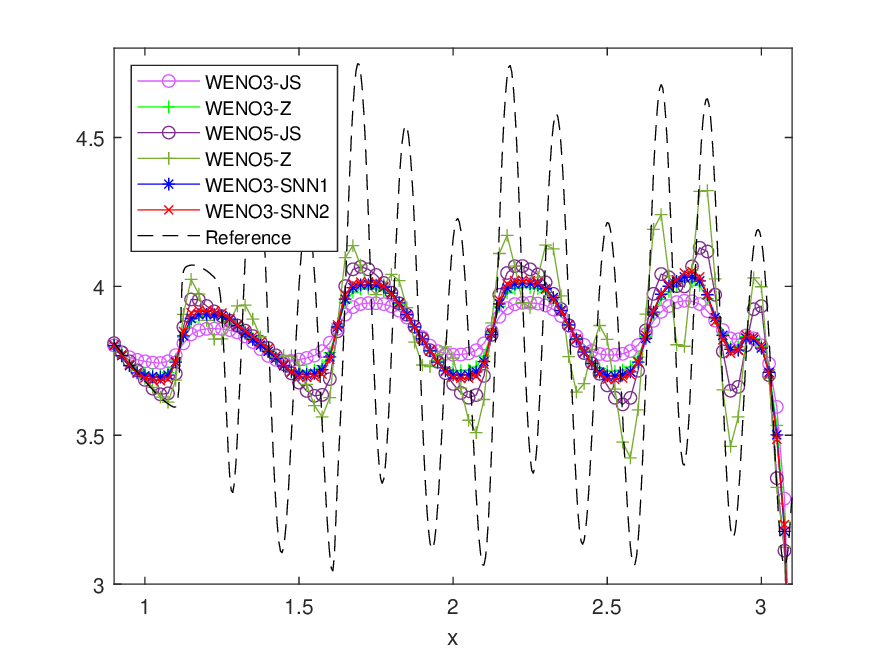}
\caption{Density profiles for Example \ref{ex:shock_entropy_wave} with $k=10$ at $T = 2$ (left), close-up view of the solutions in the box (right) computed by WENO3-JS (purple), WENO3-Z (green), WENO5-JS (dark purple), WENO5-Z (dark green), WENO3-SNN1 (blue) and WENO3-SNN2 (red) with $N = 400$.
The dashed black line is generated by fifth-order WENO5-M with $N = 2000$.}
\label{fig:shock_entropy_wave_k10}
\end{figure}

\begin{example} \label{ex:shock_entropy_wave}
The shock entropy wave interaction problem \cite{ShuOsherI} contains a right moving Mach 3 shock and an entropy wave in density, of which the initial condition is given by
$$
   (\rho, u, P ) = \left\{ 
                    \begin{array}{ll} 
                     (3.857143, \, 2.629369, \, 10.333333), & x < -4, \\ 
                     (1 + 0.2 \sin(kx), \, 0, \, 1),        & x \geqslant -4,
                    \end{array} 
                   \right. 
$$
with $k$ the wave number of the entropy wave.

For $k=5$, we take a uniform grid with $N=200$ cells on the computational domain $[-5, \, 5]$. 
The numerical solution, computed by fifth-order WENO5-M \cite{Henrick} with a high resolution of $N = 2000$ grid points, is used as the reference solution. 
The numerical solutions of density at $T=2$ are displayed in Fig. \ref{fig:shock_entropy_wave_k5}, where the solutions approximated by fifth-order WENO5-JS \cite{Jiang} and WENO5-Z \cite{Borges} are added for comparison.

For $k=10$, the computational domain $[-5, \, 5]$ is divided into $N=400$ uniform cells. 
We compute the numerical solution by fifth-order WENO5-M with $N = 2000$ grid points as the reference solution.
Fig. \ref{fig:shock_entropy_wave_k10} shows the approximate density profiles by WENO3-JS, WENO3-Z, WENO5-JS, WENO5-Z and WENO3-SNNs at $T=2$.

We observe the improved performance of WENO3-SNNs in capturing the fine structure of the density profile over WENO3-JS and WENO3-Z. 
From Fig. \ref{fig:shock_entropy_wave_k5}, the solution of WENO3-SNN2 is even comparable to the one of WENO5-JS in some regions.
\end{example}

\begin{figure}[htbp]
\centering
\includegraphics[height=0.35\textwidth]{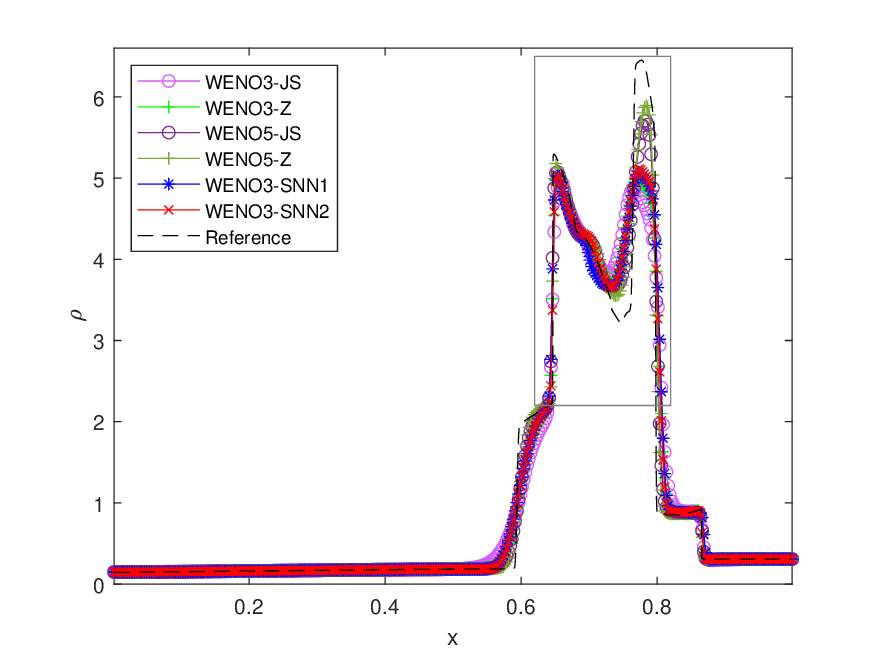}
\includegraphics[height=0.35\textwidth]{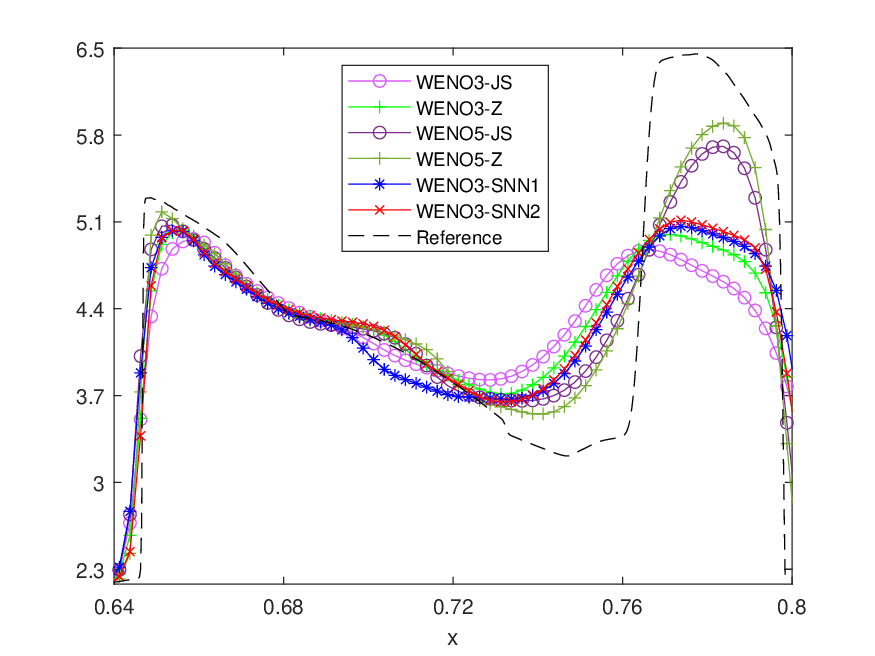}
\caption{Density profiles for Example \ref{ex:interacting_blastwave} at $T=0.038$ (left), close-up views of the solutions in the box (right) computed by WENO3-JS (purple), WENO3-Z (green), WENO5-JS (dark purple), WENO5-Z (dark green), WENO3-SNN1 (blue) and WENO3-SNN2 (red) with $N = 400$.
The dashed black lines are generated by fifth-order WENO5-M with $N = 4000$.}
\label{fig:interacting_blastwave}
\end{figure}

\begin{example} \label{ex:interacting_blastwave} 
The blastwaves interaction problem \cite{Woodward} depicts the evolution of two blast waves developing and colliding, later producing a new contact discontinuity.
The initial condition is given by
$$
   (\rho, u, P ) = \left\{ 
                    \begin{array}{ll} 
                     (1,~0,~1000), & 0 \leqslant x < 0.1, \\ 
                     (1,~0,~0.01), & 0.1 \leqslant x < 0.9, \\
                     (1,~0,~100),  & 0.9 \leqslant x \geqslant 1.
                    \end{array} 
                   \right. 
$$
The reflective boundary condition is applied to both ends.
The computational domain is $[0, \, 1]$ with $N = 400$ uniform cells and the final time is $T = 0.038$.
Fig. \ref{fig:interacting_blastwave} plots the numerical solutions of the density $\rho$ by WENO3-JS, WENO3-Z, WENO5-JS, WENO5-Z and WENO3-SNNs.
The dashed line is the reference solution computed by fifth-order WENO5-M with $N=4000$ grid points.
It can be seen that WENO3-SNNs have better resolution than WENO3-JS and WENO3-Z because of its reduced dissipation around discontinuities, but WENO5-JS and WENO5-Z provide better performance than the third-order WENO schemes due to their higher order.
\end{example}

\subsection{Two-dimensional scalar problems}
\begin{figure}[htbp]
\centering
\includegraphics[height=0.3\textwidth]{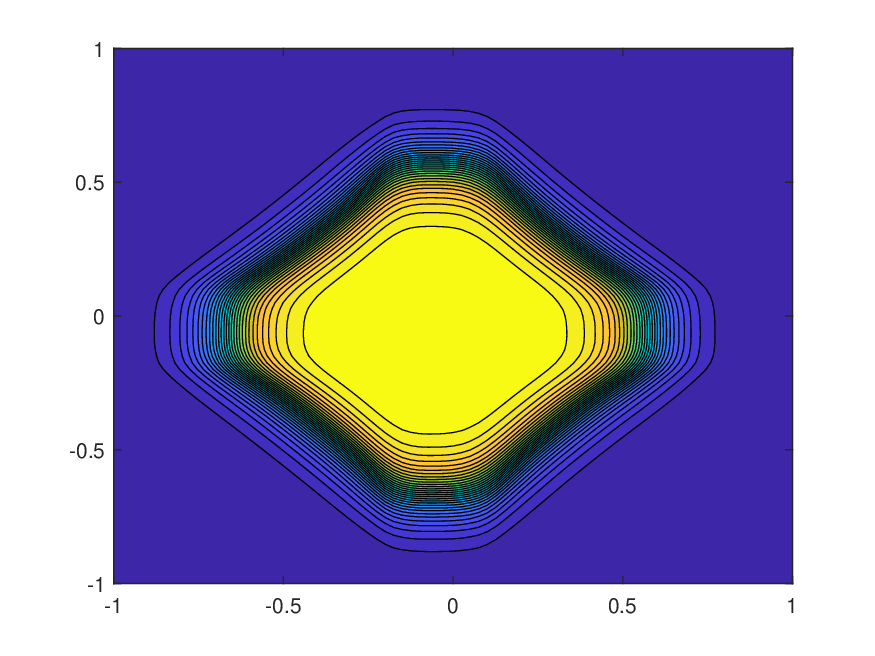}
\includegraphics[height=0.3\textwidth]{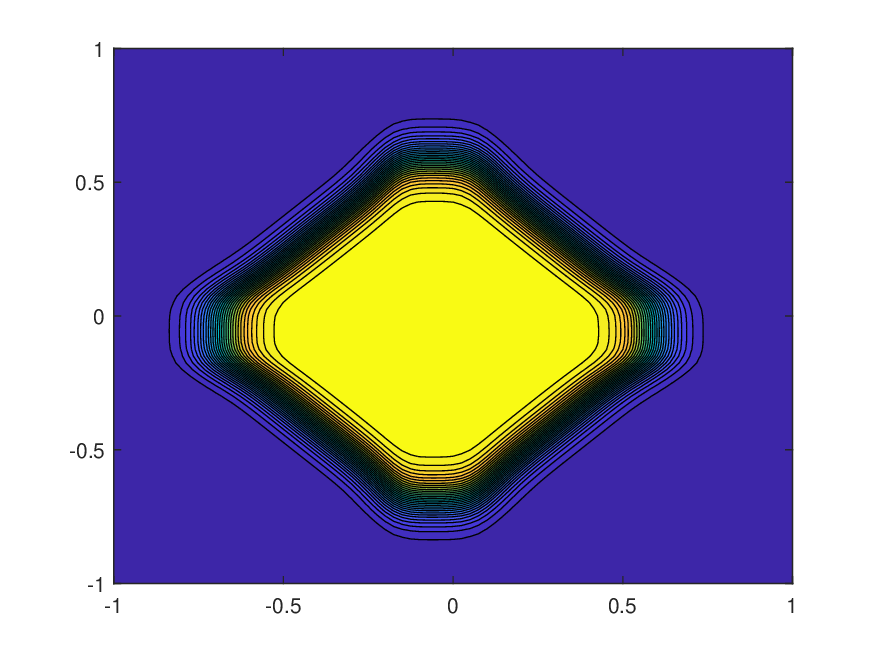}
\includegraphics[height=0.3\textwidth]{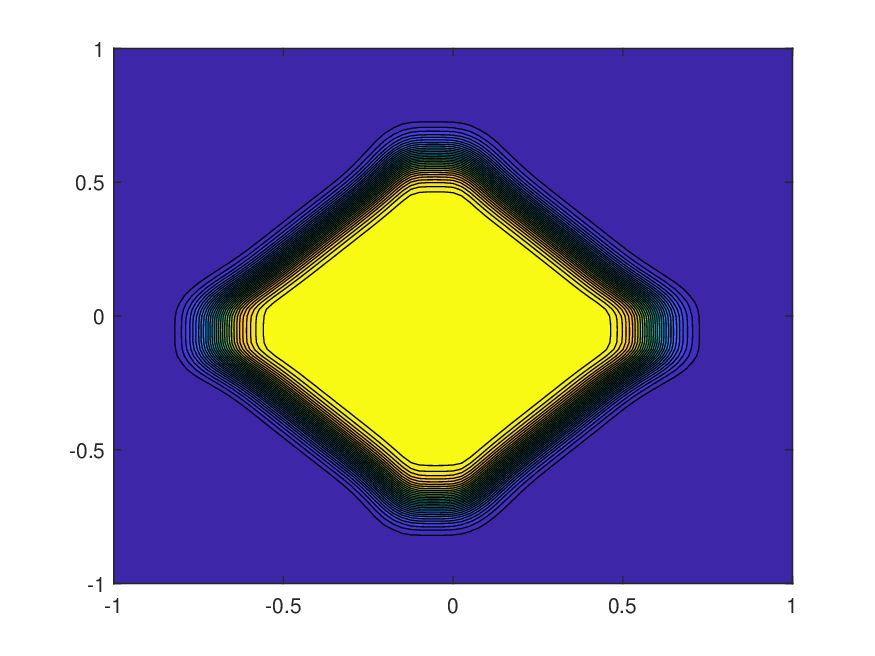}
\includegraphics[height=0.3\textwidth]{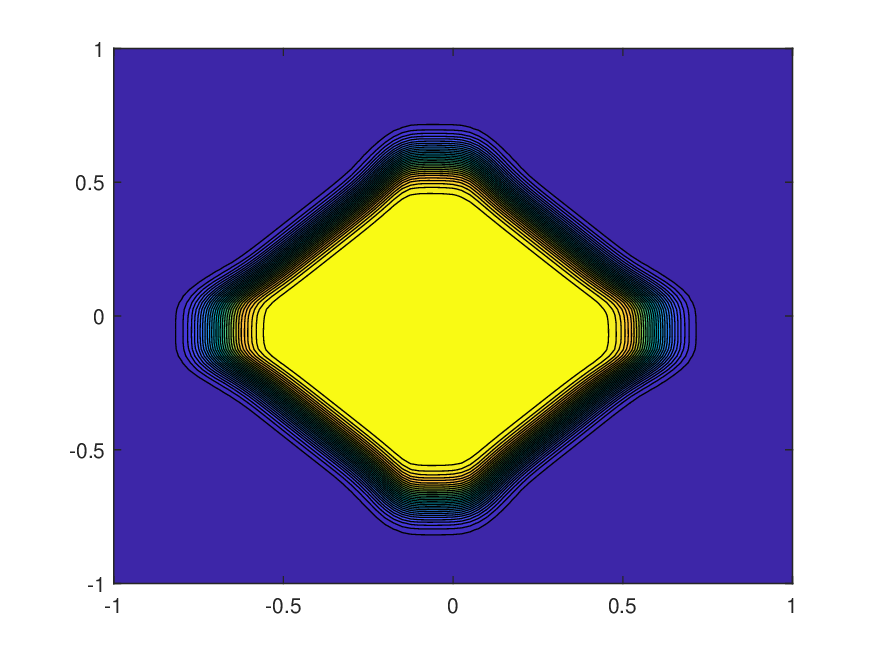}
\caption{Solutions in the filled contour plot for Example \ref{ex:advection_2d} at $T=4$ by WENO3-JS (top left), WENO3-Z (top right), WENO3-SNN1 (bottom left) and WENO3-SNN2 (bottom right) with $N_x = N_y = 80$.
Each contour plot displays contours at $30$ levels of $u$.}
\label{fig:advection_2d}
\end{figure}

\begin{example} \label{ex:advection_2d}
Consider the two-dimensional linear advection equation,
$$ u_t + u_x + u_y = 0, \quad -1 \leqslant x, y \leqslant 1, $$
with the initial condition
$$
   u(x,y,0) = \left\{
               \begin{array}{ll} 
                1, & \text{if } (x,y) \in S, \\
                0, & \text{otherwise},
               \end{array} 
              \right.
$$
with $S = \{ (x,y): \, |x \pm y| < 1/\sqrt{2} \}$ a unit square centered at the origin, and the periodic boundary condition.
We divide the computational domain $[-1, \, 1] \times [-1, \, 1]$ into $N_x \times N_y = 80 \times 80$ uniform cells and run the WENO schemes up to the final time $T=4$.
Fig. \ref{fig:advection_2d} shows the numerical solutions by those four WENO schemes at the final time, and Table \ref{tab:advection_2d} displays the $L_1, \, L_2$ and $L_\infty$ errors.
According to Table \ref{tab:advection_2d}, WENO3-SNNs provide better accuracy than WENO3-JS and WENO3-Z, where WENO3-SNN1 gives the smallest $L_1$ and $L_2$ errors while WENO3-SNN2 leads to the least $L_{\infty}$ error.
\end{example}

\begin{table}[htbp]
\renewcommand{\arraystretch}{1.1}
\scriptsize
\centering
\caption{$L_1, \, L_2, \, L_\infty$ errors for Example \ref{ex:advection_2d}.}      
\begin{tabular}{ccccc}
\hline 
Error & WENO3-JS & WENO3-Z & WENO3-SNN1 & WENO3-SNN2 \\
\hline 
$L_1$      & 0.068205 & 0.050340 & 0.045149 & 0.045808 \\
$L_2$      & 0.261161 & 0.224367 & 0.212482 & 0.220650 \\
$L_\infty$ & 0.773255 & 0.755226 & 0.745258 & 0.727796 \\  
\hline
\end{tabular}
\label{tab:advection_2d}
\end{table}

\begin{figure}[htbp]
\centering
\includegraphics[height=0.28\textwidth]{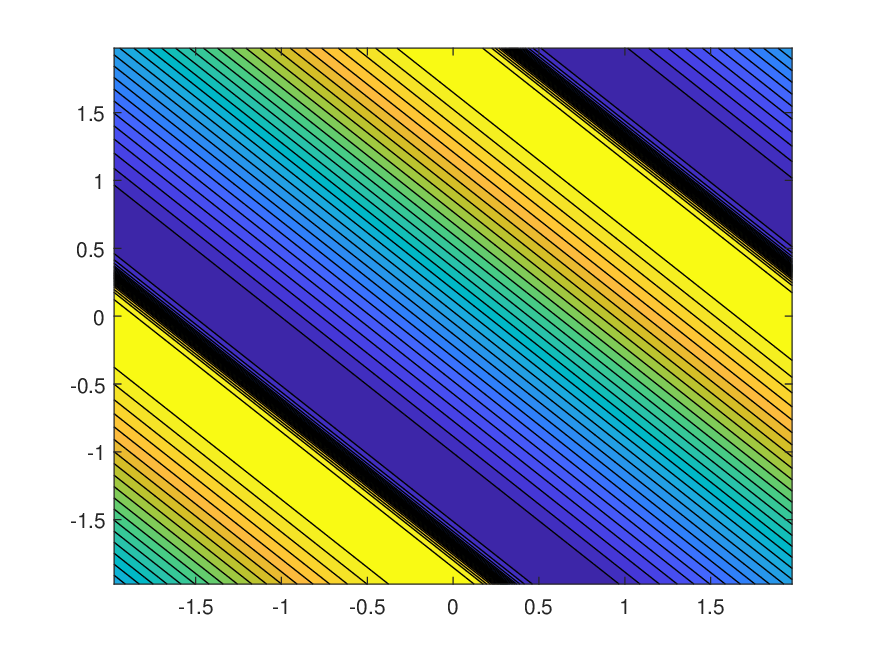}
\includegraphics[height=0.28\textwidth]{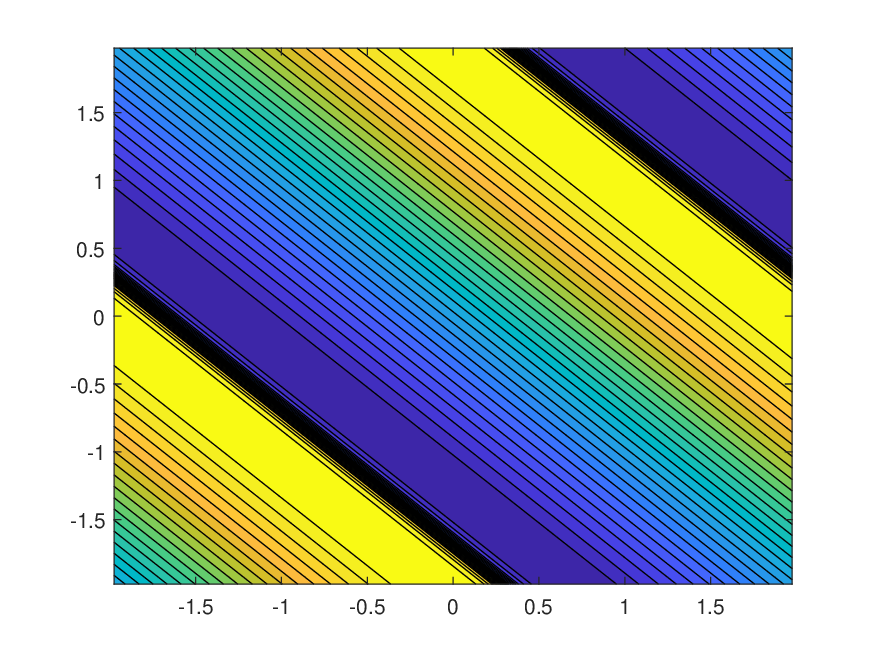}
\includegraphics[height=0.28\textwidth]{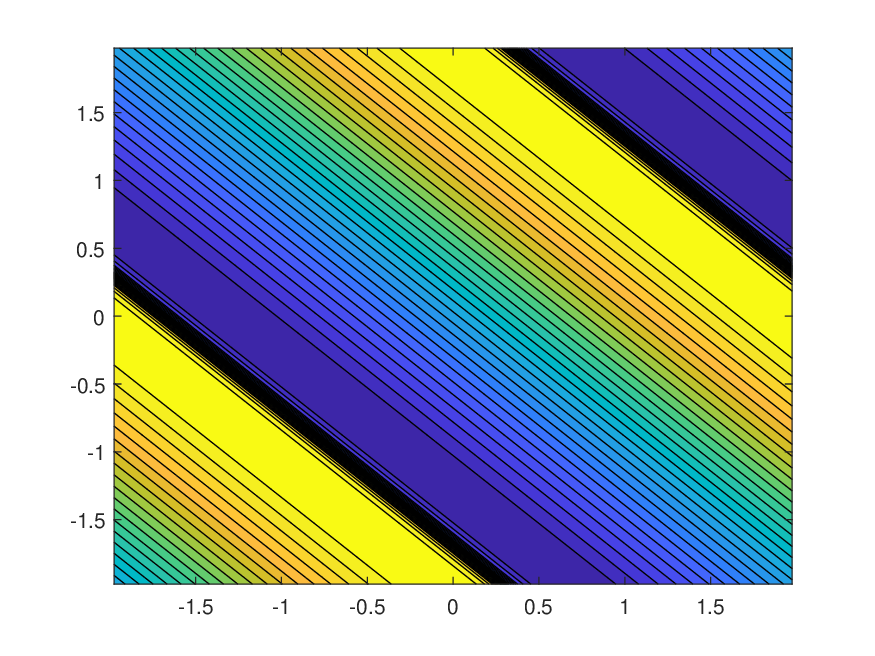}
\includegraphics[height=0.28\textwidth]{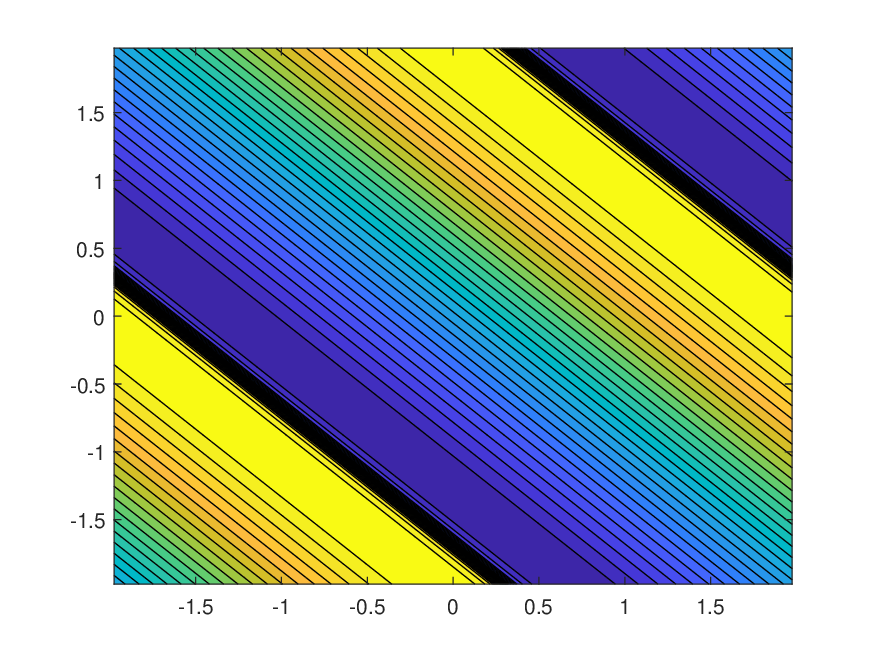}
\caption{Solutions in the filled contour plot for Example \ref{ex:burgers_2d} at $T=2/\pi$ by WENO3-JS (top left), WENO3-Z (top right), WENO3-SNN1 (bottom left) and WENO3-SNN2 (bottom right) with $N_x = N_y = 80$.
Each contour plot displays contours at $30$ levels of $u$.}
\label{fig:burgers_2d}
\end{figure}

\begin{example} \label{ex:burgers_2d}
The two-dimensional Burgers' equation has the form
$$ u_t + \left( \frac{1}{2}u^2 \right)_x+ \left( \frac{1}{2}u^2 \right)_y = 0. $$
The initial condition is
$$
   u(x,y,0) = \frac{1}{4} + \frac{1}{2} \sin \left( \pi \frac{x+y}{2} \right).
$$ 
The exact solution is smooth up to the final time $T = 2/\pi$.
The computational domain is $[-2,\, 2] \times [-2,\, 2]$ with $N_x \times N_y = 80 \times 80$ grid points.
We observe from Fig. \ref{fig:burgers_2d} that all WENO schemes generate comparable numerical profiles, which implies that our WENO3-SNN schemes work well in shock-capturing.
Besides, Table \ref{tab:burgers_2d} shows that WENO3-SNN1 gives the best accuracy in terms of $L_1, \, L_2$ and $L_\infty$ errors, but WENO3-SNN2 is slightly worse than WENO3-Z due to its low dissipation probably.
\end{example}

\begin{table}[htbp]
\renewcommand{\arraystretch}{1.1}
\scriptsize
\centering
\caption{$L_1, \, L_2, \, L_\infty$ errors for Example \ref{ex:burgers_2d}.}      
\begin{tabular}{ccccc}
\hline 
Error & WENO3-JS & WENO3-Z & WENO3-SNN1 & WENO3-SNN2 \\
\hline
$L_1$      & 0.004491 & 0.003372 & 0.003206 & 0.003423 \\
$L_2$      & 0.067012 & 0.058067 & 0.056618 & 0.058507 \\
$L_\infty$ & 0.120357 & 0.121121 & 0.118633 & 0.134010 \\  
\hline
\end{tabular}
\label{tab:burgers_2d}
\end{table}

\subsection{Two-dimensional system problems}
The two-dimensional Euler equations of gas dynamics are of the form
\begin{equation} \label{eq:euler_2d}
 \bfu_t + \bff(\bfu)_x + \bfg(\bfu)_y = 0, 
\end{equation}
where the conserved vector $\bfu$ and the flux functions $\bff, \, \bfg$ in the $x, \, y$ directions, respectively, are 
\begin{align*}
      \bfu  &= \left[ \rho, \, \rho u, \, \rho v, \, E \right]^T, \\
 \bff(\bfu) &= \left[ \rho u, \, \rho u^2+P, \, \rho u v, \, u(E+P) \right]^T, \\ 
 \bfg(\bfu) &= \left[ \rho v, \, \rho u v, \, \rho v^2+P, \, v(E+P) \right]^T.
\end{align*}
As in one-dimensional case, $\rho$ is the density and $P$ is the pressure. 
The primitive variables $u$ and $v$ denote $x$- and $y$-component velocity, respectively.
The specific kinetic energy $E$ is defined as
$$
   E = \frac{P}{\gamma - 1} + \frac{1}{2} \rho ( u^2 + v^2 ) 
$$
with $\gamma = 1.4$ for the ideal gas. 

\begin{figure}[htbp]
\centering
\includegraphics[height=0.28\textwidth]{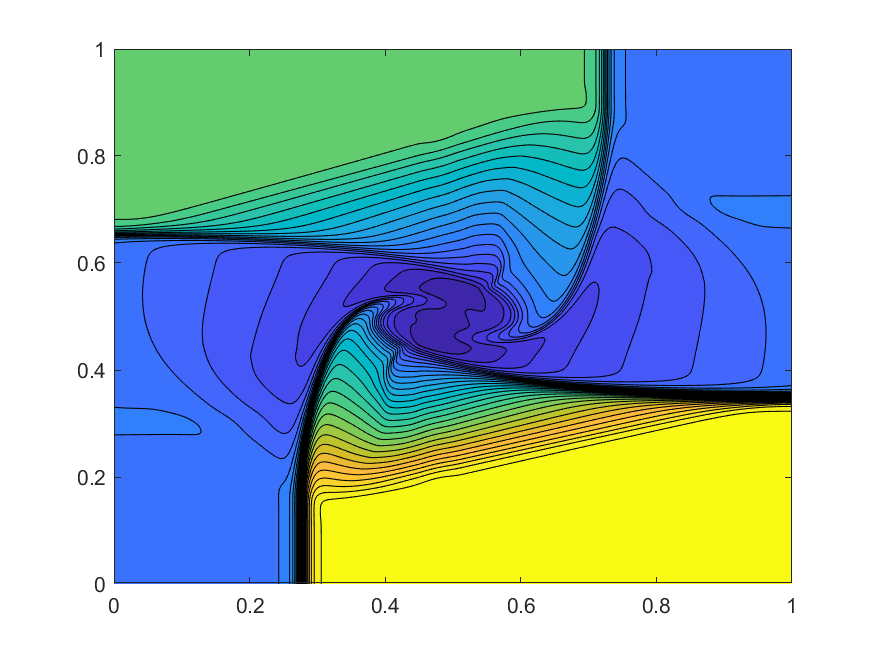}
\includegraphics[height=0.28\textwidth]{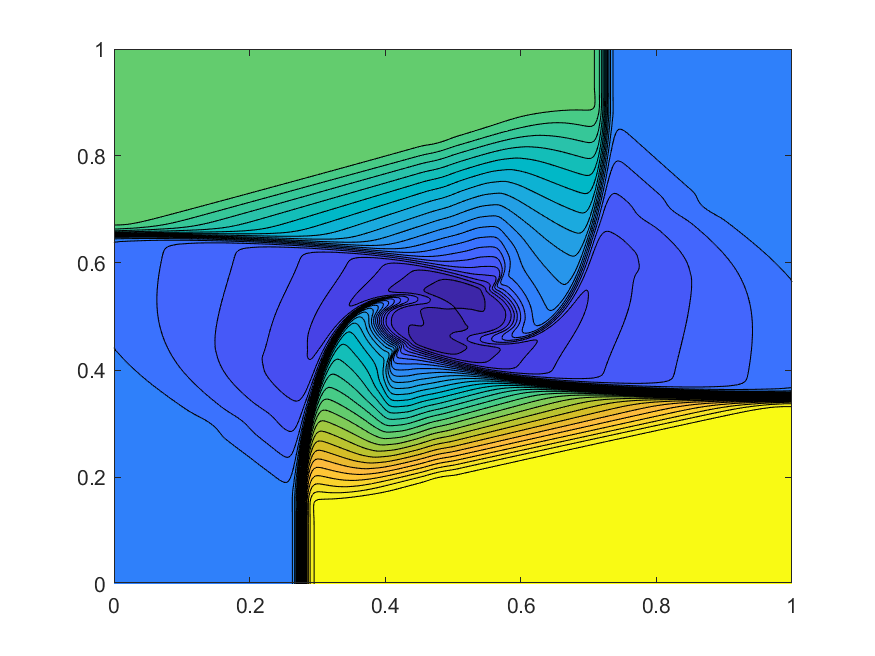}
\includegraphics[height=0.28\textwidth]{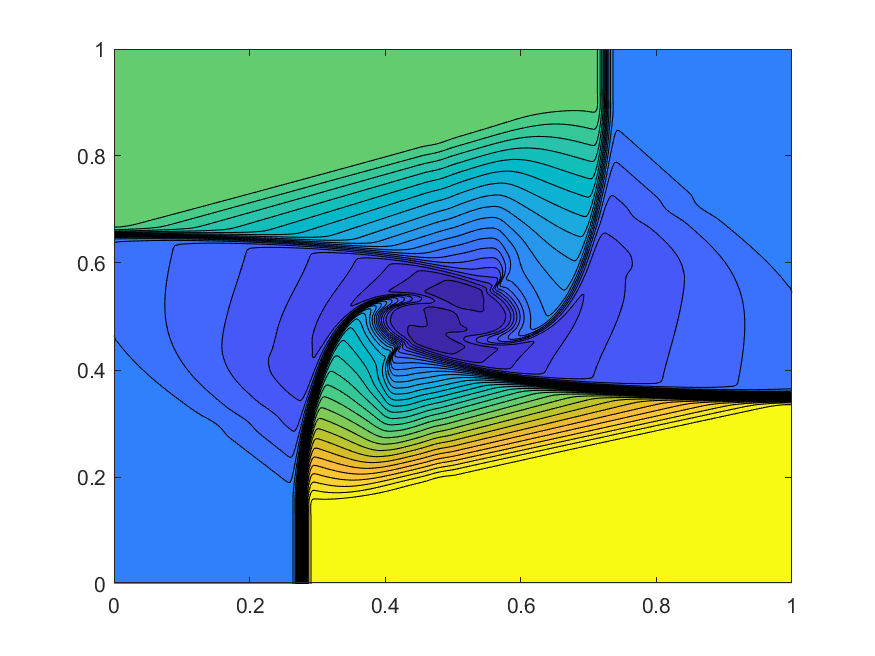}
\includegraphics[height=0.28\textwidth]{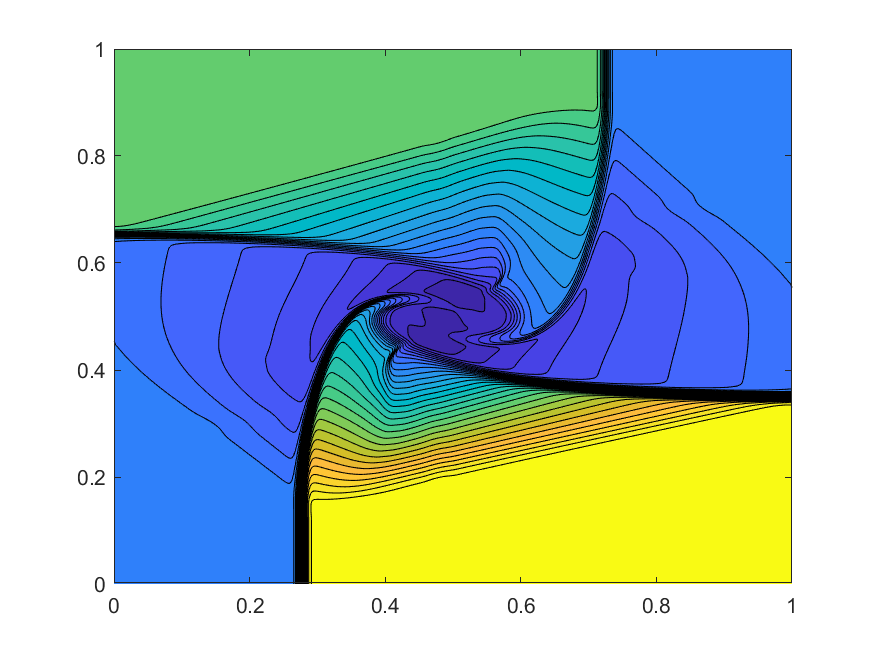}
\caption{Density in the filled contour plot for Example \ref{ex:euler_2d_riemann} at $T=0.3$ by WENO3-JS (top left), WENO3-Z (top right), WENO3-SNN1 (bottom left) and WENO3-SNN2 (bottom right) with $N_x = N_y = 400$.
Each contour plot displays contours at $30$ levels of the density.}
\label{fig:eer2d}
\end{figure}

\begin{example} \label{ex:euler_2d_riemann}
We first consider the Riemann problem \cite{Kurganov} to test our WENO schemes. 
The initial condition is
$$
   (\rho, u, v, P ) = \left\{ 
                       \begin{array}{ll} 
                        (1, \, 0.75, \, -0.5, \, 1),  & x > 0.5, \, y > 0.5, \\
                        (2, \, 0.75, \, 0.5, \, 1),   & x \leqslant 0.5, \, y > 0.5, \\
                        (1, \, -0.75, \, 0.5, \, 1),  & x \leqslant 0.5, \, y \leqslant 0.5, \\
                        (3, \, -0.75, \, -0.5, \, 1), & x > 0.5, \, y \leqslant 0.5.
                       \end{array}
                      \right.
$$
We divide the square computational domain $[0, \, 1] \times [0, \, 1]$ into $N_x \times N_y = 400 \times 400$ uniform cells.
The numerical solution of the density computed by WENO3-SNNs at the final time $T=0.3$ compared with those by WENO3-JS and WENO3-Z is presented in Fig. \ref{fig:eer2d}.
It is observed that WENO3-SNNs and WENO-Z produce richer structures of the vortex turning clockwise than WENO-JS.
\end{example}

\begin{figure}[htbp]
\centering
\includegraphics[height=0.28\textwidth]{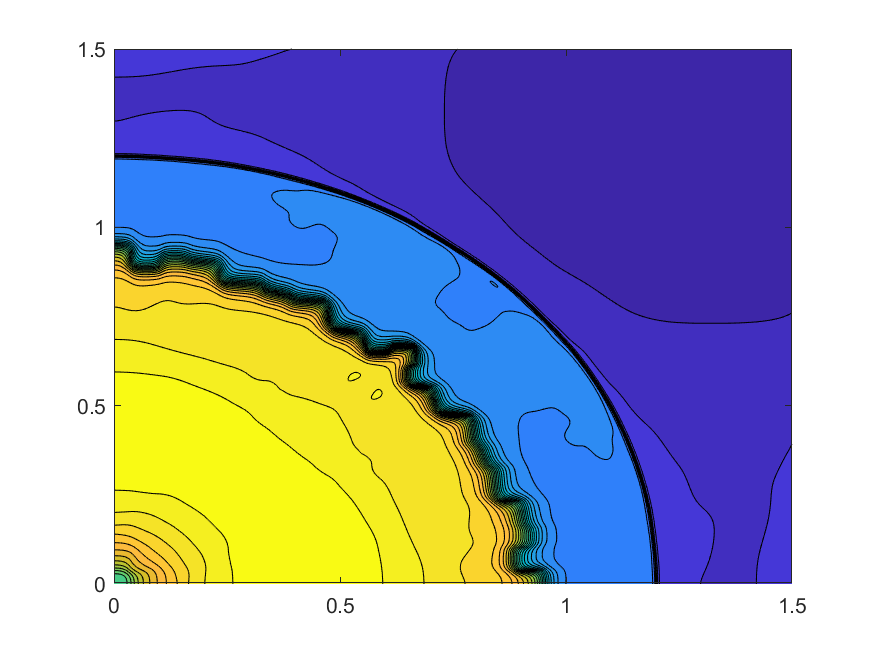}
\includegraphics[height=0.28\textwidth]{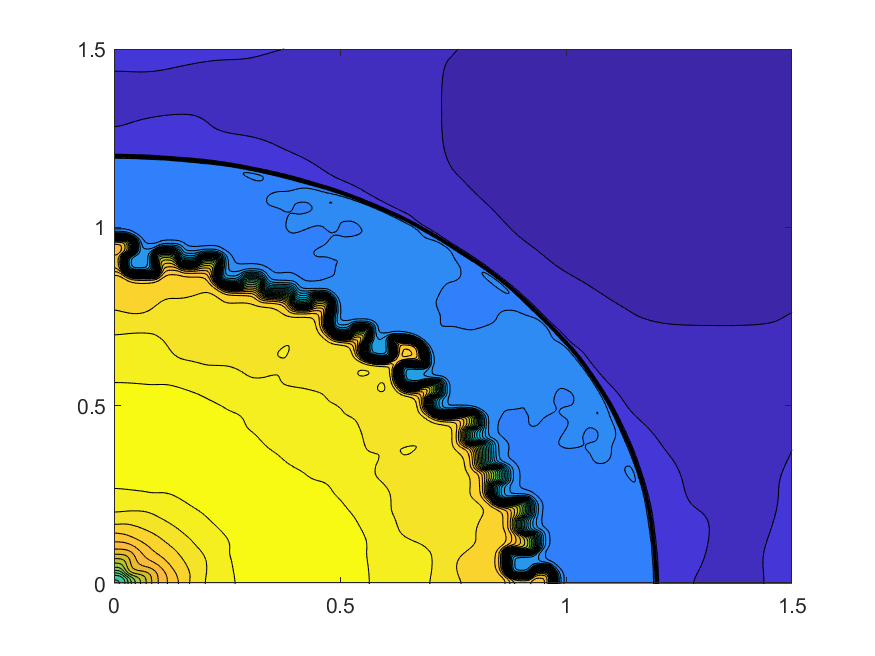}
\includegraphics[height=0.28\textwidth]{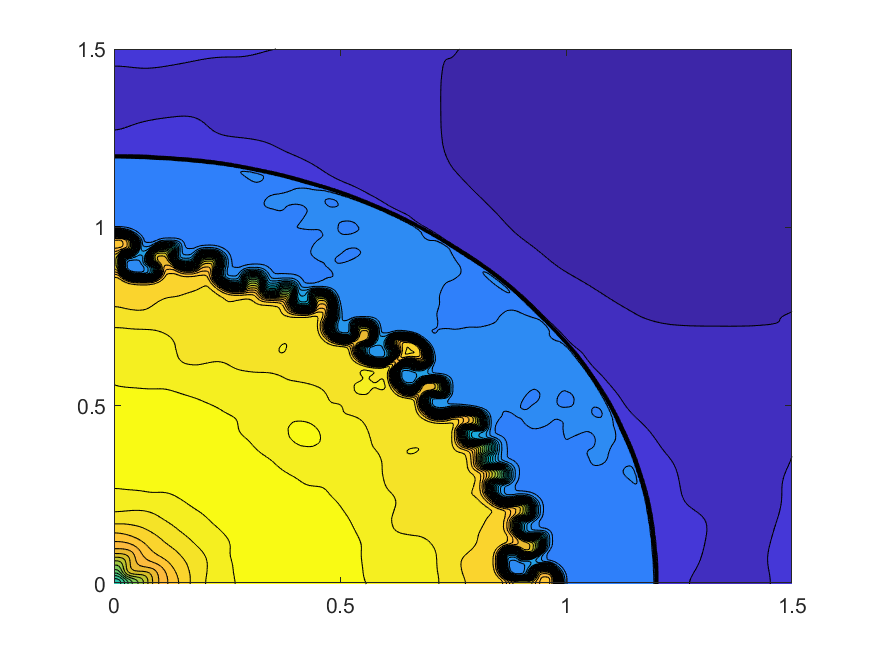}
\includegraphics[height=0.28\textwidth]{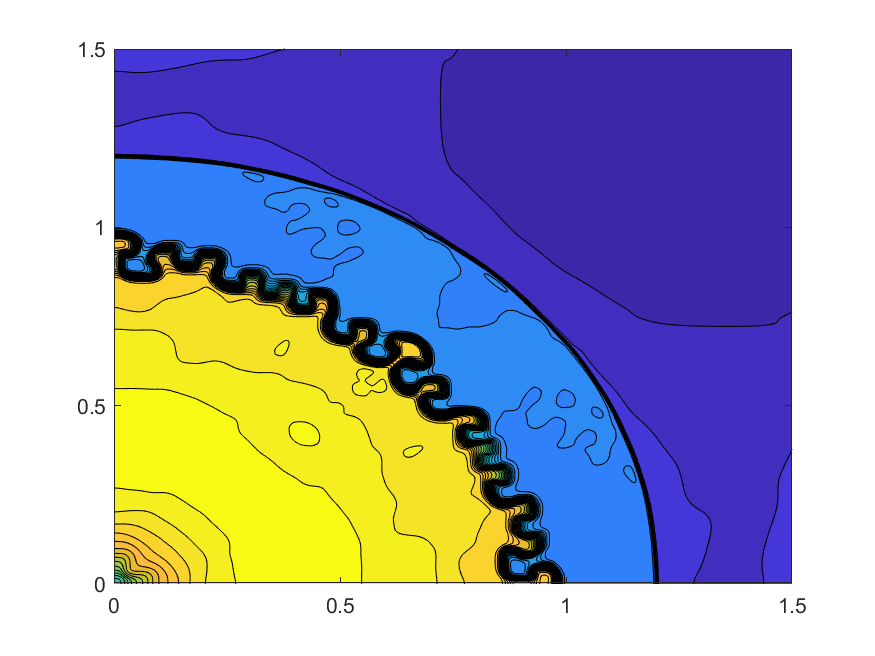}
\caption{Density in the filled contour plot for Example \eqref{ex:explosion} at $T=3.2$ approximated by WENO3-JS (top left), WENO3-Z (top right), WENO3-SNN1 (bottom left) and WENO3-SNN2 (bottom right) with $N_x = N_y=400$. }
\label{fig:explosion}
\end{figure}

\begin{example} \label{ex:explosion}
The explosion problem \cite{Liska}, which is a circularly symmetric problem, has an initial circular region of higher density and pressure:
$$
   (\rho, u, v, P ) = \left\{ 
                       \begin{array}{ll} 
                        (1, \, 0, \, 0, \, 1),       & x^2 + y^2 < 0.16, \\ 
                        (0.125, \, 0, \, 0, \, 0.1), & \text{otherwise}. 
                       \end{array} 
                      \right. 
$$
In this problem, the contact line develops instabilities as it is sensitive to the perturbations of the initially circular interface. 
The computational domain is $[0, \, 1.5] \times [0, \, 1.5]$ with $N_x = N_y = 400$ grid points. 
Fig. \ref{fig:explosion} plots the numerical density computed by the WENO schemes at the final time $T = 3.2$, where WENO3-SNNs and WENO-Z exhibit finer structures of contact curve than WENO-JS.
Besides, WENO3-SNN2 captures more complicated structures inside the circular region corresponding to the unstable contact wave. 
\end{example}

\begin{figure}[htbp]
\centering
\includegraphics[height=0.3\textwidth]{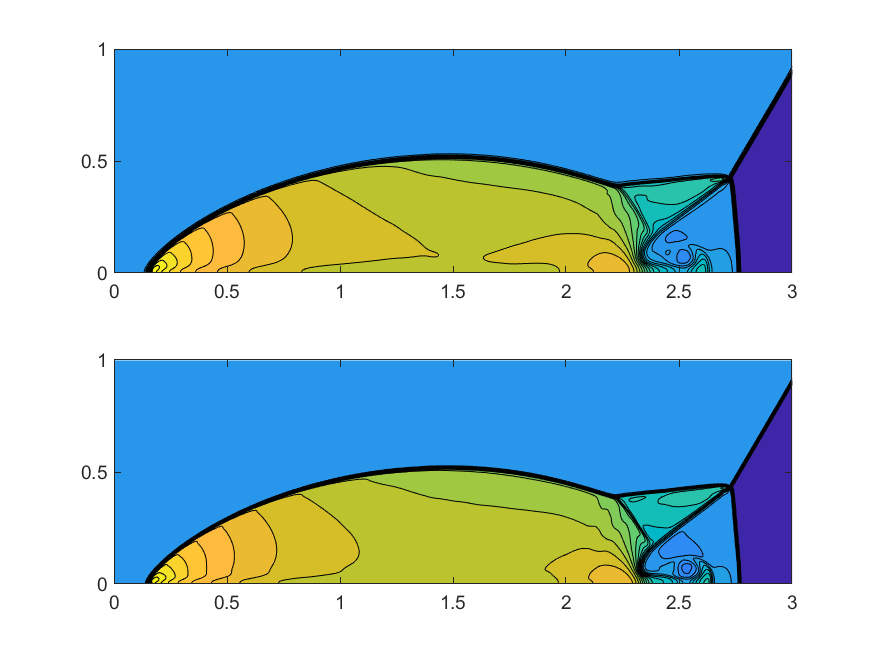}
\includegraphics[height=0.3\textwidth]{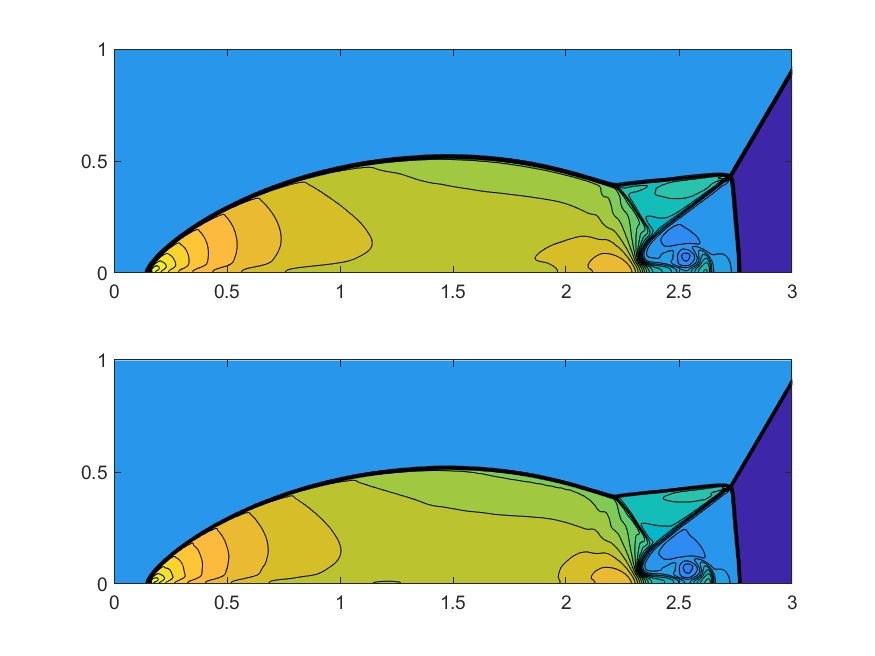}
\caption{Density in the filled contour plot for Example \ref{ex:double_mach_reflection} at $T=0.2$ by WENO3-JS (top left), WENO3-Z (top right), WENO3-SNN1 (bottom left) and WENO3-SNN2 (bottom right) with $N_x = 800$ and $N_y = 200$.
Each contour plot displays contours at $30$ levels of the density.}
\label{fig:dmr}
\end{figure}

\begin{figure}[htbp]
\centering
\includegraphics[height=0.28\textwidth]{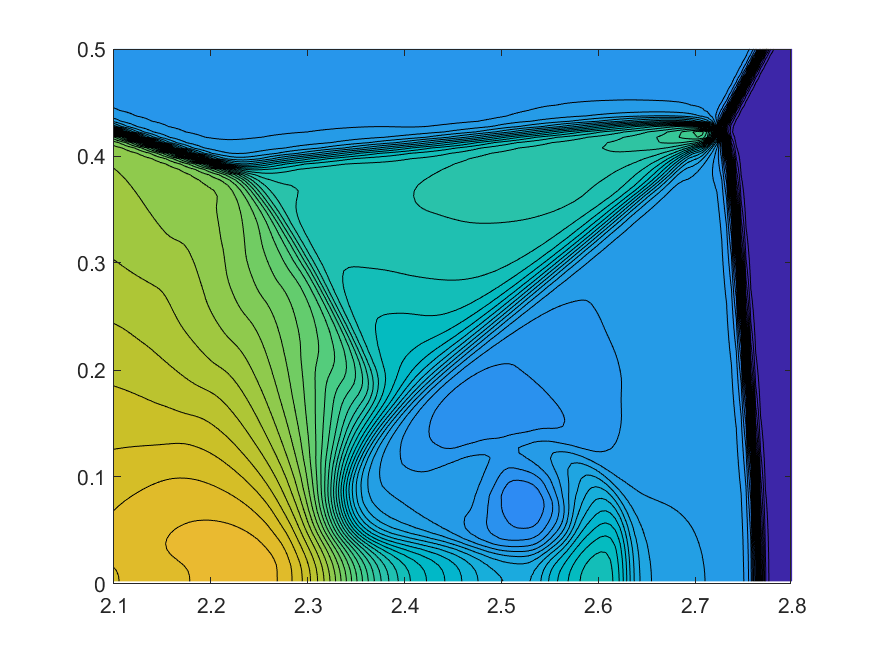}
\includegraphics[height=0.28\textwidth]{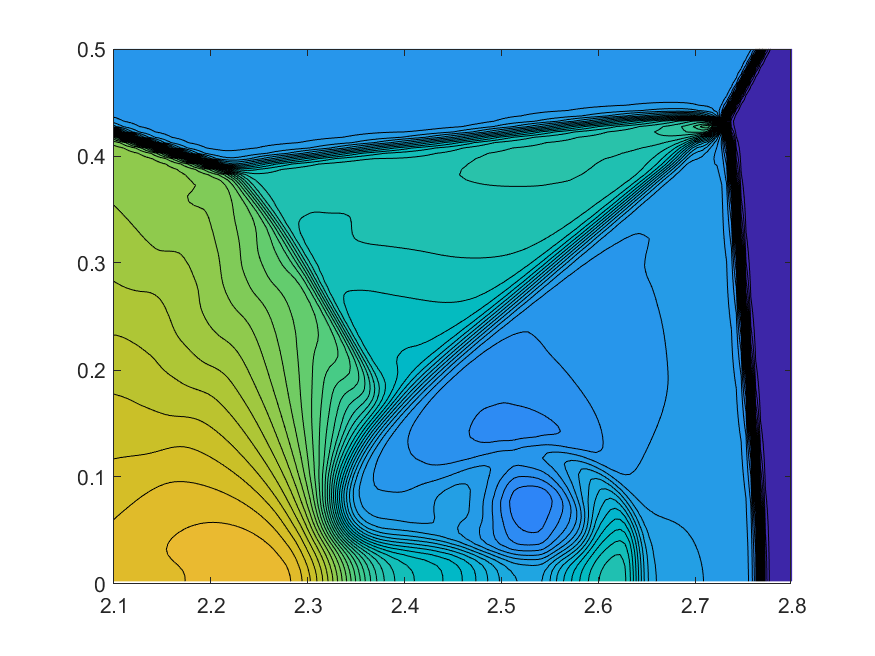}
\includegraphics[height=0.28\textwidth]{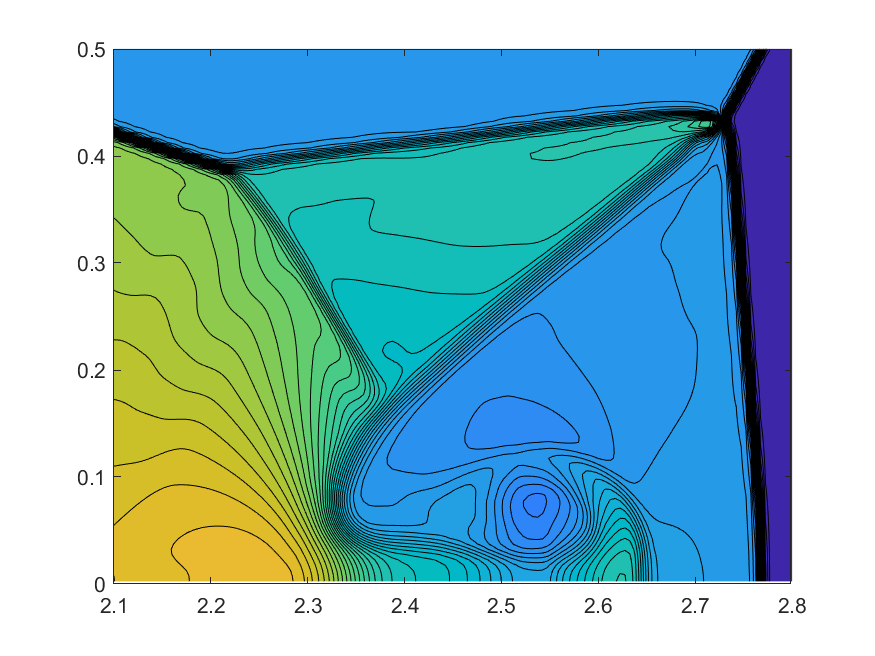}
\includegraphics[height=0.28\textwidth]{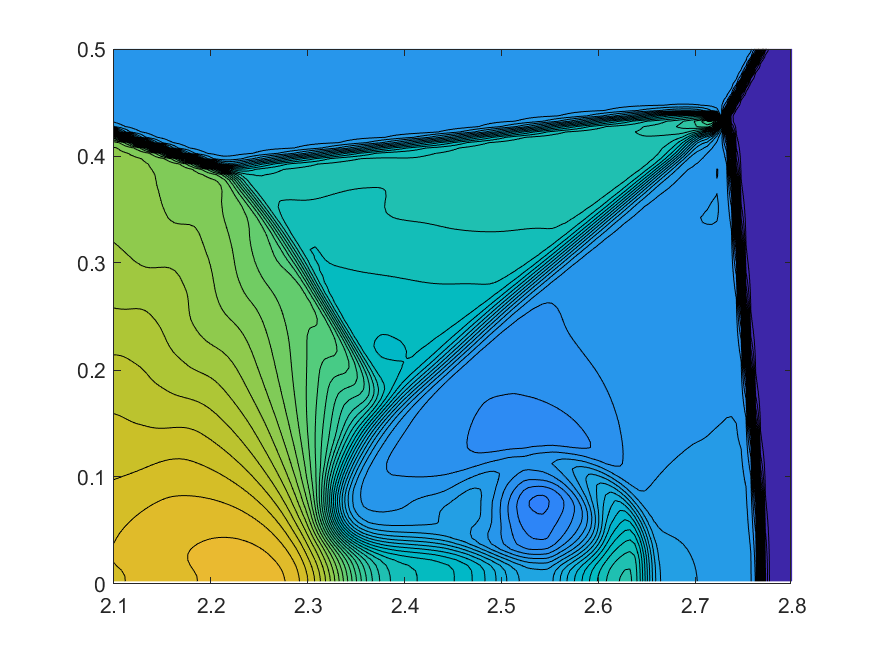}
\caption{Zooming-in density in the filled contour plot for Example \ref{ex:double_mach_reflection} at $T=0.2$ by WENO3-JS (top left), WENO3-Z (top right), WENO3-SNN1 (bottom left) and WENO3-SNN2 (bottom right) with $N_x = 800$ and $N_y = 200$.
Each contour plot displays contours at $30$ levels of the density.}
\label{fig:dmr_zoom}
\end{figure}

\begin{example} \label{ex:double_mach_reflection}
In this example, we consider the double Mach reflection problem introduced by Woodward and Colella \cite{Woodward}.
The initial condition is given by 
$$
   (\rho, u, v, P ) = \left\{ 
                       \begin{array}{ll} 
                        (8, \, 8.25 \cos \theta, \, -8.25 \sin \theta, \, 116.5), & x < \frac{1}{6} + \frac{y}{\sqrt{3}}, \\ 
                        (1.4, \, 0, \, 0, \, 1), & x \geqslant \frac{1}{6} + \frac{y}{\sqrt{3}}, 
                       \end{array} 
                      \right. 
$$
with $\theta = \frac{\pi}{6}$.
We divide the computational domain $[0,4] \times [0,1]$ into $N_x \times N_y = 800 \times 200$ uniform cells. 
The simulation is carried out until the final time $T = 0.2$, when the strong shock, joining the contact surface and transverse wave, sharpens.
We show the density profile of each WENO scheme in $[0,3] \times [0,1]$ at the final time in Fig. \ref{fig:dmr}.
We further zoom in on the solution for the region $[2.2, 2.8] \times [0, 0.5]$ in Fig. \ref{fig:dmr_zoom}.
We can see that WENO3-SNNs better capture the wave structures near the second triple point, and predicts a stronger jet near the wall.
\end{example}

\begin{figure}[htbp]
\centering
\includegraphics[height=0.3\textwidth]{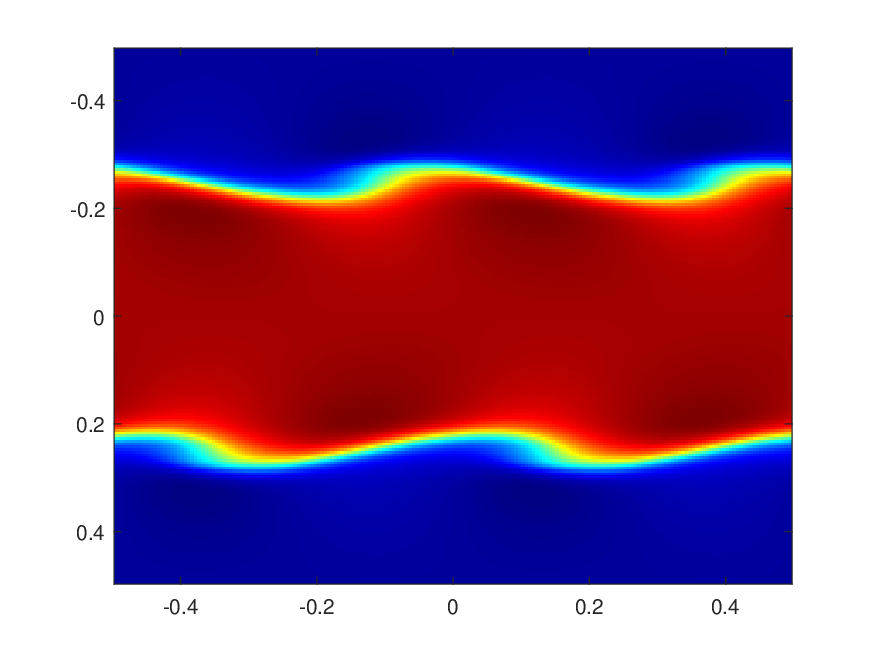}
\includegraphics[height=0.3\textwidth]{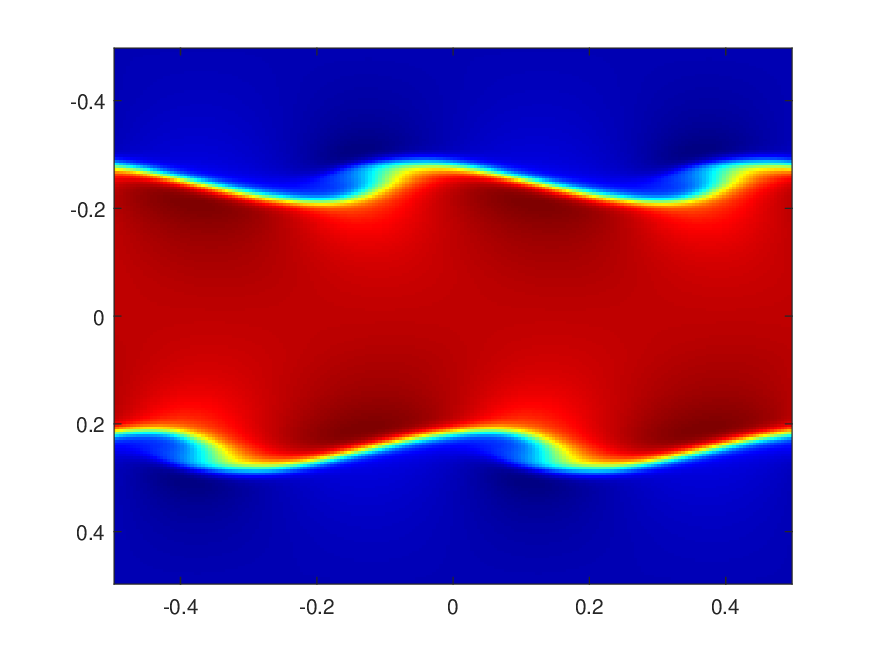}
\includegraphics[height=0.3\textwidth]{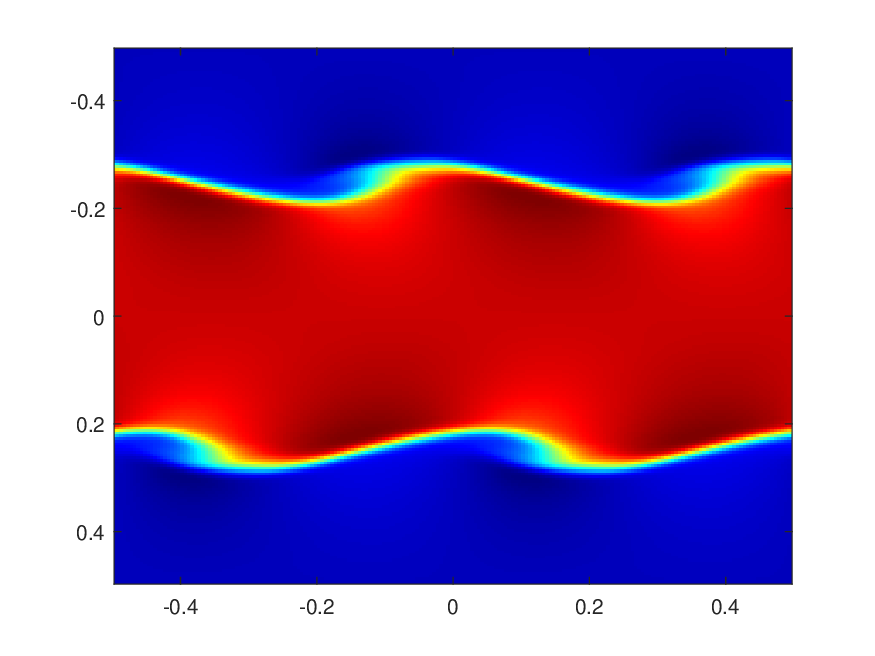}
\includegraphics[height=0.3\textwidth]{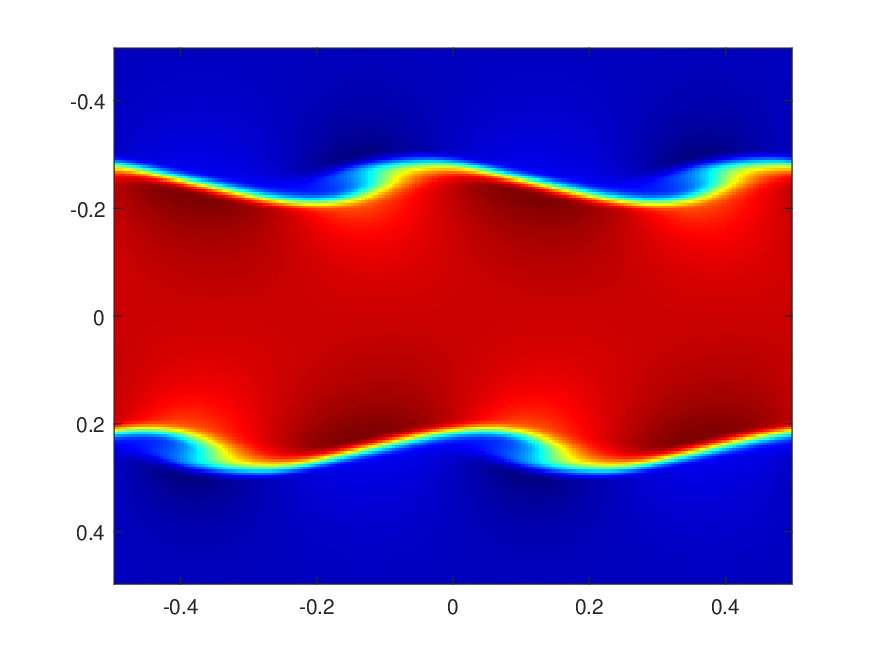}
\caption{Density in the plot of scaled colors for Example \ref{ex:kelvin_helmholtz_instability} at $t=1$ by WENO3-JS (top left), WENO3-Z (top right), WENO3-SNN1 (bottom left) and WENO3-SNN2 (bottom right) with $N_x = N_y = 200$.}
\label{fig:khi_t1}
\end{figure}

\begin{figure}[htbp]
\centering
\includegraphics[height=0.3\textwidth]{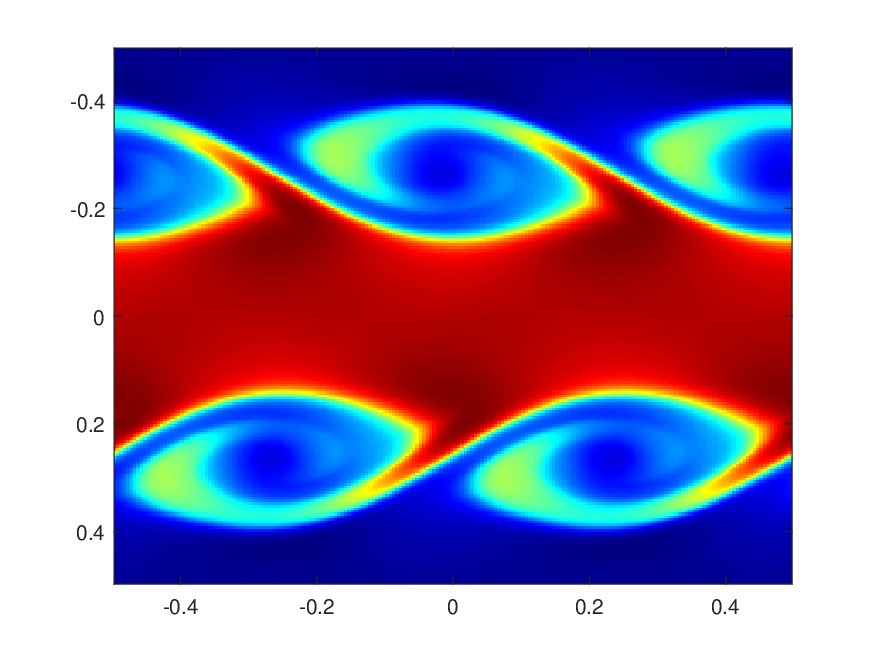}
\includegraphics[height=0.3\textwidth]{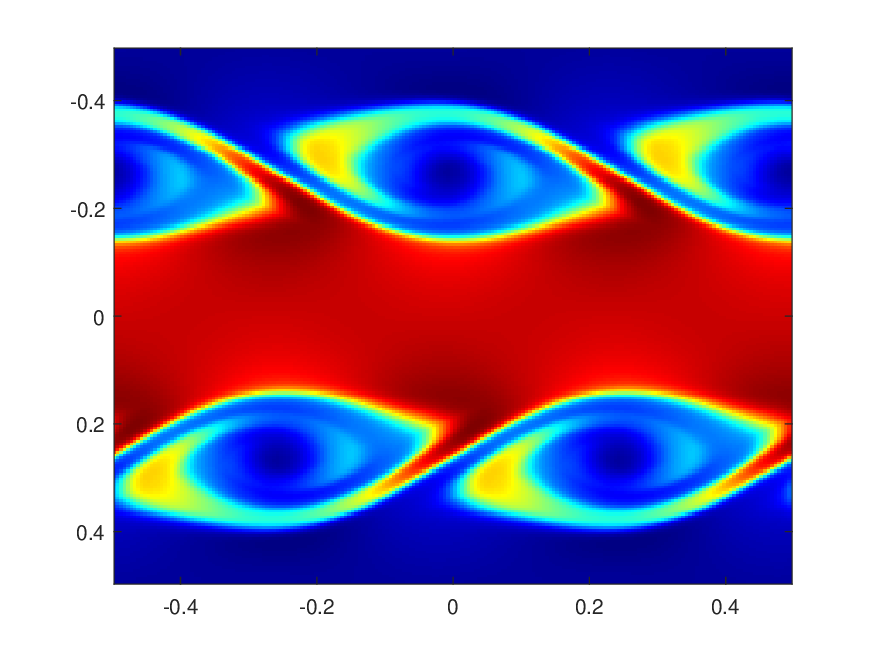}
\includegraphics[height=0.3\textwidth]{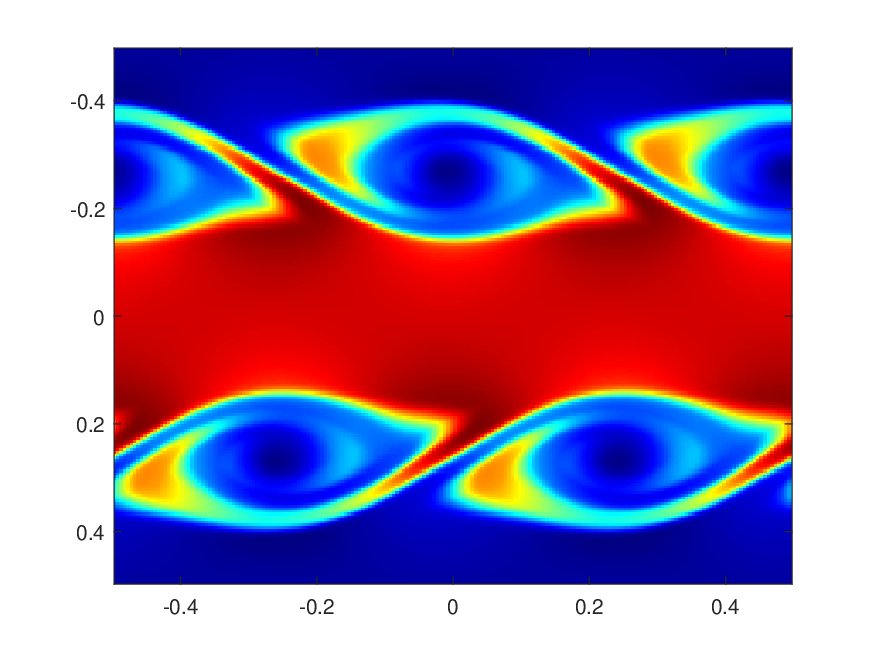}
\includegraphics[height=0.3\textwidth]{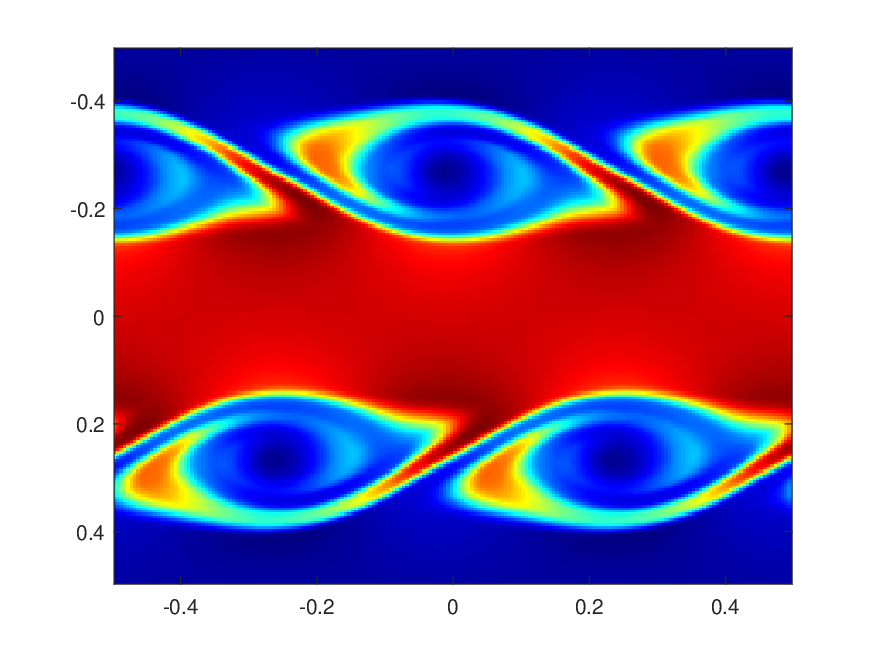}
\caption{Density in the plot of scaled colors for Example \ref{ex:kelvin_helmholtz_instability} at $t=2.5$ by WENO3-JS (top left), WENO3-Z (top right), WENO3-SNN1 (bottom left) and WENO3-SNN2 (bottom right) with $N_x = N_y = 200$.}
\label{fig:khi_t2}
\end{figure}

\begin{figure}[htbp]
\centering
\includegraphics[height=0.3\textwidth]{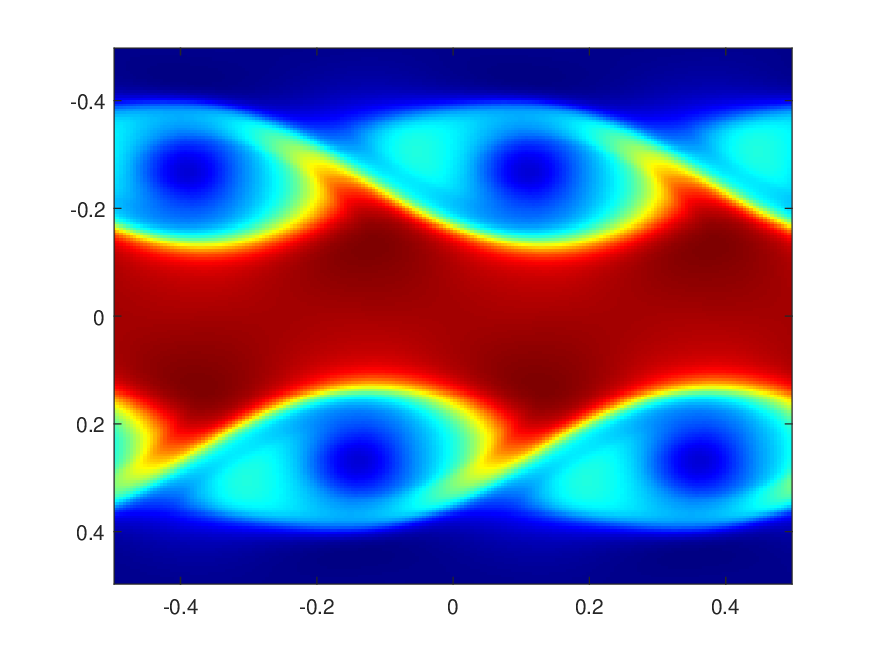}
\includegraphics[height=0.3\textwidth]{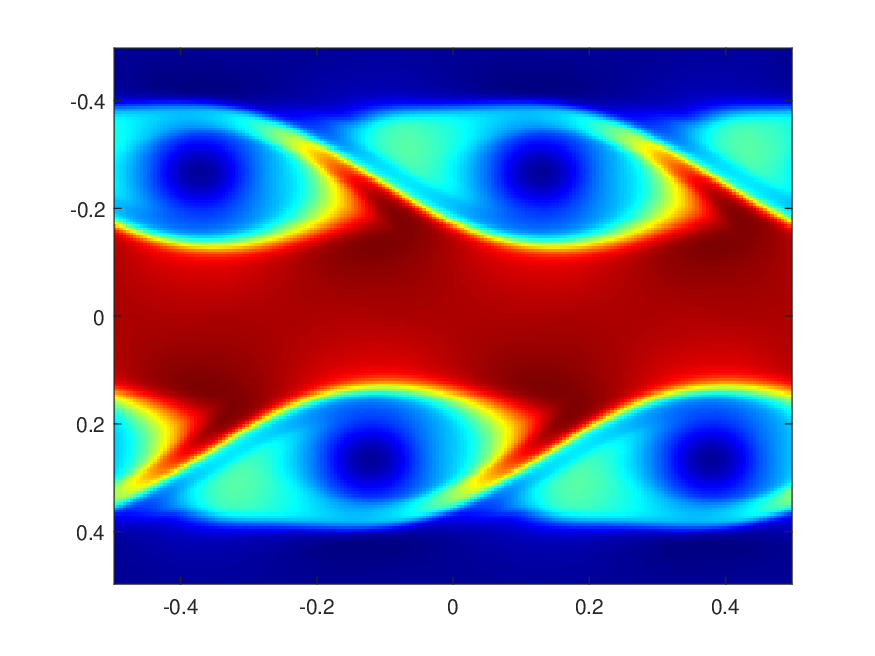}
\includegraphics[height=0.3\textwidth]{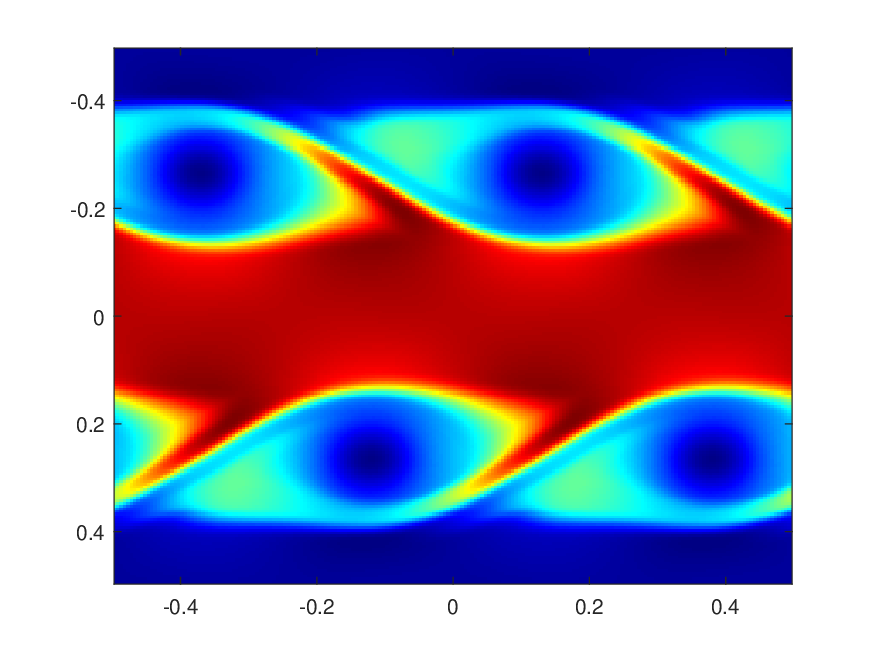}
\includegraphics[height=0.3\textwidth]{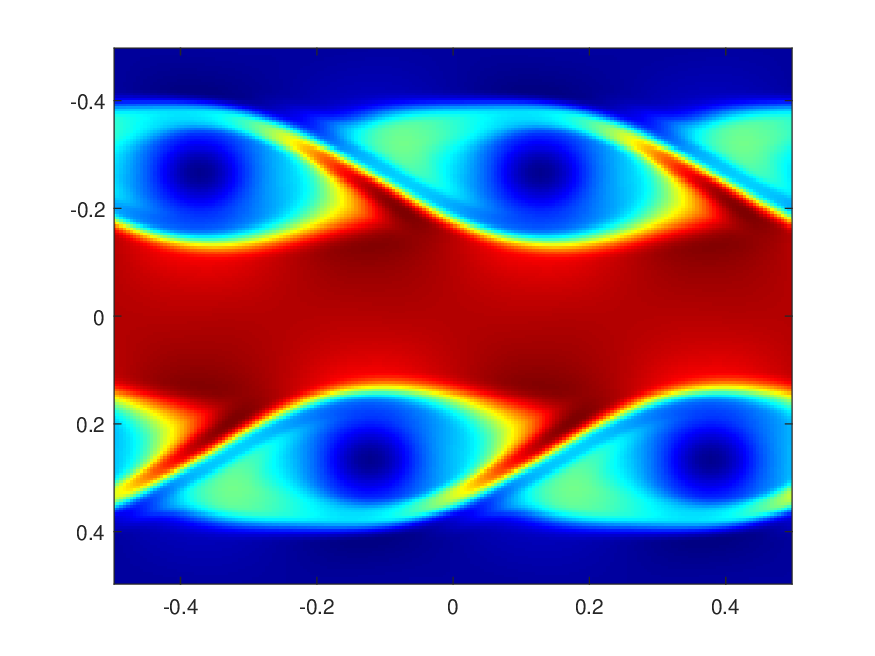}
\caption{Density in the plot of scaled colors for Example \ref{ex:kelvin_helmholtz_instability} at $T=4$ by WENO3-JS (top left), WENO3-Z (top right), WENO3-SNN1 (bottom left) and WENO3-SNN2 (bottom right) with $N_x = N_y = 200$.}
\label{fig:khi_T4}
\end{figure}

\begin{example} \label{ex:kelvin_helmholtz_instability}
We end this section with the Kelvin-Helmholtz (KH) instability, which has the initial condition \cite{Garg},
\begin{align*}
 & \left( \rho(x,y,0), u(x,y,0) \right) = \left\{ 
                                           \begin{array}{ll} 
                                            (1, \, -0.5+0.5 \e^{(y+0.25)/L}),  & -0.5 \leqslant y < -0.25, \\
                                            (2, \, 0.5-0.5 \e^{(-y-0.25)/L}),  & -0.25 \leqslant y < 0, \\
                                            (2, \, 0.5-0.5 \e^{(y-0.25)/L}),   & 0 \leqslant y < 0.25, \\
                                            (1, \, -0.5+0.5 \e^{(-y+0.25)/L}), & 0.25 \leqslant y \leqslant 0.5,
                                           \end{array}
                                          \right. \\
 & v(x,y,0) = 0.01 \sin (4 \pi x), \ P(x,y,0) = 1.5,
\end{align*}
where $L = 0.00625$ is a smoothing parameter corresponding to a thin shear interface in the simulations.
We employ the uniform grid with $N_x = N_y = 200$ for the square computational domain $[-0.5,0.5] \times [-0.5,0.5]$.
The numerical solutions of the density at $t = 1, \, 2.5$ and the final time $T=4$ are plotted in Figs. \ref{fig:khi_t1}, \ref{fig:khi_t2} and \ref{fig:khi_T4}, respectively.
In Fig. \ref{fig:khi_t1}, WENO3-JS, WENO3-Z and WENO3-SNNs at $t=1$ produce comparable swirl structures.
At later times $t = 2.5$ and $T=4$, WENO3-SNNs and WENO3-Z displays more complex turbulent structures, as shown in Figs. \ref{fig:khi_t2} and \ref{fig:khi_T4}, indicating that they can capture the KH instability and achieves a better resolution of KH vortices along the interface.
\end{example}

\section{Conclusion} \label{sec:conclusion} 
In this paper, we propose the WENO schemes based on the shallow neural network. 
The neural network integrates the Delta layer to the architecture of the shallow neural network. 
We define two loss functions with MSE and MSLE. 
Using WENO3-JS as the labels, we design the loss functions as the weighted sum of two errors with WENO3-JS and linear weights, respectively. 
The the neural network is trained to learn the linear weights for smooth regions and the WENO3-JS weighting function for discontinuities.
Numerical results indicate the improved behavior of less dissipation around discontinuities while preserving the ENO behavior in smooth regions for two proposed WENO schemes WENO3-SNNs.
We would like to upgrade to the fifth-order WENO scheme with the neural network in our future work.

\section*{Acknowledgments}
The research of the first author was supported by Institute of Information $\&$ Communications Technology Planning $\&$ Evaluation (IITP) grant funded by the Korea government(MSIT) (No.RS-2019-II191906, Artificial Intelligence Graduate School Program(POSTECH)).
The research of the fourth and fifth authors was supported by POSTECH Basic Science Research Institute under the NRF grant number 2021R1A6A1A1004294412 and 2021R1A6A1A10042944.
The research of the fifth author is supported by National Research Foundation of Korea under the grant number 2021R1A2C3009648 and partially by NRF MSIT (RS-2023-00219980).


\providecommand{\bysame}{\leavevmode\hbox to3em{\hrulefill}\thinspace}
\providecommand{\MR}{\relax\ifhmode\unskip\space\fi MR }
\providecommand{\MRhref}[2]{%
  \href{http://www.ams.org/mathscinet-getitem?mr=#1}{#2}
}
\providecommand{\href}[2]{#2}

\end{document}